\documentclass[preprint,12pt,nonatbib]{elsarticle}
\usepackage[utf8]{inputenc}                
\usepackage[x11names]{xcolor}              
\usepackage[
  colorlinks=true,        
  citecolor=ForestGreen,  
  linkcolor=MidnightBlue, 
  urlcolor=Teal,          
  hypertexnames=false     
]{hyperref}
\usepackage[natbibapa]{apacite}

\makeatletter
\newcommand{\apaseven@doihref}[1]{\href{https://doi.org/#1}{https://doi.org/#1}\endgroup}
\AtBeginDocument{%
  \renewcommand{\doi}{\begingroup\@sanitize\catcode`\\=0\relax\apaseven@doihref}%
}
\makeatother
\usepackage{etoolbox}
\AtBeginDocument{%
  \apptocmd{\thebibliography}{\raggedright}{}{}%
}
\usepackage{multibib}
\newcites{supp}{Supplementary References}
\usepackage{changepage}

\usepackage{graphicx}
\usepackage{placeins}
\usepackage{float}

\usepackage{booktabs}
\usepackage{multirow}
\usepackage{caption}
\usepackage{subcaption}
\usepackage{booktabs, makecell}
\usepackage{array}
\usepackage{siunitx}
\usepackage{rotating}
\newcolumntype{R}{S[table-format=1.3]}
\newlength{\gwotcolwd}
\newlength{\gwotcolaux}
\newcommand{\setgwotcolwd}[2]{%
  \settowidth{\gwotcolwd}{#1}%
  \addtolength{\gwotcolwd}{-2\tabcolsep}%
  \setlength{\gwotcolwd}{0.5\gwotcolwd}%
  \settowidth{\gwotcolaux}{#2}%
  \addtolength{\gwotcolaux}{-6\tabcolsep}%
  \setlength{\gwotcolaux}{0.25\gwotcolaux}%
  \ifdim\gwotcolaux>\gwotcolwd\setlength{\gwotcolwd}{\gwotcolaux}\fi
}
\newcommand{\gwotcell}[1]{\makebox[\gwotcolwd][c]{#1}}

\usepackage{amssymb}
\usepackage{dsfont}
\usepackage{amsmath}
\DeclareUnicodeCharacter{2022}{\ensuremath{\bullet}}
\DeclareUnicodeCharacter{2194}{\ensuremath{\leftrightarrow}}
\DeclareUnicodeCharacter{2212}{\ensuremath{-}}
\DeclareUnicodeCharacter{2248}{\ensuremath{\approx}}
\DeclareUnicodeCharacter{1D434}{\ensuremath{\mathbf{A}}}
\DeclareUnicodeCharacter{1D437}{\ensuremath{\mathbf{D}}}
\DeclareUnicodeCharacter{1D43E}{\ensuremath{\mathbf{K}}}
\DeclareUnicodeCharacter{1D43F}{\ensuremath{\mathbf{L}}}
\DeclareUnicodeCharacter{1D456}{\ensuremath{i}}
\DeclareUnicodeCharacter{1D457}{\ensuremath{j}}
\DeclareUnicodeCharacter{1D458}{\ensuremath{k}}
\DeclareUnicodeCharacter{1D459}{\ensuremath{l}}
\DeclareUnicodeCharacter{1D45A}{\ensuremath{m}}
\DeclareUnicodeCharacter{1D460}{\ensuremath{s}}
\DeclareUnicodeCharacter{1D461}{\ensuremath{t}}
\DeclareUnicodeCharacter{1D472}{\ensuremath{\boldsymbol{K}}}
\DeclareUnicodeCharacter{1D70C}{\ensuremath{\rho}}
\usepackage{xcolor}

\usepackage{lineno}

\journal{Neural Networks}

\begin{document}

\begin{frontmatter}



\title{Relational Knowledge Distillation Brings\\
DNN Representations Close Enough to Humans\\
to Be Aligned Without Supervision}
\author[1]{Yuria Shimizu}
\author[1]{Soh Takahashi}
\author[2]{Takato Horii}
\author[1]{Masafumi Oizumi}
\ead{c-oizumi@g.ecc.u-tokyo.ac.jp}
\affiliation[1]{organization={Graduate School of Arts and Sciences, The University of Tokyo},
addressline={3-8-1 Komaba},
postcode={153-8902},
city={Meguro, Tokyo},
country={Japan}}
\affiliation[2]{organization={Graduate School of Engineering Science, The University of Osaka},
addressline={1-3 Machikaneyama-cho},
postcode={560-8531},
city={Toyonaka, Osaka},
country={Japan}}

\begin{abstract}
    Linking the internal representations of deep neural networks (DNNs) to human mental representations is important for using DNNs as computational models of human vision. Existing DNN representations remain insufficiently similar to human mental representations, which are not directly observable and are therefore commonly measured through large-scale similarity judgments of object images. A natural approach to narrowing this gap is to directly transfer the relational structure of human representations into DNNs, and previous studies have reported improved human--DNN representational similarity. However, whether this improvement holds under stricter evaluation remains untested in two respects: fine-grained alignment at the individual-object level, and generalization to a human embedding derived from a dataset independent of the training data. Here, we employ an unsupervised comparison method, Gromov--Wasserstein optimal transport (GWOT), which estimates human--DNN correspondences from the internal distance structure alone and thereby tests fine-grained alignment. We further assess generalization on a curated test set of concepts non-overlapping with the training data. We show that fine-tuning pre-trained DNNs with Relational Knowledge Distillation (RKD), an established relational transfer method, brings DNNs close enough to humans to be aligned at the individual-object level on this test set. We also show that this improvement is driven by a more human-like global structure, as reflected in the ordering of distances among coarse categories, while the local human--DNN nearest-neighbor overlap rate remains largely unchanged. These findings indicate that relational transfer from humans brings the global structure of pre-trained DNNs close enough to the human structure to enable fine-grained human--DNN alignment without supervision.
\end{abstract}


\end{frontmatter}


\section{Introduction}
\label{Introduction}

    Deep neural networks (DNNs) have become the leading computational models of human vision, yet whether their internal representations correspond to the mental representations that humans form of objects remains an open question. DNNs have earned this status through their demonstrated ability to predict neural responses, measured with fMRI in humans and with electrophysiology in monkeys, as well as human object recognition behavior \citep{Yamins2014-xv, Khaligh-Razavi2014-qu, Guclu2015-jk, Kheradpisheh2016-jj, Cichy2016-dz, Storrs2021-dd, Conwell2024-ls}. However, predicting these responses does not establish that DNNs represent objects as humans do. This correspondence matters, because it would render the internal structure of a DNN interpretable in human terms. The DNN could then be understood through the same concepts by which humans organize objects. This interpretability, in turn, allows the mechanisms of human object representation to be probed through the DNN \citep{Kriegeskorte2015-bi, Yamins2016-uf}.

    Despite this motivation, current DNN representations remain insufficiently similar to human mental representations of objects. Human mental representations are not directly observable and are therefore measured, for example, through similarity judgments \citep{Hebart2020-pd, Roads2021-nz}. These judgments reveal the similarity structure of human representations: the more similar people judge two objects to be, the closer those objects lie in the representational space. This structure is relational, because it is defined by how objects stand in relation to one another rather than by properties of any object in isolation. Modern DNNs, in contrast, are typically optimized for supervised image classification or for self-supervised and weakly supervised representation-learning objectives \citep{He2016-je, Gidaris2018-qs, Chen2020-rt, Caron2020-um, Radford2021-fu}. None of these objectives targets the human similarity structure directly, and there is no guarantee that it emerges from them. Consistent with this, DNN representations capture this human similarity structure only to a limited extent \citep{Jozwik2017-ha, Peterson2018-hb}. Moreover, this limitation does not diminish as models become more accurate on standard benchmarks, because neither object classification accuracy nor model scale predicts agreement with human similarity judgments \citep{Roads2021-nz, Muttenthaler2022-mf}.

    A natural approach to narrowing this representational gap is to use the relational structure of human representations as an explicit training target for DNNs. Prior studies have used this structure to adapt DNN representations, either by learning a transformation on top of a pre-trained network \citep{Peterson2016-hx, Peterson2018-hb, Attarian2020-bo, Jha2023-ou, Muttenthaler2023-aw} or by fine-tuning the network \citep{Fu2023-ds, Muttenthaler2025-nb}. These adaptations have improved the representational similarity between humans and DNNs, as measured against the human similarity judgments that each study targeted.

\begin{figure}[p]
    \centering
    \resizebox{0.971\textwidth}{!}{%
        \includegraphics{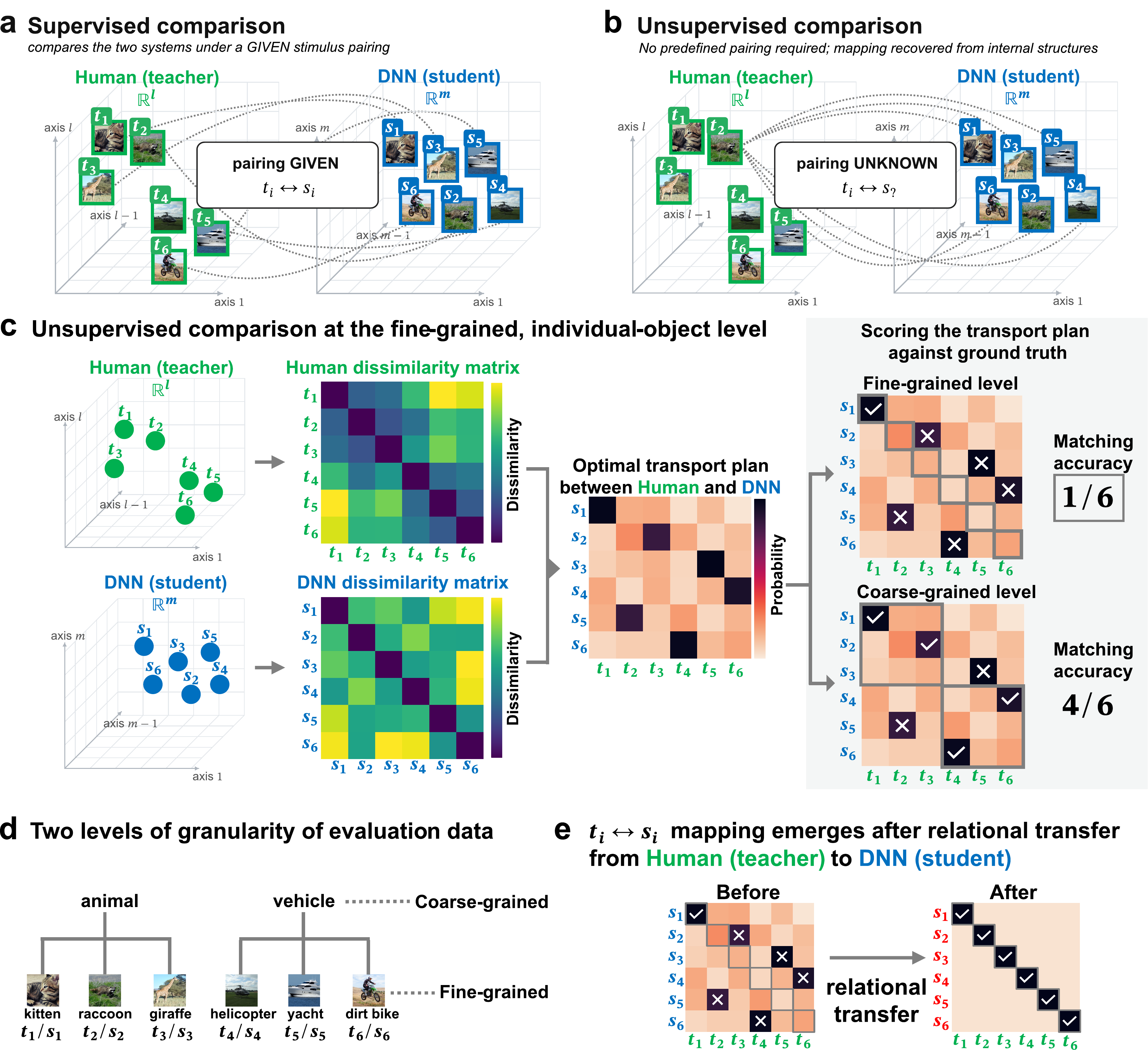}%
    }
    \caption{\textbf{Evaluating relational transfer from humans to DNNs by unsupervised comparison at the fine-grained, individual-object level.}
  \textbf{(a)}~Supervised comparison evaluates a human representation (teacher,
  $\mathbb{R}^{l}$) and a DNN representation (student, $\mathbb{R}^{m}$) under a
  pre-specified stimulus pairing ($t_i \leftrightarrow s_i$): the object
  correspondence is given, not inferred.
  \textbf{(b)}~Unsupervised comparison assumes no such pairing; the
  object-to-object correspondence is estimated from each space's internal
  distance structure alone.
  \textbf{(c)}~Unsupervised comparison evaluates the estimated correspondence at the
  fine-grained, individual-object level. Each
  representation is summarized as a dissimilarity matrix, and
  the method estimates a transport
  plan between the human and DNN dissimilarity matrices. Scoring this plan against the ground-truth identity of each individual object yields fine-grained matching. Scoring the same plan instead against the coarse category of each object yields coarse-grained matching (in this example, $1/6$ and $4/6$, respectively).
  \textbf{(d)}~The evaluation dataset has two levels of granularity: each individual object at the fine-grained level belongs to one coarse-grained category.
  \textbf{(e)}~When relational transfer brings the human and DNN representational structures closer, the transport plan reflects this correspondence as a pronounced identity diagonal, so individual objects become matchable.}
    \label{fig:intro}
\end{figure}

    However, the conventional methods used to evaluate these approaches cannot reveal whether this improved representational similarity reaches the fine-grained, individual-object level. In these methods, human and DNN representations are compared under a pre-specified pairing of stimuli. For example, the human representation of the stimulus ``dog'' is assumed to correspond to the DNN representation of ``dog''. We term this evaluation paradigm supervised comparison \citep{Takeda2025-za, Takeda2025-wi, Takahashi2026-lk} (Fig.~\ref{fig:intro}a). This paradigm is straightforward, and it quantifies how similar two representations are overall. Yet, because the object correspondence is given rather than inferred, supervised comparison does not test whether a different pairing of objects would make the two representations appear just as similar. Indeed, a simulation has demonstrated that two representations can remain highly similar under supervised comparison even when they correspond only at the coarse-category level and not at the individual-object level within each coarse category \citep{Takeda2025-wi}.

    Moreover, generalization of the human--DNN alignment beyond the human similarity dataset used for training remains untested at this fine-grained, individual-object level. This generalization matters because human similarity judgments cannot be collected for every object, so a model of human object representation must extend to objects for which no such judgments are available. Previous studies evaluated human--DNN alignment within the same dataset used for training, in some cases on a held-out split of that dataset or on an entirely held-out image domain \citep{Peterson2016-hx, Peterson2018-hb, Attarian2020-bo, Jha2023-ou, Fu2023-ds, Muttenthaler2023-aw, Muttenthaler2025-nb}. Some of these studies
    additionally evaluated it on a separate human similarity dataset \citep{Fu2023-ds, Muttenthaler2023-aw, Muttenthaler2025-nb}. All of these evaluations, however, relied on supervised comparison, so any generalization they reported could have been confined to the coarse-category level.

    Here, we use an unsupervised comparison method to test whether directly transferring the relational structure of human representations into DNNs brings them close enough to humans to be aligned at the fine-grained, individual-object level. Specifically, we adopt Gromov--Wasserstein optimal transport (GWOT) \citep{Memoli2011-nt, Peyre2016-dj}. GWOT aligns two representational spaces using only their internal distance structures, with no predefined pairing of objects. It estimates an optimal transport plan, a matrix whose entries give the probability that each human object corresponds to each DNN object. Among all possible transport plans, GWOT selects the one under which the pairwise distances among the human objects best match the pairwise distances among the DNN objects assigned to them. For example, the correspondence between the human representation of ``dog'' and the DNN representation of ``dog'' is not assumed but inferred from the optimal transport plan. We accordingly term this evaluation paradigm unsupervised comparison \citep{Takeda2025-za, Takeda2025-wi, Takahashi2026-lk} (Fig.~\ref{fig:intro}b). Crucially, this enables us to measure alignment at the fine-grained, individual-object level: fine-grained matching tests whether the most probable DNN match for each human object is the identical object itself, the strictest possible criterion (Fig.~\ref{fig:intro}c). The same transport plan can additionally be scored at the coarse-grained level, where any DNN match within the same coarse category counts as correct (Fig.~\ref{fig:intro}c,d). If transferring the relational structure brings the human and DNN representational structures closer, the transport plan should concentrate on the identity diagonal, so that individual objects become matchable (Fig.~\ref{fig:intro}e).

\begin{figure}[t]
    \centering
    \includegraphics[width=\textwidth]{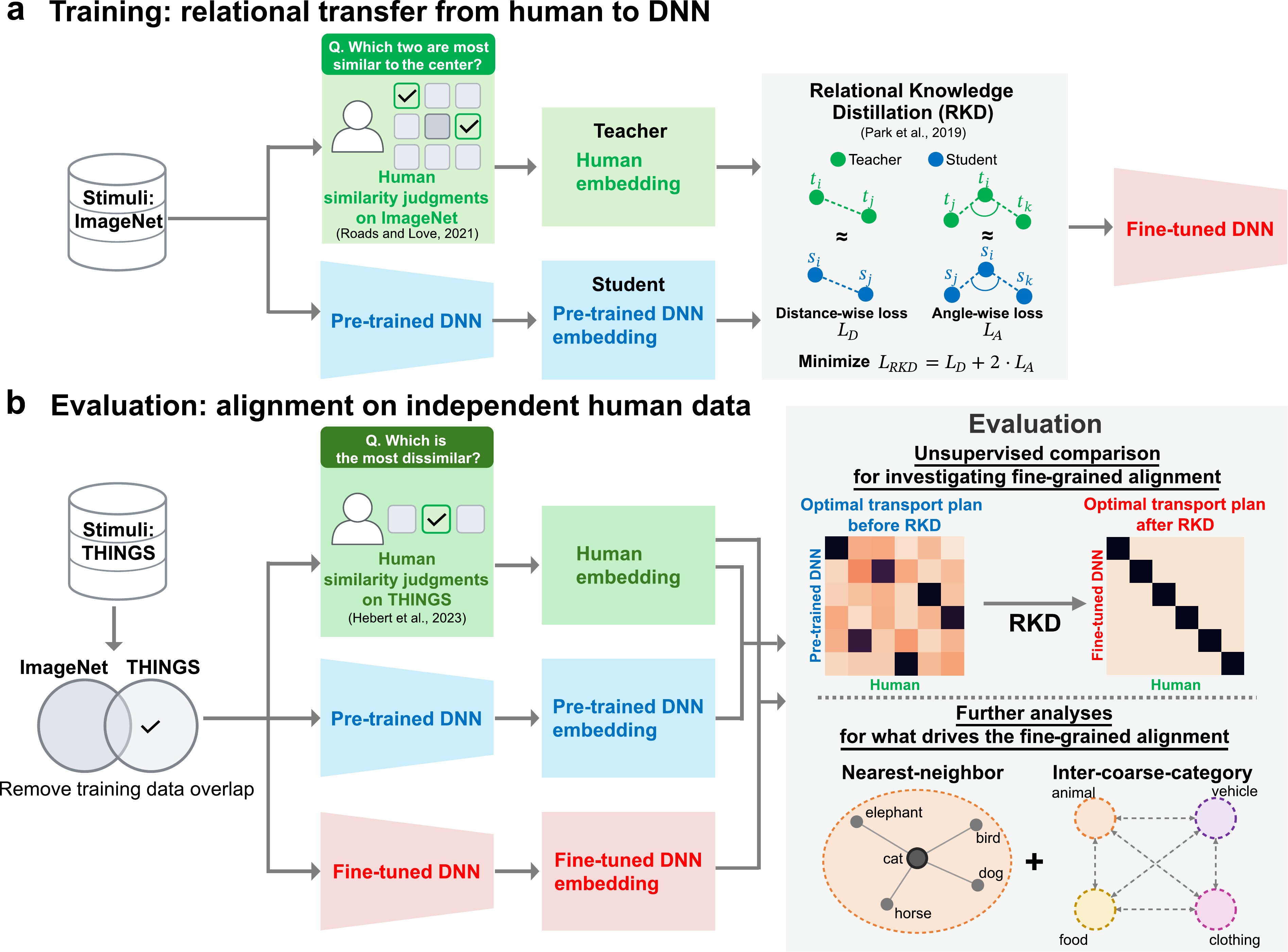}
    \caption{\textbf{Pipeline for relational transfer from humans to DNNs and its
    evaluation.}
    \textbf{(a)}~\textbf{Training.} We performed relational transfer using ImageNet stimuli. We obtained a human psychological embedding (teacher) from human similarity judgments on ImageNet images. We used a pre-trained DNN embedding, taken from the vision encoder's projection head, as the student. We then fine-tuned this DNN using Relational Knowledge Distillation (RKD; \citealp{Park2019-nt}) to bring the relations among objects in the DNN embedding closer to those in the human psychological embedding. Specifically, we minimized a distance-wise loss ($\mathcal{L}_{D}$) and an angle-wise loss ($\mathcal{L}_{A}$) defined over the relations among objects.
    \textbf{(b)}~\textbf{Evaluation.} To test generalization, we evaluated relational transfer using THINGS stimuli, after removing concepts that overlapped with the ImageNet training data. We then compared the relational structure of the human, pre-trained-DNN, and fine-tuned-DNN embeddings using an unsupervised comparison method, Gromov--Wasserstein optimal transport. To investigate what drove the fine-grained alignment, we further probed the underlying representational change. One analysis addressed the local neighborhood structure, and the other addressed the global structure. Within the global structure, we focused on how the coarse categories were arranged relative to one another.}
    \label{fig:pipeline}
    \end{figure}

    Beyond testing fine-grained alignment, we investigate whether this alignment generalizes to an independent human psychological embedding of concepts that do not overlap with the training data. To this end, we draw on two independent datasets of human similarity judgments. For training, we use judgments collected on ImageNet images \citep{Roads2021-nz}, from which we derive the human psychological embedding that serves as the target of the transfer (Fig.~\ref{fig:pipeline}a). For evaluation, we use a publicly released human psychological embedding derived from judgments collected on THINGS images \citep{Hebart2023-uf}. We then evaluate the pre-trained and the fine-tuned DNNs against this evaluation embedding. To separate the evaluation concepts from the training concepts, we systematically remove overlapping concepts using a WordNet-based criterion \citep{Miller1990-nw} (Fig.~\ref{fig:pipeline}b).

    We show that directly transferring human relational structure brings human and DNN representations into correspondence at the fine-grained, individual-object level. Moreover, this correspondence holds for an evaluation embedding whose concepts were curated not to overlap with the training data. To perform this transfer, we fine-tune pre-trained DNNs with Relational Knowledge Distillation (RKD; \citealp{Park2019-nt}), an established relational transfer method. RKD matches the pairwise distances and triplet-wise angles among objects across the human and DNN embeddings (Fig.~\ref{fig:pipeline}a). Using Representational Similarity Analysis (RSA; \citealp{Kriegeskorte2008-ep}), a representative example of supervised comparison methods, we confirm an overall increase in human--DNN similarity. Using GWOT, we further find that individual objects, not only coarse categories, become matchable. To investigate what drives this fine-grained gain, we perform two follow-up analyses (Fig.~\ref{fig:pipeline}b). These analyses trace the gain to a more human-like global structure, as reflected in the relative ordering of the distances among coarse categories. In contrast, the local neighborhood structure does not contribute to this gain, because its agreement with the human neighborhoods does not increase. Thus, relational transfer from humans brings the global structure of pre-trained DNNs close enough to the human structure to enable individual objects to be matched without supervision.

\section{Methods}
    \subsection{Datasets}
        To evaluate whether Relational Knowledge Distillation (RKD) generalizes to human representations that were not used for training, we used two independent large-scale human similarity judgment datasets on natural object images: (i) judgments collected on ImageNet images \citep{Roads2021-nz} served as the training dataset for RKD, and (ii) judgments collected on THINGS images \citep{Hebart2023-uf} served as the evaluation dataset. Each set of judgments provided its own human psychological embedding.
    
    \subsubsection{Training dataset: ImageNet}
    \label{sec:dataset_imagenet}
        For the RKD training data, we used the ILSVRC 2012 validation set of ImageNet \citep{Deng2009-ea, Russakovsky2015-il} as the image stimuli, paired with a large-scale set of human similarity judgments collected on these same images \citep{Roads2021-nz}. This image set contains 50{,}000 natural object images spanning 1{,}000 object classes. In each judgment trial, participants viewed a central query image surrounded by eight reference images and selected the two references most similar to the query (an 8-rank-2 trial format). Trials were selected using an active-learning paradigm that maximized information gain. We used the training split of the judgment data, as distributed through the PsiZ library (TensorFlow Datasets builder \texttt{ilsvrc2012\_val\_hsj}, builder version 1.0.1, built from the data release \texttt{ilsvrc2012\_val\_hsj\_v0.3.0} archived at \url{https://osf.io/7ck3s}), comprising 0.50 million similarity judgments on the 50{,}000 images, collected from 3{,}688 unique participants.

        For the human representation in our training, we derived an embedding that places the 50{,}000 ImageNet stimuli in a Euclidean space whose pairwise distances best reproduce these human similarity judgments. Specifically, we followed the procedure implemented in the Python library PsiZ \citep{Roads2019-jk, Roads2021-nz}. In this procedure, stimulus coordinates in $\mathbb{R}^{D}$, where $D$ is the dimensionality of the embedding space, are optimized to minimize the cross-entropy between predicted and observed 8-rank-2 choice probabilities \citep{Roads2019-jk}. Similarities are computed via an exponential kernel \citep{Shepard1987-ky, Roads2019-jk}. We selected $D$ and the initialization seed on a held-out validation split and left all other parameters at their default values. To this end, we performed a grid search over $D \in \{50, 75, 100, \ldots, 300\}$ crossed with five random initialization seeds, evaluated on a held-out 5\% of these judgments (95\%/5\% split). We selected the $(D, \mathrm{seed})$ combination with the lowest validation loss. This yielded $D = 125$ with seed 2. The resulting 125-dimensional embedding of all 50{,}000 ImageNet validation images constituted the teacher representation.

    \subsubsection{Evaluation dataset: THINGS}
    \label{sec:dataset_things}
        For the evaluation data, we used the THINGS image dataset \citep{Hebart2019-bn} as the image stimuli, paired with a large-scale set of human similarity judgments collected on these images \citep{Hebart2023-uf}. This image set contains 26{,}107 natural object images spanning 1{,}854 object concepts. In each judgment trial, participants viewed three object images and selected the most dissimilar one (a triplet odd-one-out format). Judgments were collected on a subset of 1{,}854 representative images (one per concept), yielding 4.70 million judgments from 12{,}340 unique participants.

        For the human representation in our evaluation, we used a psychological embedding that places the 1{,}854 THINGS concepts in a vector space whose pairwise dot products best reproduce these human similarity judgments. Unlike the teacher representation, which we fitted ourselves (Section~\ref{sec:dataset_imagenet}), this evaluation embedding was used exactly as publicly released \citep{Hebart2023-uf}. It was derived with the Sparse Positive Similarity Embedding (SPoSE) procedure \citep{Zheng2019-io}, which had previously been applied to the same 1{,}854 THINGS objects \citep{Hebart2020-pd}. This embedding is 66-dimensional, sparse, and non-negative. In this embedding, the similarity of two concepts is defined as the dot product of their coordinate vectors, and the coordinates were optimized so that these dot products predicted the observed odd-one-out choices.

        To assess whether RKD generalizes to object concepts not related to the training data, we removed the THINGS concepts that overlapped with the ImageNet training concepts. Specifically, we excluded the THINGS concepts whose WordNet synset \citep{Miller1990-nw} was hierarchically related to the synset of any of the 1{,}000 ILSVRC classes. We used WordNet version 3.0 throughout, accessed through the Python library NLTK \citep{Bird2009-nl}. Each THINGS concept is associated with a single WordNet synset (1{,}792 of the 1{,}854 concepts; the remaining 62 concepts carry no synset identifier in the THINGS release or one that is not a valid synset offset in WordNet 3.0, and we retained them), and each ILSVRC class likewise corresponds to a single synset. To capture both directions of the WordNet hierarchy, we performed a breadth-first search over WordNet hypernym and hyponym links starting from the synset of a THINGS concept. We traversed hypernym links up to the WordNet root, and hyponym links up to a maximum depth of 12. We thereby obtained our non-overlapping test set, consisting of the remaining 1{,}249 THINGS concepts. We excluded 605 concepts: 334 identical, 212 hypernym, and 59 hyponym concepts.

        \paragraph{Coarse-category reassignment}
            For our coarse-category-level evaluation, we assigned each THINGS object to exactly one coarse category. We based this assignment on the manual coarse-category annotation released with the THINGS database, which provides 27 categories \citep{Hebart2019-bn}. In this released annotation, 559 of the 1{,}854 objects carry no category label and 302 carry more than one. We therefore applied the following procedure to all 1{,}854 THINGS objects. First, we assigned every object with no category label to \emph{others}. Second, we moved each multiply assigned object to whichever of its categories contained the most objects. Third, we merged every category that then held fewer than 10 objects into \emph{others}. Among the objects left in these small categories, only ``milkshake'' (originally in both ``dessert'' and ``drink'') had a second original category that survived the merge, so it was reassigned to ``drink'', not to \emph{others}. This procedure assigned each of the 1{,}854 objects to exactly one coarse category, yielding 20 categories plus an \emph{others} category, which contained 570 objects.

    \subsection{Deep neural networks}
        We used Contrastive Language--Image Pre-training (CLIP) models \citep{Radford2021-fu}, because previous studies have shown that CLIP representations are more strongly aligned with human mental representations than the representations of other DNNs \citep{Muttenthaler2022-mf, Takahashi2026-lk}. As the vision encoder architecture, we adopted Vision Transformer B/16 (ViT-B/16) \citep{Dosovitskiy2020-ay}, a widely used model whose Base size, denoted by the ``B'', allowed us to run RKD within our computational resources. To test whether the effects of RKD are consistent across different pre-training data, we selected five CLIP ViT-B/16 variants: LAION-2B \citep{Schuhmann2022-lx}, LAION-400M \citep{Schuhmann2021-dn}, DataComp-L, DataComp-XL \citep{Gadre2023-dr}, and the OpenAI dataset \citep{Radford2021-fu}. For all five models, we extracted the 512-dimensional embedding from the vision encoder's projection head as the feature representation. 
           
        To examine whether the effect of RKD depends on the ViT-B/16 architecture and on CLIP-style image--text pre-training, we further examined seven additional DNN models. These models together span two architectures (convolutional ResNet and Vision Transformer) and four pre-training schemes (image--text contrastive, supervised classification, self-supervised, and no pre-training). We list these models and report the results in Supplementary Section~\ref{sec:supp_additional}.

    \subsection{Relational Knowledge Distillation}\label{sec:methods_rkd}
        To transfer the relational structure of the human psychological embedding into the DNNs, we adopted Relational Knowledge Distillation (RKD), an established method for relational transfer \citep{Park2019-nt}. RKD was originally developed to transfer knowledge from one model to another. It trains the student model to reproduce the relations among the teacher model's output embeddings, namely their pairwise distances and triplet-wise angles. We applied this relational transfer with a human teacher in place of a model teacher. Specifically, we used the human psychological embedding as the teacher and the CLIP vision encoder as the student.

        We adopted RKD because it does not require the teacher and student spaces to share a coordinate system. The teacher embedding is 125-dimensional (Section~\ref{sec:dataset_imagenet}), whereas the student output is 512-dimensional, and their axes have no a priori correspondence. Matching the two spaces coordinate by coordinate would therefore require learning an additional transformation between them. RKD removes this step, because its relational quantities are scalars computed from pairs or triples of embeddings, regardless of the dimensionality of each space. We could thus reshape the DNN representation by updating the parameters of the DNN itself, rather than by fitting a separate transformation to its output.

        We used the RKD loss in its original formulation \citep{Park2019-nt}, without modification. It preserves the relations among objects through two complementary terms, a distance-wise loss and an angle-wise loss:

        \begin{equation}
            \mathcal{L}_{\text{RKD}} = \lambda_D \mathcal{L}_D + \lambda_A \mathcal{L}_A \label{eq:rkd}
        \end{equation}

        Here, $\mathcal{L}_D$ penalizes discrepancies in pairwise distances between the teacher and student embeddings, and $\mathcal{L}_A$ penalizes discrepancies in triplet-wise angles. We denote the teacher embedding of an object by $t$ and the student embedding of the same object by $s$. Following the original work, we set $\lambda_D = 1$ and $\lambda_A = 2$.

        We added no other term to this objective: neither a task-specific loss, such as a classification cross-entropy, nor a regularization term. This choice isolates the effect of transferring relational structure, because any additional objective would confound this effect with task-specific optimization. Preserving downstream task performance therefore lies outside the scope of this study. We quantify how this choice affects downstream performance, taking few-shot classification as a representative task. We report the results in Supplementary Table~\ref{tab:fewshot_results}. See Supplementary Section~\ref{sec:supp_fewshot} for details.

        Both terms of this loss take the same form: the discrepancy in a relational quantity between teacher and student, averaged over the tuples formed within a mini-batch. The Huber loss $l_\delta$ \citep{Huber1964-rb} with $\delta = 1$ measures each discrepancy,

        \begin{equation}
            l_\delta(a, b) = \begin{cases} \frac{1}{2}(a - b)^2 & \text{if } |a - b| \leq 1 \\ |a - b| - \frac{1}{2} & \text{otherwise} \end{cases} \label{eq:huber}
        \end{equation}

        where $a$ and $b$ are the teacher-side and student-side values of the relational quantity being compared. This loss grows quadratically for small discrepancies and linearly for large ones, so that a few outlying tuples do not dominate the total.

        For the distance-wise loss, $l_\delta$ is applied to normalized pairwise distances, averaged over all ordered pairs $(x_i, x_j)$ in the mini-batch $\mathcal{P}$:

        \begin{equation}
            \mathcal{L}_D = \frac{1}{|\mathcal{P}|}\sum_{(x_i, x_j) \in \mathcal{P}} l_\delta\!\left(\psi_D(t_i, t_j),\, \psi_D(s_i, s_j)\right) \label{eq:rkd-d}
        \end{equation}

        The distance-wise potential $\psi_D$ is a normalized Euclidean distance between two embeddings. For the teacher, $\psi_D(t_i, t_j) = \|t_i - t_j\|_2 / \mu_t$, which divides the distance between $t_i$ and $t_j$ by the mean non-zero pairwise distance $\mu_t$ among the teacher embeddings in the mini-batch. For the student, $\psi_D(s_i, s_j) = \|s_i - s_j\|_2 / \mu_s$ is defined in the same way, with $\mu_s$ computed over the student embeddings. This normalization removes the difference in absolute distance scale between the two spaces, so that the loss compares only the relative distance structure.

        For the angle-wise loss, $l_\delta$ is applied to the cosines of triplet-wise angles, averaged over all ordered triples $(x_i, x_j, x_k)$ in the mini-batch $\mathcal{T}$:

        \begin{equation}
            \mathcal{L}_A = \frac{1}{|\mathcal{T}|}\sum_{(x_i, x_j, x_k) \in \mathcal{T}} l_\delta\!\left(\psi_A(t_i, t_j, t_k),\, \psi_A(s_i, s_j, s_k)\right) \label{eq:rkd-a}
        \end{equation}

        The angle-wise potential $\psi_A$ is the cosine of an angle of the triangle formed by three embeddings. For the teacher, $\psi_A(t_i, t_j, t_k) = \langle e^{ij},\, e^{kj} \rangle$ is the cosine of the angle at vertex $t_j$, where $e^{ij} = (t_i - t_j) / \|t_i - t_j\|_2$ and $e^{kj} = (t_k - t_j) / \|t_k - t_j\|_2$ are the unit vectors pointing from $t_j$ to $t_i$ and to $t_k$. For the student, $\psi_A(s_i, s_j, s_k)$ is defined in the same way, using the student embeddings. As these vectors are unit-normalized, this potential is invariant to the absolute scale of each space, so that the loss compares only the relative angular structure.

        \paragraph{Implementation details: RKD training}\label{sec:rkd_training_setup}
            We fine-tuned each CLIP model with this objective under a shared optimization protocol, selecting only the learning rate separately for each model. We initialized every model from its publicly released OpenCLIP pre-trained weights \citep{Ilharco2021-oc, Cherti2022-fl}. We then updated the entire vision encoder (ViT backbone, layer normalization, and 768$\to$512 projection head), whereas the text encoder was not used in the forward pass. We optimized with Adam \citep{Kingma2014-av} at a constant learning rate for 50 epochs with a mini-batch size of 256, using all 50{,}000 ImageNet images. To select the learning rate, we held out the last five images of each of the 1{,}000 ImageNet classes (5{,}000 images in total) as a validation set and trained on the remaining 45 images per class (45{,}000 images in total), which kept the classes balanced in both sets. We compared three candidates ($1 \times 10^{-4}$, $1 \times 10^{-5}$, $1 \times 10^{-6}$) on this validation set, and $1 \times 10^{-6}$ was selected for all five models. Our implementation adapted the original open-source RKD code (\citealp{Park2019-nt}; available at \url{https://github.com/lenscloth/RKD}).

    \subsection{Evaluation methods}
        \subsubsection{Dissimilarity matrix construction}
        \label{sec:dissimilarity}
            We used the same definition of representational dissimilarity for all our evaluation methods. To capture both the angular and the distance structure that RKD optimizes, we combined cosine and Euclidean distances using the same $\lambda_A : \lambda_D = 2 : 1$ weighting as the RKD loss. Specifically, for each pair of embedding vectors $i$ and $j$, we computed the dissimilarity as
            \begin{equation}
                d(i, j) = \mathcal{N}\!\left[\, 2\,\mathcal{N}[d_{\cos}] + \mathcal{N}[d_{\mathrm{euc}}] \,\right](i, j),
                \label{eq:combined_distance}
            \end{equation}
            where $d_{\cos}$ and $d_{\mathrm{euc}}$ are the cosine and Euclidean distance matrices, and $\mathcal{N}[\cdot]$ denotes min-max normalization to $[0, 1]$ over all entries of a matrix, $\mathcal{N}[m] = (m - \min m) / (\max m - \min m)$. Finally, we set the diagonal of $d$ to zero.

        \subsubsection{Supervised comparison: Representational Similarity Analysis}
        \label{sec:methods_rsa}
            To quantify representational similarity between human and DNN spaces under a fixed object pairing, we computed the Spearman correlation between the human and DNN dissimilarity matrices (Eq.~\ref{eq:combined_distance}), following the standard Representational Similarity Analysis (RSA) procedure \citep{Kriegeskorte2008-ep}. We refer to RSA as a supervised comparison because it requires a pre-specified pairing of stimuli across the two spaces. We computed this correlation over the strict upper triangle of the two matrices, that is, over all unique object pairs excluding self-pairs.

        \subsubsection{Unsupervised comparison: Gromov--Wasserstein optimal transport}
        \label{sec:methods_gwot}
            To quantify representational similarity between human and DNN spaces without a fixed object pairing, we estimated an optimal coupling (transport plan $\Gamma$) between the human and DNN dissimilarity matrices (Eq.~\ref{eq:combined_distance}), following the Gromov--Wasserstein optimal transport (GWOT) procedure \citep{Memoli2011-nt, Peyre2016-dj}. We refer to GWOT as an unsupervised comparison because it is not given the pairing of stimuli across the two spaces and instead recovers the correspondence from the internal distance structure of each space alone. The GWOT objective is to find $\Gamma$ that minimizes
            \begin{equation}
                \min_{\Gamma} \sum_{i,j,k,l} (D_{ij} - D'_{kl})^2 \, \Gamma_{ik} \, \Gamma_{jl}, \label{eq:gwot_objective}
            \end{equation}
            where $D_{ij}$ is the dissimilarity between human objects $i$ and $j$ (the source space), $D'_{kl}$ is the dissimilarity between DNN objects $k$ and $l$ (the target space), and $\Gamma_{ik}$ can be interpreted as the probability that human object $i$ is aligned with DNN object $k$. We constrained $\Gamma$ to have uniform marginals, $\sum_{k} \Gamma_{ik} = \sum_{i} \Gamma_{ik} = 1/n$, where $n$ is the number of test objects, so that every object carries equal mass in both spaces.

            To handle the non-convexity of Eq.~\ref{eq:gwot_objective}, we adopted two safeguards against poor local optima. First, we ran the optimization from random initializations and retained the solution with the lowest GW cost. Second, we added an entropic regularization term to smooth the loss landscape \citep{Cuturi2013-sk, Peyre2016-dj}. We denote the discrete entropy of the transport plan by $H(\Gamma) = -\sum_{ik} \Gamma_{ik} (\log \Gamma_{ik} - 1)$ \citep{Peyre2019-ct}. The regularized objective is to find $\Gamma$ that minimizes
            \begin{equation}
                \min_{\Gamma} \sum_{i,j,k,l} (D_{ij} - D'_{kl})^2 \, \Gamma_{ik} \, \Gamma_{jl} - \varepsilon \, H(\Gamma),
            \end{equation}
            where $\varepsilon > 0$ is a hyperparameter controlling the trade-off between the original GWOT objective and the entropy term; a larger $\varepsilon$ drives $\Gamma$ closer to the uniform coupling, in which every entry equals $1/n^2$.

            \paragraph{Implementation details: GWOT optimization}
                We implemented GWOT and tuned its entropic regularization coefficient $\varepsilon$ with the GW-Tune toolbox \citep{Takeda2025-wi}, which uses the POT library \citep{Flamary2021-mg} for GWOT optimization and Optuna \citep{Akiba2019-yo} for hyperparameter tuning. We ran 1{,}000 optimization trials, each evaluating a different value of $\varepsilon$ drawn from the range $[10^{-4}, 10^{-3}]$. We adopted this range from a previous study of human--DNN alignment on the THINGS dataset \citep{Takahashi2026-lk}. We doubled the 500 trials used in that study to further reduce the risk of poor local optima. We did not run more trials because of the computational cost: each trial required a full GWOT optimization on dissimilarity matrices of up to $1{,}249 \times 1{,}249$, and we repeated this search for every model before and after RKD. We drew each value of $\varepsilon$ on a logarithmic scale using the Tree-structured Parzen Estimator (TPE) sampler \citep{Bergstra2011-mr}, which samples adaptively rather than uniformly over the range. In each trial, we randomly initialized the transport plan by sampling each element from a uniform distribution over $[0, 1]$ and normalizing the matrix to satisfy the marginal constraints. Across the 1{,}000 trials, we selected the transport plan with the lowest GW cost. We computed this cost without the entropy term, because $\varepsilon$ differs across trials.

            \paragraph{Matching accuracy}
                To evaluate how well the estimated transport plan $\Gamma$ recovers the ground-truth human--DNN correspondence, we computed the matching accuracy, the percentage of human source objects for which the DNN targets receiving the most transport mass include a correct target. We applied this criterion at two granularities, the individual-object (fine-grained) level and the coarse-category (coarse-grained) level, which differ only in what is counted as a correct target. As the same $n$ test objects are represented in both spaces, source and target objects share a common index. We write $\mathrm{Top}_K(\Gamma_{i,:})$ for the indices of the $K$ largest entries of row $i$, that is, the $K$ DNN target objects receiving the most transport mass from human source object $i$ (including all indices tied at the $K$-th largest value).

                At the individual-object (fine-grained) level, the criterion was the strictest possible: we counted source object $i$ as matched at rank $K$ only if the identical target object was among these $K$ targets,
                \begin{equation}
                    \mathrm{Match}^{\mathrm{fine}}_K(i) = \mathds{1}\!\left[ \, i \in \mathrm{Top}_K(\Gamma_{i,:}) \,\right].
                \end{equation}
                At the coarse-category (coarse-grained) level, any target object from the same coarse category counts as correct. Writing $\mathrm{cat}(i)$ for the coarse category of object $i$ (the 20 categories plus \emph{others} defined in Section~\ref{sec:dataset_things}), we counted source object $i$ as matched at rank $K$ if
                \begin{equation}
                    \mathrm{Match}^{\mathrm{coarse}}_K(i) = \mathds{1}\!\left[ \,\exists\, j \in \mathrm{Top}_K(\Gamma_{i,:}) \text{ such that } \mathrm{cat}(j) = \mathrm{cat}(i) \,\right].
                \end{equation}
                At both levels, we defined the top-$K$ matching accuracy as the percentage of source objects matched at rank $K$, that is, the mean of $\mathrm{Match}^{\mathrm{fine}}_K$ or $\mathrm{Match}^{\mathrm{coarse}}_K$ over all $n$ source objects, multiplied by 100. We report top-1 and top-5 matching accuracies at both levels.

                For each matching accuracy, we also report its chance level, the value expected from an uninformative transport plan. We defined such a plan as one whose $K$ largest entries in each row fall on $K$ target objects drawn uniformly at random from the $n$ target objects. At the individual-object (fine-grained) level, the chance level is then $K/n$. At the coarse-category (coarse-grained) level, it is the probability that at least one of the $K$ random targets shares the coarse category of the source object, averaged over all source objects.
        
        \subsubsection{Local structure: \texorpdfstring{$k$}{k}-nearest-neighbor overlap rate}
        \label{sec:methods_knn}
            To assess whether RKD increased the similarity of local neighborhood structure between human and DNN embedding spaces, we computed the $k$-nearest-neighbor overlap rate on the 1{,}249-object THINGS test set. This metric quantifies whether each object is surrounded by the same neighboring objects in the two spaces. Previous studies have used the same nearest-neighbor overlap to compare the local neighborhood structure of representations across models \citep{Huh2024-wv, Groger2026-pc}. For each query object, we identified the $k$ nearest neighbors in each space from the dissimilarity matrix (Eq.~\ref{eq:combined_distance}) by sorting the corresponding row in ascending order and selecting the $k$ objects with the smallest dissimilarities (excluding the query itself). We then computed the size of the overlap between the human and DNN neighbor sets for each query, averaged this overlap across all 1{,}249 queries, and divided it by $k$ to obtain the $k$-nearest-neighbor overlap rate. We computed this metric for $k = 5$ and $k = 10$. For each condition, we report the mean overlap rate across the 1{,}249 query objects together with a 95\% confidence interval. We computed this interval as the mean plus or minus $1.96$ times the standard error of the mean. We obtained this standard error as the sample standard deviation of the per-query overlap rates across the 1{,}249 query objects, divided by $\sqrt{1{,}249}$.

        \subsubsection{Global structure: three-dimensional projection of the embeddings}
        \label{sec:pca}
            To qualitatively compare the global structure across the embedding spaces, we visualized them in three dimensions. We projected each of the three embeddings (human, pre-RKD DNN, post-RKD DNN) separately with principal component analysis (PCA) using scikit-learn \citep{Pedregosa2011-sk} and retained the first three components. We then centered each of the three point clouds and divided it by its own Frobenius norm, so that the three clouds had the same overall size. Finally, we placed the two DNN clouds in the frame of the human cloud by orthogonal Procrustes alignment, which pairs each object with the same object in the human cloud \citep{Schonemann1966-op, Virtanen2020-sp}. This alignment applies a distance-preserving transformation, that is, a rotation or a rotation combined with a reflection, so these steps preserve the relative arrangement of the objects within each embedding. We restricted this visualization to the 794 objects of the 1{,}249-object THINGS test set that belonged to one of the 20 coarse categories defined in Section~\ref{sec:dataset_things}, that is, excluding the heterogeneous \emph{others} category.

        \subsubsection{Global structure: inter-coarse-category distance}
        \label{sec:inter_category}
            To quantitatively compare the global structure across the embedding spaces, we computed the Spearman correlation between the human and DNN coarse-category-level dissimilarity matrices. This analysis quantifies whether the coarse categories are arranged in the same relative order in the two spaces. It complements the GWOT coarse-category-level matching accuracy, which measures the extent to which the two spaces agree on coarse-category membership. We excluded the heterogeneous \emph{others} category. We then measured the distance between every pair of the remaining 20 coarse categories (794 objects) within each embedding space, using the standard Wasserstein distance \citep{Peyre2019-ct}. This distance treats each category as the distribution of its member objects rather than as a single centroid. We computed each distance with the entropy-regularized Sinkhorn algorithm \citep{Cuturi2013-sk}, taking the ground cost from the corresponding block of the object-level dissimilarity matrix (Eq.~\ref{eq:combined_distance}) and using uniform marginals, with the entropic regularization coefficient fixed at $0.05$ across spaces.

\FloatBarrier
\section{Results}
    \subsection{RKD brought DNN representations closer to humans on non-overlapping test concepts}
    \label{sec:results_alignment}
    
        \subsubsection{Supervised comparison: RSA correlation increased after RKD}
        \label{sec:rsa_results}
            To test whether the representational similarity gain from Relational Knowledge Distillation (RKD) generalized beyond the training data, we first performed Representational Similarity Analysis (RSA), a supervised comparison method. Specifically, we computed the Spearman correlation between the human and DNN dissimilarity matrices on the 1{,}249 THINGS object concepts of our non-overlapping test set.

            As expected, we found that the correlation between human and DNN representations increased for all five models after RKD, confirming that our RKD procedure successfully reshaped the DNN representations toward the human similarity structure. Specifically, the correlation rose from a pre-RKD range of $\rho = 0.407$--$0.445$ to a post-RKD range of $\rho = 0.527$--$0.568$ (Table~\ref{tab:rsa_gwot}). However, the increase in RSA correlation was moderate, and an overall correlation alone cannot reveal whether the alignment is at the fine-grained, individual-object level or only at the coarse-category level.

    \begin{table}[t]
        \centering
        \caption{\textbf{RSA and GWOT results for five CLIP ViT-B/16 variants before and after RKD on the 1{,}249-object THINGS test set.} Values in the RSA column are Spearman correlations ($\rho$) between human and DNN dissimilarity matrices. Values in the GWOT columns are matching accuracies. For each human object, the top-1 and top-5 columns consider, respectively, the single DNN object and the five DNN objects that receive the most transport mass from it. Fine-grained (individual-object-level) matching counts only the identical object as correct, whereas coarse-grained (coarse-category-level) matching counts any object from the same coarse category as correct. Chance-level values are listed for reference.}
        \label{tab:rsa_gwot}
        \small
        \setlength{\tabcolsep}{4pt}
        \setgwotcolwd{Coarse-grained}{GWOT matching accuracy (\%)}
        \begin{tabular}{@{}llRcccc}
            \toprule
            \multirow{3}{*}{Pre-training data} & & {\multirow{3}{*}{\makecell{RSA correlation\\(Spearman $\rho$)}}} & \multicolumn{4}{c}{GWOT matching accuracy (\%)} \\
            \cmidrule(lr){4-7}
            & & & \multicolumn{2}{c}{Fine-grained} & \multicolumn{2}{c}{Coarse-grained} \\
            \cmidrule(lr){4-5} \cmidrule(lr){6-7}
            & & & \gwotcell{Top-1} & \gwotcell{Top-5} & \gwotcell{Top-1} & \gwotcell{Top-5} \\
            \midrule
            \multirow{2}{*}{LAION-2B}
                & Before & 0.407 & 0.560 & 2.64 & 34.8 & 60.4 \\
                & After  & 0.568 & 19.5 & 42.5 & 61.2 & 90.6 \\
            \midrule
            \multirow{2}{*}{LAION-400M}
                & Before & 0.423 & 1.28 & 6.49 & 37.3 & 62.8 \\
                & After  & 0.548 & 16.1 & 37.7 & 58.7 & 89.7 \\
            \midrule
            \multirow{2}{*}{DataComp-L}
                & Before & 0.416 & 2.96 & 8.41 & 40.6 & 70.2 \\
                & After  & 0.527 & 14.7 & 38.4 & 57.6 & 88.4 \\
            \midrule
            \multirow{2}{*}{DataComp-XL}
                & Before & 0.436 & 2.24 & 7.21 & 38.5 & 68.0 \\
                & After  & 0.550 & 16.7 & 38.0 & 61.0 & 89.8 \\
            \midrule
            \multirow{2}{*}{OpenAI}
                & Before & 0.445 & 0.641 & 1.92 & 37.0 & 61.1 \\
                & After  & 0.547 & 13.1 & 32.1 & 57.6 & 88.8 \\
            \midrule
            Chance rate
                & & {N/A} & 0.0801 & 0.400 & 18.4 & 51.9 \\
            \bottomrule
        \end{tabular}%
    \end{table}

        \subsubsection{Unsupervised comparison: GWOT matching accuracy increased not only at the coarse-grained level but also at the fine-grained, individual-object level after RKD}
        \label{sec:gwot_results}
            To address the question left open by the RSA analysis, namely whether the similarity improved at the fine-grained level of individual objects or only at the coarse-grained level of categories, we employed Gromov--Wasserstein optimal transport (GWOT), an unsupervised method that estimates human--DNN matching from the internal distance structure alone, without the known stimulus pairing. The resulting object-to-object transport plan can be scored at both the fine-grained and the coarse-grained levels.

            We found that GWOT matching accuracy between human and DNN representations increased after RKD for all five CLIP ViT-B/16 variants, at both the fine- and coarse-grained levels (Table~\ref{tab:rsa_gwot}). At the fine-grained level, where only the single identical object out of the 1{,}249 test objects counts as a correct match (chance $0.0801\%$), top-1 matching accuracy rose from a pre-RKD range of $0.560\%$--$2.96\%$ to a post-RKD range of $13.1\%$--$19.5\%$, more than two orders of magnitude above chance. Before RKD, almost no individual object could be recovered. At the coarse-grained level, top-1 matching accuracy rose from a pre-RKD range of $34.8\%$--$40.6\%$ (chance $18.4\%$) to a post-RKD range of $57.6\%$--$61.2\%$. Top-5 matching accuracies showed comparable gains at both levels (Table~\ref{tab:rsa_gwot}). 
            
\begin{figure}[p]
            \centering
            \includegraphics[width=0.933\textwidth]{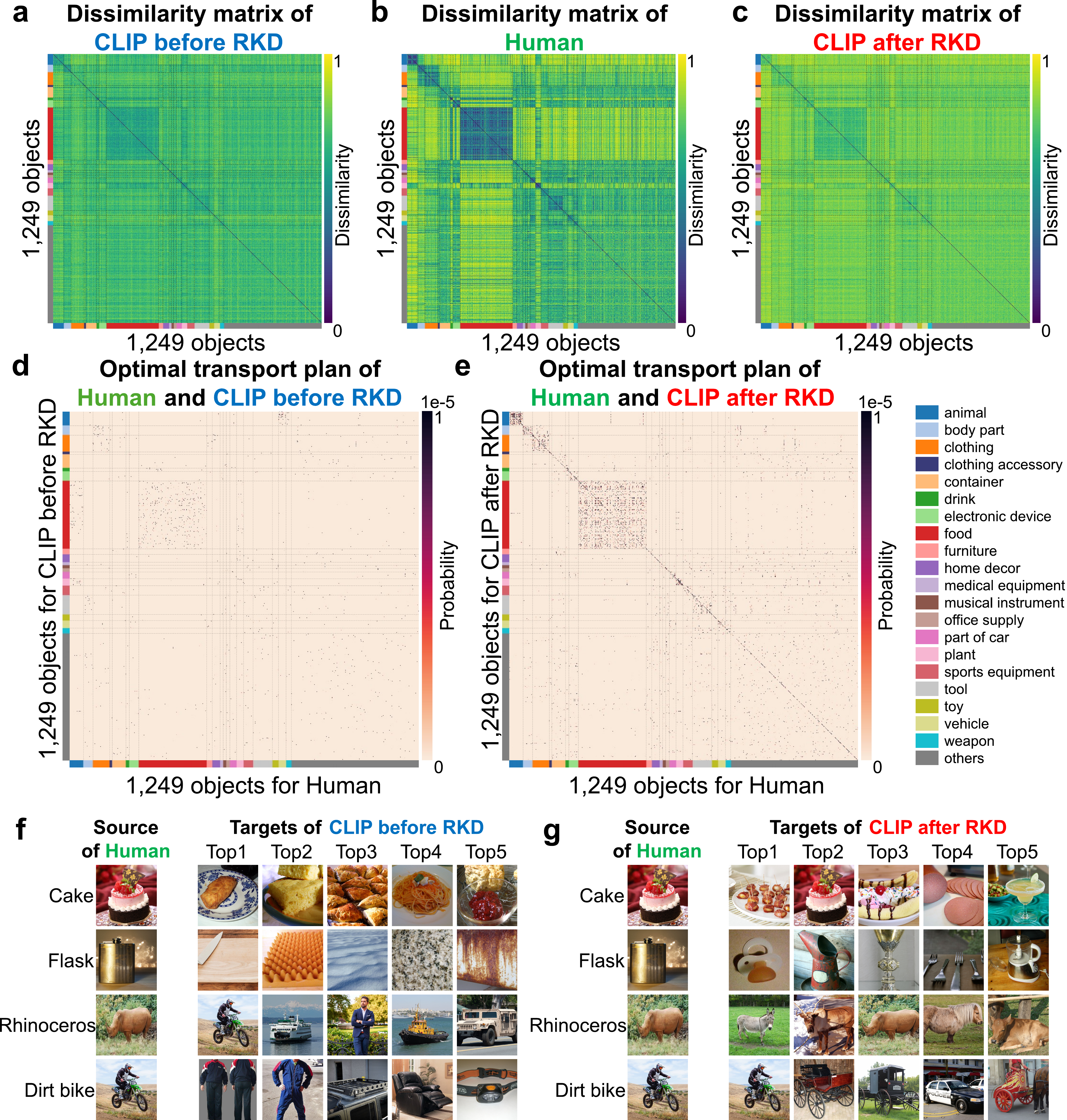}
            \caption{%
                \textbf{Unsupervised comparison between the human and DNN representations before and after RKD, for the CLIP ViT-B/16 (LAION-2B) model on the 1{,}249 THINGS test objects.}
                In (a--e), the 1{,}249 objects follow the same ordering, in which objects are sorted by coarse category. This ordering is annotated on both axes of (a--e), where the color bars indicate coarse-category membership and the grid lines mark the category boundaries (legend at right).
                \textbf{(a--c)} Dissimilarity matrices for CLIP before RKD (a), the human embedding (b), and CLIP after RKD (c).
                \textbf{(d, e)} GWOT transport plans $\Gamma$ between the human and DNN dissimilarity matrices, before RKD (d) and after RKD (e). Rows are human source objects, and columns are DNN target objects.
                \textbf{(f, g)} For four randomly selected human source objects (identical in both panels), the five DNN target objects receiving the largest transport mass in $\Gamma$: before RKD (f) and after RKD (g). In each row, the source object is shown on the left, and the five targets follow in order of decreasing transport mass.
            }
            \label{fig:gwot}
        \end{figure}

            This quantitative shift is also visible in the structure of the GWOT transport plan $\Gamma$. We illustrate this change with the CLIP ViT-B/16 (LAION-2B) model as an example (Fig.~\ref{fig:gwot}d, e; dissimilarity matrices in Fig.~\ref{fig:gwot}a--c). Before RKD, $\Gamma$ exhibited only a faint coarse-category block structure and a weak diagonal line (Fig.~\ref{fig:gwot}d), whereas after RKD, both patterns became visually apparent (Fig.~\ref{fig:gwot}e), corresponding to coarse-category matching and even to individual-object matching, respectively.           

            To illustrate how this improvement manifests in concrete matches between humans and DNNs in the GWOT transport plan, we examined the top-5 DNN target objects receiving the largest transport mass for human source objects. Using the same CLIP ViT-B/16 (LAION-2B) model as an example, we show four randomly selected human source objects, which spanned four distinct coarse categories (Fig.~\ref{fig:gwot}f, g). Overall, the top-5 targets before RKD were often unrelated to the source object, whereas after RKD they tended to be objects of the same kind. As an example, the top-5 targets for ``rhinoceros'' were mostly inanimate objects, such as vehicles, before RKD. After RKD, they were all animals and included ``rhinoceros'' itself.

            To rule out the possibility that the coarse-category-level gain was specific to the human-annotated grouping, we rescored the same GWOT transport plans using a grouping obtained by $k$-means clustering of the test objects in the human psychological embedding into 21 clusters. We found the same pattern: cluster-level top-1 matching accuracy increased after RKD for all five variants and reached a level comparable to that for the human-annotated categories (Supplementary Fig.~\ref{fig:data_driven_clustering}b). This consistency between the human-annotated categories and the data-driven clusters indicates that the reorganization was toward the coarse structure of the human embedding itself, rather than specific to the annotated categories. See Supplementary Section~\ref{sec:supp_data_driven} for details.

            We also confirmed that the gains in RSA correlation and GWOT matching accuracy were not specific to the CLIP ViT-B/16 architecture or to CLIP-style image--text pre-training. Both measures rose after RKD for the DNN models spanning different architectures and pre-training schemes (Supplementary Table~\ref{tab:rsa_gwot_additional}). See Supplementary Section~\ref{sec:supp_additional} for details.

        \subsubsection{Control experiment: a within-dataset split of THINGS showed how strict our generalization test was}
        \label{sec:results_things_split}

            To assess how demanding our primary evaluation (Section~\ref{sec:rsa_results} and Section~\ref{sec:gwot_results}) is as a generalization test, we conducted a control experiment in which we applied RKD and the corresponding evaluation under a within-dataset split of THINGS, sorting the 1{,}854 objects alphabetically and assigning the first 1{,}554 to training and the remaining 300 to testing. This setting removes both of the demands that our primary evaluation imposes, because the teacher and the evaluation reference are now the same THINGS embedding, rather than embeddings derived from different judgment tasks on different image datasets and by different embedding models, and because the test objects are no longer separated from the training concepts by our concept-overlap criterion. We found that, after RKD, the RSA correlation rose from $0.448$--$0.464$ to $0.803$--$0.825$, individual-object top-1 matching accuracy rose from $0.00\%$--$1.33\%$ to $58.0\%$--$65.0\%$ (chance $0.333\%$), and coarse-category top-1 matching accuracy rose from $16.0\%$--$24.7\%$ to $77.7\%$--$84.7\%$ (chance $15.0\%$; Supplementary Table~\ref{tab:rsa_gwot_things_split_ep50}). The two settings use test sets that differ in size (300 versus 1{,}249 objects) and hence in chance level, so we do not compare these values directly with those obtained under our primary evaluation. Even so, every measure was substantially higher under the within-dataset split than under our primary evaluation (Table~\ref{tab:rsa_gwot}), indicating that our primary evaluation is a far stricter generalization test. See Supplementary Section~\ref{sec:supp_split} for details.

    \subsection{The gain in individual-object matching arose from a more human-like global structure, not from more human-like local neighborhoods}
    \label{sec:results_reorganization}

        Having demonstrated that RKD improved GWOT matching at the fine-grained, individual-object level, we next investigated what drove this improvement. In principle, this improvement could originate from a change in the local neighborhood structure, in the global structure, or in both. In this paper, we use the terms local neighborhood structure and global structure as follows. The local neighborhood structure is which other objects lie nearest to each object. The global structure, in contrast, is the overall arrangement of the representation beyond these local neighborhoods. We therefore examined both possibilities on the same 1{,}249-object THINGS test set used for the RSA and GWOT analyses.
            
        \subsubsection{Local structure: human--DNN nearest-neighbor overlap rate remained largely unchanged after RKD}
        \label{sec:knn}
            To examine whether the gain in individual-object matching reflects a more human-like local neighborhood structure in the DNN space, we computed the $k$-nearest-neighbor overlap rate between the human and DNN representations for $k = 5$ and $k = 10$. To illustrate qualitatively what the overlap rate represents, we also visualized the five nearest neighbors of four query objects, the same four human source objects as in the GWOT transport-plan examples (Fig.~\ref{fig:gwot}f, g), in the human, pre-RKD, and post-RKD CLIP ViT-B/16 (LAION-2B) embedding spaces.

        \begin{figure}[t]
            \centering
            \includegraphics[width=\textwidth]{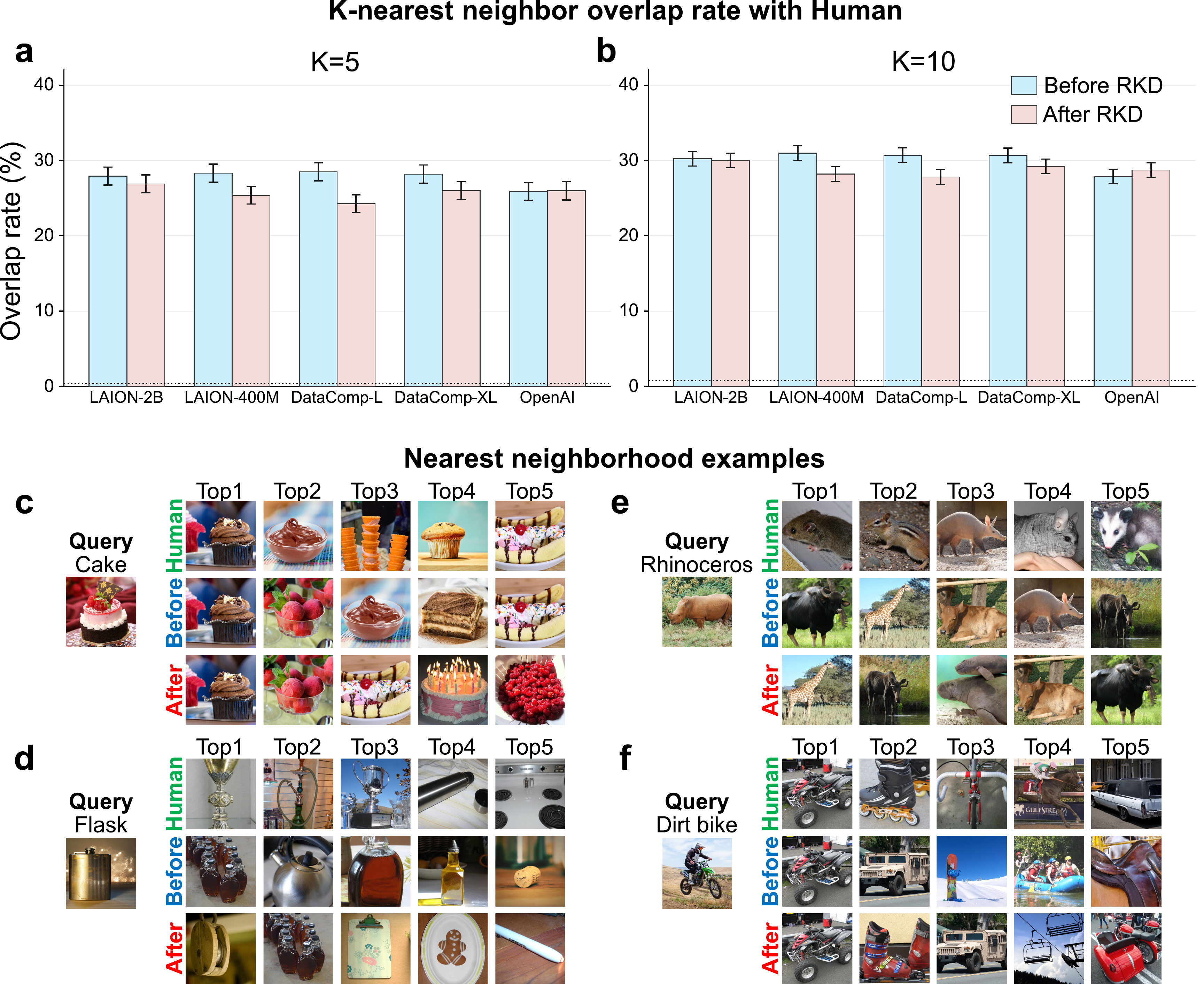}
        \caption{%
            \textbf{$k$-nearest-neighbor overlap rate between the human and DNN representations on the 1{,}249-object THINGS test set, before and after RKD.} \textbf{(a, b)} Mean overlap rates for $k = 5$ and $k = 10$, respectively. In each panel, each pair of bars gives the mean overlap rate for one CLIP ViT-B/16 variant, named on the horizontal axis; the blue bar (Before) shows the rate before RKD and the red bar (After) shows the rate after RKD. Error bars denote 95\% confidence intervals of the mean across the 1{,}249 query objects ($\pm 1.96 \times \mathrm{SEM}$). The dotted horizontal line marks the chance overlap rate ($0.4\%$ for $k = 5$ and $0.8\%$ for $k = 10$). \textbf{(c--f)} Five-nearest-neighbor examples for four randomly selected THINGS query objects, the same four objects as in Fig.~\ref{fig:gwot}f, g. In each panel, the query object is shown on the left. The three rows give the five nearest neighbors retrieved under the human embedding (Human), under CLIP ViT-B/16 (LAION-2B) before RKD (Before), and under the same model after RKD (After). Within each row, neighbors are ordered from left to right by increasing dissimilarity from the query (Top-1 to Top-5).}
            \label{fig:nearest}
        \end{figure}       

            We found that the $k$-nearest-neighbor overlap rate remained largely unchanged after RKD across all five CLIP ViT-B/16 variants and at both values of $k$ (Fig.~\ref{fig:nearest}a, b). Specifically, the change in the mean overlap rate ranged from $-4.2$ to $+0.8$ percentage points across all variant--$k$ combinations, and therefore no variant became substantially more human-like in its local neighborhood structure. Notably, this overlap rate was already far above chance before RKD for all five variants; the $k = 5$ overlap rate ranged from $25.9\%$ to $28.5\%$, against a chance level of $0.4\%$.
            
            We observed the same pattern in the five nearest neighbors of the four query objects, the same four objects as in the GWOT transport-plan examples. As an example, we show these neighbors for CLIP ViT-B/16 (LAION-2B) (Fig.~\ref{fig:nearest}c--f). In the human embedding and in the DNN embedding both before and after RKD, the neighbors were largely objects of a kind similar to the query, and the extent to which the DNN neighbors corresponded to the human neighbors did not change substantially after RKD.
            
            Motivated by the finding that RKD left the human--DNN overlap rate largely unchanged, we additionally measured how much RKD changed the DNN's own local neighborhoods, independently of the human reference. We found that RKD replaced the majority of each object's nearest-neighbor set (Supplementary Table~\ref{tab:knn_before_after}), yet the human--DNN overlap rate remained at the same level. The neighbors that RKD introduced therefore agreed with the human neighborhoods just as well as the neighbors they replaced had. See Supplementary Section~\ref{sec:supp_neighborhoods} for details.
            
            Taken together, these results show that RKD reorganized the DNN's own local neighborhoods but did not increase their agreement with the human neighborhoods. Therefore, the substantial improvement in GWOT matching accuracy reported in Section~\ref{sec:gwot_results} cannot be attributed to a change that made local neighborhoods more human-like; it likely arose from changes in global structure, which we examine in Section~\ref{sec:category}.

        \subsubsection[Global structure: the coarse-category arrangement became more human-like after RKD]
        {Global structure: the coarse-category arrangement became more\protect\newline human-like after RKD}
        \label{sec:category}
            The results reported in Section~\ref{sec:knn} imply that the improvement in fine-grained alignment should be attributed to a more global structural change, as the human--DNN agreement in local neighborhood structure remained largely unchanged. Comprehensively analyzing changes in global relations is generally more difficult than analyzing changes in local relations because it involves more relations. Here, as an example of an analysis revealing global structural changes, we examined the relations among the coarse categories, the 20 human-annotated categories assigned to the test objects.

            \begin{figure}[t]
                \centering
                \includegraphics[width=\textwidth]{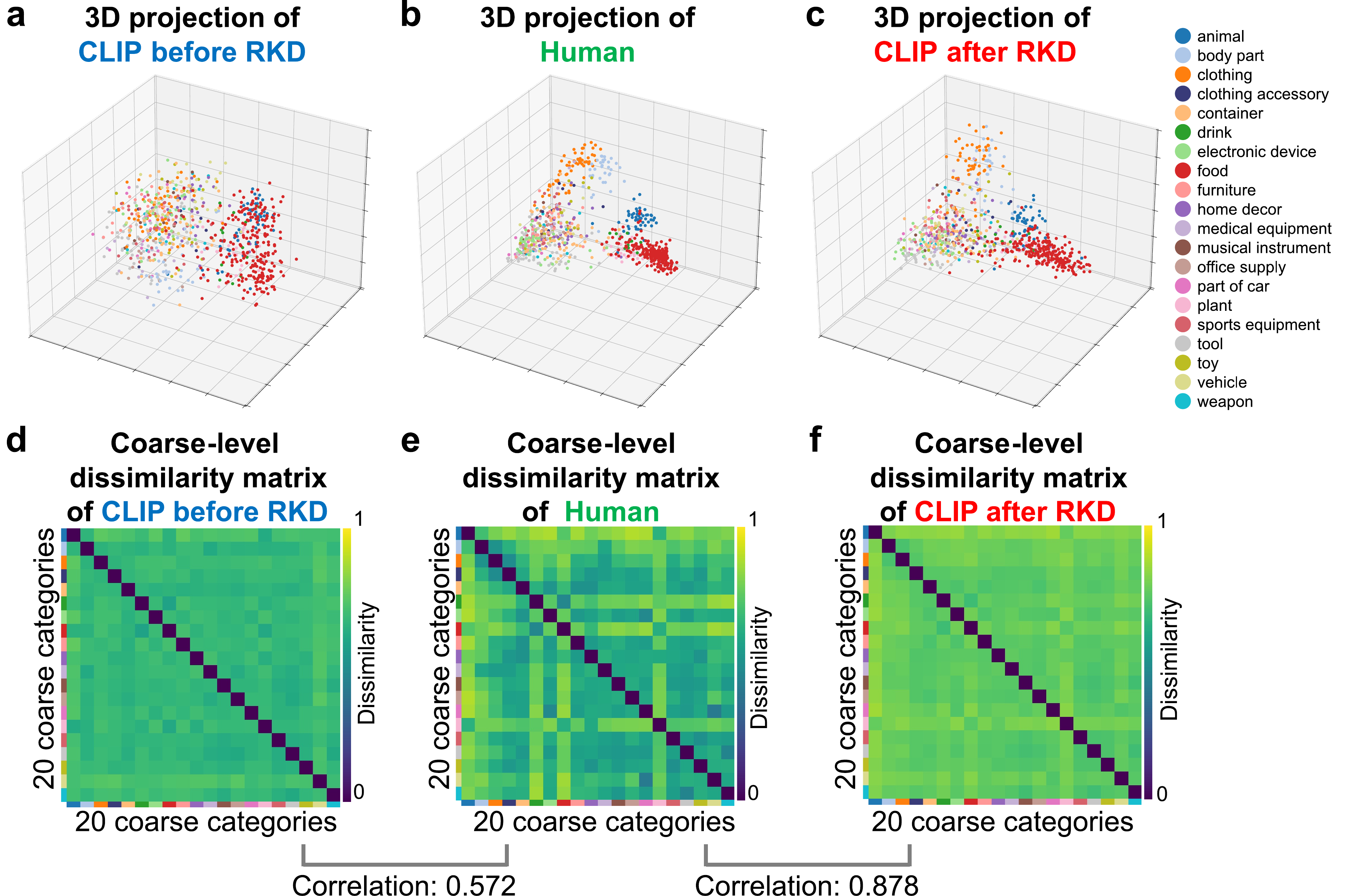}
            \caption{%
                \textbf{Arrangement among the coarse categories in the human and DNN representations before and after RKD, for the CLIP ViT-B/16 (LAION-2B) model.} \textbf{(a--c)}~Three-dimensional projections of the 794 THINGS test objects across 20 coarse categories, excluding the heterogeneous \emph{others} category: (a) CLIP before RKD, (b) human psychological embedding, and (c) CLIP after RKD. The projections were obtained with principal component analysis and placed in a shared coordinate frame by Procrustes alignment. Each object is colored by its coarse category according to the legend at the right of the top row. \textbf{(d--f)}~$20 \times 20$ inter-coarse-category dissimilarity matrices: (d) CLIP before RKD, (e) human psychological embedding, and (f) CLIP after RKD. The color bars along the left and bottom edges indicate the coarse category of each row and column, using the same legend as in (a--c). Brackets below the matrices indicate Spearman correlations between the DNN and human inter-coarse-category dissimilarity matrices before and after RKD.}
                \label{fig:category}
            \end{figure}

            To compare the arrangement of the coarse categories qualitatively, we projected the human, pre-RKD, and post-RKD embeddings into three dimensions and placed them in a shared coordinate frame, taking CLIP ViT-B/16 (LAION-2B) as an example (Fig.~\ref{fig:category}a--c). We found that the post-RKD categories were arranged in much the same way as in the human embedding. Before RKD, in contrast, the categories were largely intermixed. This change concerned the relative placement of the categories, and therefore occurred at the global scale.

            To quantify how much this visual correspondence increased after RKD, we computed the Spearman correlation between the human and CLIP inter-coarse-category dissimilarity matrices (Fig.~\ref{fig:category}d--f). This correlation captures whether the same category pairs are relatively near or far in the two embeddings. We found that it increased after RKD for all five variants, from a pre-RKD range of $\rho = 0.572$--$0.751$ to a post-RKD range of $\rho = 0.838$--$0.879$. As an example, we show the inter-coarse-category dissimilarity matrices for CLIP ViT-B/16 (LAION-2B) (Fig.~\ref{fig:category}d--f). For this model, the correlation rose from $\rho = 0.572$ to $\rho = 0.878$.

            To confirm that these findings did not depend on the human-annotated categories, we repeated both analyses on the data-driven clusters derived from the human embedding, as we did in the GWOT coarse-grained matching analysis. The arrangement again became more human-like after RKD in the three-dimensional projections (Supplementary Fig.~\ref{fig:data_driven_clustering}c--e), and the correlation increased for all five variants (Supplementary Fig.~\ref{fig:data_driven_clustering}f--h). These results therefore did not depend on the human-annotated categories. See Supplementary Section~\ref{sec:supp_data_driven} for details.

            Taken together with the local-structure analysis (Section~\ref{sec:knn}), these results indicate that the gain in individual-object matching reflects a reorganization of the global structure toward the human structure, rather than a change in the human-likeness of the local neighborhood structure.

\FloatBarrier
\section{Discussion}
\label{Discussion}

    To test whether relational transfer from humans to DNNs holds under a strict evaluation, we fine-tuned DNNs with Relational Knowledge Distillation (RKD), and evaluated them against an independently obtained human psychological embedding on object concepts curated not to overlap with the concepts used for the transfer. As we expected, under supervised comparison, the RSA correlation increased after RKD (Table~\ref{tab:rsa_gwot}). Under unsupervised comparison with Gromov--Wasserstein optimal transport (GWOT), matching accuracy increased not only at the coarse-category level but also at the fine-grained, individual-object level (Fig.~\ref{fig:gwot}d, e; Table~\ref{tab:rsa_gwot}). To identify which aspect of the representational structure gave rise to this fine-grained alignment, we characterized the local and the global structure before and after RKD. The human--DNN nearest-neighbor overlap rate, which reflects the local structure, remained largely unchanged after RKD (Fig.~\ref{fig:nearest}a, b). The local structure therefore did not contribute to the gain in fine-grained matching. In contrast, the arrangement among the coarse categories, one aspect of the global structure, became more human-like after RKD (Fig.~\ref{fig:category}).
    
    \subsection{Unsupervised comparison is necessary for evaluating relational transfer at the fine-grained, individual-object level}
    \label{sec:disc_unsupervised}

        We argue that an unsupervised comparison method, GWOT, is necessary for evaluating human--DNN alignment at the individual-object level. Before RKD, the RSA correlation was already moderately high, at $\rho = 0.407$--$0.445$. After RKD, it increased by $0.10$--$0.16$, reaching $\rho = 0.527$--$0.568$ (Table~\ref{tab:rsa_gwot}). However, this increase alone cannot reveal whether the alignment improved at the coarse-category level or at the individual-object level. In contrast, GWOT matching accuracy at the individual-object level distinguishes fine-grained alignment from alignment that is only coarse-grained. This accuracy rose from $0.560\%$--$2.96\%$ before RKD to $13.1\%$--$19.5\%$ after RKD (Table~\ref{tab:rsa_gwot}). Accordingly, the transport plan showed a visible identity diagonal only after RKD (Fig.~\ref{fig:gwot}d, e). We should also mention CLIP ResNet-50, one of the additional models we evaluated, as a particularly clear case in which the two measures did not move together. Its RSA correlation barely changed after RKD ($\rho = 0.500$ to $\rho = 0.501$), whereas its individual-object matching accuracy rose from $0.400\%$ to $10.3\%$ (Supplementary Table~\ref{tab:rsa_gwot_additional}). Supervised comparison alone would therefore have reported no gain for this model, even though its representation had in fact been reorganized toward the human structure.

    \subsection{Relational transfer conveys a general organization of object representations, not adjustments specific to the trained concepts}
    \label{sec:disc_generalization}

        We demonstrate that relational transfer conveyed a general organization of object representations, rather than adjustments specific to the concepts used for the transfer. Our primary evaluation was demanding in two respects (Fig.~\ref{fig:pipeline}). First, the test concepts were curated to exclude hierarchical relations in WordNet to the training concepts. Second, the evaluation embedding was obtained independently of the one we transferred, from a different judgment task on a different image dataset, and by a different embedding model. Alignment without supervision under both demands was a strict requirement: unsupervised comparison provides no stimulus pairing, so each individual object had to be matched from the internal distance structure alone. Indeed, prior studies that used supervised comparison have reported that alignment can fall below its pre-adaptation level when the evaluation data differ from the data used for fitting \citep{Peterson2018-hb, Fu2023-ds}. To gauge how much these two demands cost, we repeated the RKD and evaluation procedure under a within-dataset training--test split of THINGS, in which neither demand was present. Every measure was substantially higher in this control experiment (Section~\ref{sec:results_things_split}). This gap confirms how difficult the required generalization is. Even so, individual objects remained matchable under our primary evaluation. We therefore conclude that what RKD conveyed was the organization of the human space itself, and not just the coordinates of the concepts on which it was trained.

    \subsection{Relational transfer from humans supplies the human-like global structure that does not emerge spontaneously from pre-training}
    \label{sec:disc_global}

        We interpret the human--DNN representational similarity observed before relational transfer through the lens of two hypotheses about representational convergence among DNNs. One is the Platonic Representation Hypothesis, which holds that independently trained DNNs converge toward a shared representation \citep{Huh2024-wv}. The other is the Aristotelian Representation Hypothesis, which refines this claim: after correcting for the confounding effects of model width and depth, the apparent agreement in global geometry largely disappears, and only local neighborhood relations remain shared across models \citep{Groger2026-pc}. We asked the same question with the human representation in place of one of the two DNNs, using at the local level the same nearest-neighbor overlap measure as these studies used (Section~\ref{sec:methods_knn}). Our results fit the Aristotelian view. Before relational transfer, the human--DNN agreement in local neighborhood structure was already far above chance, at $25.9\%$--$28.5\%$ for $k = 5$ against a chance level of $0.4\%$ (Fig.~\ref{fig:nearest}a). In contrast, the arrangement among the coarse categories, one aspect of the global structure, was only partly human-like, at $\rho = 0.572$--$0.751$ (Fig.~\ref{fig:category}d, e). The correspondence that GWOT recovered at the coarse-category level was likewise modest, at $34.8\%$--$40.6\%$ against a chance level of $18.4\%$ (Table~\ref{tab:rsa_gwot}).

        We therefore suggest that relational transfer from humans supplied the human-like global structure that did not emerge spontaneously from pre-training. After the transfer, this global arrangement of the coarse categories became more human-like ($\rho = 0.838$--$0.879$; Fig.~\ref{fig:category}e, f). The coarse-category-level GWOT matching accuracy also increased, reaching $57.6\%$--$61.2\%$ (Table~\ref{tab:rsa_gwot}). The local neighborhood overlap rate, in contrast, remained largely unchanged (Fig.~\ref{fig:nearest}a, b). This dissociation held for all five CLIP ViT-B/16 variants, which differ in the scale and the curation of their pre-training data. Pre-training alone did not bring the global structure close enough to the human structure for individual objects to be matched without supervision. 

    \subsection{Future work: toward transferring neural and person-specific representations}
    \label{sec:future_work}

        Although this study transfers human psychological embeddings estimated from behavioral data, one direction would be to transfer human neural activity data to DNNs. If we interpret neural activity as directly corresponding to embeddings in DNNs, then transferring the relations among neural activity patterns can be regarded as a more direct method of transferring human-like representations. Indeed, neural measurements have been reported to capture aspects of human mental representation that behavioral judgments do not capture \citep{Bauer2019-cp, King2019-jw}. Because the two signals are complementary in this way, transferring both in a single objective would convey relational structure that neither signal carries alone. Our current pipeline accommodates this extension directly. RKD uses only the pairwise distances and the triplet-wise angles among data points, so switching from behavioral to neural data, or using both at once, requires no change to the loss function or to the DNN architecture. We chose RKD as the method of relational transfer with this extensibility in mind.

        A second direction is to move from the population-level human representation transferred in this study to an individual-level one. Such a transfer would enable the resulting DNN to serve as a proxy for a specific person rather than for people in general. The human embedding we transferred was estimated from judgments pooled across thousands of participants, and therefore captures the average structure shared across people. Because these judgments were distributed across so many participants, each participant contributed a small share of them. Transferring an individual's own embedding would instead convey that person's idiosyncratic structure. However, the judgments could then no longer be distributed, because every one of them would have to come from the same person. In that case, the amount of data required from that person would be the main obstacle. Collecting a full individual embedding from behavioral judgments alone would therefore be impractical. Neural measurements complement behavioral judgments, so incorporating them offers one possible solution.

\clearpage

\section*{CRediT authorship contribution statement}

\textbf{Yuria Shimizu:} Writing -- review \& editing, Writing -- original draft, Visualization, Validation, Software, Methodology, Investigation, Formal analysis, Data curation, Conceptualization; \textbf{Soh Takahashi:} Writing -- review \& editing, Methodology; \textbf{Takato Horii:} Writing -- review \& editing, Methodology; \textbf{Masafumi Oizumi:} Writing -- review \& editing, Writing -- original draft, Supervision, Resources, Project administration, Methodology, Funding acquisition, Conceptualization.

\section*{Declaration of competing interest}

The authors declare no competing interests.

\section*{Funding}

This work was supported by JST Moonshot R\&D [grant number JPMJMS2012] and
the Japan Society for the Promotion of Science, Grant-in-Aid for
Transformative Research Areas [grant number 23H04834].

\section*{Declaration of generative AI and AI-assisted technologies in the manuscript preparation process}

During the preparation of this work, the authors used Claude (Anthropic) in order to check grammar and improve the clarity and fluency of the English text. After using this tool, the authors reviewed and edited the content as needed and take full responsibility for the content of the published article.

\clearpage
\bibliographystyle{apa7}
\bibliography{references}

\begin{thebibliography}{}

\bibitem [\protect \citeauthoryear {%
Akiba%
\ \protect \BOthers {.}}{%
Akiba%
\ \protect \BOthers {.}}{%
{\protect \APACyear {2019}}%
}]{%
Akiba2019-yo}
\APACinsertmetastar {%
Akiba2019-yo}%
\begin{APACrefauthors}%
Akiba, T.%
, Sano, S.%
, Yanase, T.%
, Ohta, T.%
\BCBL {}\ \BBA {} Koyama, M.%
\end{APACrefauthors}%
\unskip\
\newblock
\APACrefYearMonthDay{2019}{}{}.
\newblock
{\BBOQ}\APACrefatitle {{Optuna}: A next-generation hyperparameter optimization
  framework} {{Optuna}: A next-generation hyperparameter optimization
  framework}.{\BBCQ}
\newblock
\BIn{} \APACrefbtitle {{Proceedings of the 25th ACM SIGKDD International
  Conference on Knowledge Discovery \& Data Mining}} {{Proceedings of the 25th
  ACM SIGKDD International Conference on Knowledge Discovery \& Data Mining}}\
  (\BPGS\ 2623--2631).
\newblock
\APACaddressPublisher{}{{Association for Computing Machinery}}.
\newblock
\begin{APACrefDOI} \doi{10.1145/3292500.3330701} \end{APACrefDOI}
\PrintBackRefs{\CurrentBib}

\bibitem [\protect \citeauthoryear {%
Attarian%
\ \protect \BOthers {.}}{%
Attarian%
\ \protect \BOthers {.}}{%
{\protect \APACyear {2020}}%
}]{%
Attarian2020-bo}
\APACinsertmetastar {%
Attarian2020-bo}%
\begin{APACrefauthors}%
Attarian, M.%
, Roads, B\BPBI D.%
\BCBL {}\ \BBA {} Mozer, M\BPBI C.%
\end{APACrefauthors}%
\unskip\
\newblock
\APACrefYearMonthDay{2020}{}{}.
\newblock
\APACrefbtitle {Transforming neural network visual representations to predict
  human judgments of similarity} {Transforming neural network visual
  representations to predict human judgments of similarity}\ [{Preprint}].
\newblock
\APAChowpublished {ar{X}iv}.
\newblock
\begin{APACrefDOI} \doi{10.48550/arXiv.2010.06512} \end{APACrefDOI}
\PrintBackRefs{\CurrentBib}

\bibitem [\protect \citeauthoryear {%
Bauer%
\ \BBA {} Just%
}{%
Bauer%
\ \BBA {} Just%
}{%
{\protect \APACyear {2019}}%
}]{%
Bauer2019-cp}
\APACinsertmetastar {%
Bauer2019-cp}%
\begin{APACrefauthors}%
Bauer, A\BPBI J.%
\BCBT {}\ \BBA {} Just, M\BPBI A.%
\end{APACrefauthors}%
\unskip\
\newblock
\APACrefYearMonthDay{2019}{}{}.
\newblock
{\BBOQ}\APACrefatitle {Brain reading and behavioral methods provide
  complementary perspectives on the representation of concepts} {Brain reading
  and behavioral methods provide complementary perspectives on the
  representation of concepts}.{\BBCQ}
\newblock
\APACjournalVolNumPages{NeuroImage}{186}{}{794--805}.
\newblock
\begin{APACrefDOI} \doi{10.1016/j.neuroimage.2018.11.022} \end{APACrefDOI}
\PrintBackRefs{\CurrentBib}

\bibitem [\protect \citeauthoryear {%
Bergstra%
\ \protect \BOthers {.}}{%
Bergstra%
\ \protect \BOthers {.}}{%
{\protect \APACyear {2011}}%
}]{%
Bergstra2011-mr}
\APACinsertmetastar {%
Bergstra2011-mr}%
\begin{APACrefauthors}%
Bergstra, J.%
, Bardenet, R.%
, Bengio, Y.%
\BCBL {}\ \BBA {} K{\'e}gl, B.%
\end{APACrefauthors}%
\unskip\
\newblock
\APACrefYearMonthDay{2011}{}{}.
\newblock
{\BBOQ}\APACrefatitle {Algorithms for hyper-parameter optimization} {Algorithms
  for hyper-parameter optimization}.{\BBCQ}
\newblock
\BIn{} \APACrefbtitle {{Advances in Neural Information Processing Systems}}
  {{Advances in Neural Information Processing Systems}}\ (\BVOL~24, \BPGS\
  2546--2554).
\newblock
\APACaddressPublisher{}{{Curran Associates}}.
\PrintBackRefs{\CurrentBib}

\bibitem [\protect \citeauthoryear {%
Bird%
\ \protect \BOthers {.}}{%
Bird%
\ \protect \BOthers {.}}{%
{\protect \APACyear {2009}}%
}]{%
Bird2009-nl}
\APACinsertmetastar {%
Bird2009-nl}%
\begin{APACrefauthors}%
Bird, S.%
, Loper, E.%
\BCBL {}\ \BBA {} Klein, E.%
\end{APACrefauthors}%
\unskip\
\newblock
\APACrefYear{2009}.
\newblock
\APACrefbtitle {Natural language processing with {Python}: Analyzing text with
  the {Natural} {Language} {Toolkit}} {Natural language processing with
  {Python}: Analyzing text with the {Natural} {Language} {Toolkit}}.
\newblock
\APACaddressPublisher{}{{O'Reilly Media}}.
\PrintBackRefs{\CurrentBib}

\bibitem [\protect \citeauthoryear {%
Caron%
\ \protect \BOthers {.}}{%
Caron%
\ \protect \BOthers {.}}{%
{\protect \APACyear {2020}}%
}]{%
Caron2020-um}
\APACinsertmetastar {%
Caron2020-um}%
\begin{APACrefauthors}%
Caron, M.%
, Misra, I.%
, Mairal, J.%
, Goyal, P.%
, Bojanowski, P.%
\BCBL {}\ \BBA {} Joulin, A.%
\end{APACrefauthors}%
\unskip\
\newblock
\APACrefYearMonthDay{2020}{}{}.
\newblock
{\BBOQ}\APACrefatitle {Unsupervised learning of visual features by contrasting
  cluster assignments} {Unsupervised learning of visual features by contrasting
  cluster assignments}.{\BBCQ}
\newblock
\BIn{} \APACrefbtitle {{Advances in Neural Information Processing Systems}}
  {{Advances in Neural Information Processing Systems}}\ (\BVOL~33, \BPGS\
  9912--9924).
\newblock
\APACaddressPublisher{}{{Curran Associates}}.
\PrintBackRefs{\CurrentBib}

\bibitem [\protect \citeauthoryear {%
Chen%
\ \protect \BOthers {.}}{%
Chen%
\ \protect \BOthers {.}}{%
{\protect \APACyear {2020}}%
}]{%
Chen2020-rt}
\APACinsertmetastar {%
Chen2020-rt}%
\begin{APACrefauthors}%
Chen, T.%
, Kornblith, S.%
, Norouzi, M.%
\BCBL {}\ \BBA {} Hinton, G.%
\end{APACrefauthors}%
\unskip\
\newblock
\APACrefYearMonthDay{2020}{}{}.
\newblock
{\BBOQ}\APACrefatitle {A simple framework for contrastive learning of visual
  representations} {A simple framework for contrastive learning of visual
  representations}.{\BBCQ}
\newblock
\BIn{} \APACrefbtitle {{Proceedings of the 37th International Conference on
  Machine Learning}} {{Proceedings of the 37th International Conference on
  Machine Learning}}\ (\BVOL~119, \BPGS\ 1597--1607).
\newblock
\APACaddressPublisher{}{{PMLR}}.
\PrintBackRefs{\CurrentBib}

\bibitem [\protect \citeauthoryear {%
Cherti%
\ \protect \BOthers {.}}{%
Cherti%
\ \protect \BOthers {.}}{%
{\protect \APACyear {2023}}%
}]{%
Cherti2022-fl}
\APACinsertmetastar {%
Cherti2022-fl}%
\begin{APACrefauthors}%
Cherti, M.%
, Beaumont, R.%
, Wightman, R.%
, Wortsman, M.%
, Ilharco, G.%
, Gordon, C.%
, Schuhmann, C.%
, Schmidt, L.%
\BCBL {}\ \BBA {} Jitsev, J.%
\end{APACrefauthors}%
\unskip\
\newblock
\APACrefYearMonthDay{2023}{}{}.
\newblock
{\BBOQ}\APACrefatitle {Reproducible scaling laws for contrastive language-image
  learning} {Reproducible scaling laws for contrastive language-image
  learning}.{\BBCQ}
\newblock
\BIn{} \APACrefbtitle {{2023 IEEE/CVF Conference on Computer Vision and Pattern
  Recognition}} {{2023 IEEE/CVF Conference on Computer Vision and Pattern
  Recognition}}\ (\BPGS\ 2818--2829).
\newblock
\APACaddressPublisher{}{{IEEE}}.
\newblock
\begin{APACrefDOI} \doi{10.1109/CVPR52729.2023.00276} \end{APACrefDOI}
\PrintBackRefs{\CurrentBib}

\bibitem [\protect \citeauthoryear {%
Cichy%
\ \protect \BOthers {.}}{%
Cichy%
\ \protect \BOthers {.}}{%
{\protect \APACyear {2016}}%
}]{%
Cichy2016-dz}
\APACinsertmetastar {%
Cichy2016-dz}%
\begin{APACrefauthors}%
Cichy, R\BPBI M.%
, Khosla, A.%
, Pantazis, D.%
, Torralba, A.%
\BCBL {}\ \BBA {} Oliva, A.%
\end{APACrefauthors}%
\unskip\
\newblock
\APACrefYearMonthDay{2016}{}{}.
\newblock
{\BBOQ}\APACrefatitle {Comparison of deep neural networks to spatio-temporal
  cortical dynamics of human visual object recognition reveals hierarchical
  correspondence} {Comparison of deep neural networks to spatio-temporal
  cortical dynamics of human visual object recognition reveals hierarchical
  correspondence}.{\BBCQ}
\newblock
\APACjournalVolNumPages{Scientific Reports}{6}{}{Article 27755}.
\newblock
\begin{APACrefDOI} \doi{10.1038/srep27755} \end{APACrefDOI}
\PrintBackRefs{\CurrentBib}

\bibitem [\protect \citeauthoryear {%
Conwell%
\ \protect \BOthers {.}}{%
Conwell%
\ \protect \BOthers {.}}{%
{\protect \APACyear {2024}}%
}]{%
Conwell2024-ls}
\APACinsertmetastar {%
Conwell2024-ls}%
\begin{APACrefauthors}%
Conwell, C.%
, Prince, J\BPBI S.%
, Kay, K\BPBI N.%
, Alvarez, G\BPBI A.%
\BCBL {}\ \BBA {} Konkle, T.%
\end{APACrefauthors}%
\unskip\
\newblock
\APACrefYearMonthDay{2024}{}{}.
\newblock
{\BBOQ}\APACrefatitle {A large-scale examination of inductive biases shaping
  high-level visual representation in brains and machines} {A large-scale
  examination of inductive biases shaping high-level visual representation in
  brains and machines}.{\BBCQ}
\newblock
\APACjournalVolNumPages{Nature Communications}{15}{}{Article 9383}.
\newblock
\begin{APACrefDOI} \doi{10.1038/s41467-024-53147-y} \end{APACrefDOI}
\PrintBackRefs{\CurrentBib}

\bibitem [\protect \citeauthoryear {%
Cuturi%
}{%
Cuturi%
}{%
{\protect \APACyear {2013}}%
}]{%
Cuturi2013-sk}
\APACinsertmetastar {%
Cuturi2013-sk}%
\begin{APACrefauthors}%
Cuturi, M.%
\end{APACrefauthors}%
\unskip\
\newblock
\APACrefYearMonthDay{2013}{}{}.
\newblock
{\BBOQ}\APACrefatitle {{Sinkhorn} distances: Lightspeed computation of optimal
  transport} {{Sinkhorn} distances: Lightspeed computation of optimal
  transport}.{\BBCQ}
\newblock
\BIn{} \APACrefbtitle {{Advances in Neural Information Processing Systems}}
  {{Advances in Neural Information Processing Systems}}\ (\BVOL~26, \BPGS\
  2292--2300).
\newblock
\APACaddressPublisher{}{{Curran Associates}}.
\PrintBackRefs{\CurrentBib}

\bibitem [\protect \citeauthoryear {%
Deng%
\ \protect \BOthers {.}}{%
Deng%
\ \protect \BOthers {.}}{%
{\protect \APACyear {2009}}%
}]{%
Deng2009-ea}
\APACinsertmetastar {%
Deng2009-ea}%
\begin{APACrefauthors}%
Deng, J.%
, Dong, W.%
, Socher, R.%
, Li, L\BHBI J.%
, Li, K.%
\BCBL {}\ \BBA {} Fei-Fei, L.%
\end{APACrefauthors}%
\unskip\
\newblock
\APACrefYearMonthDay{2009}{}{}.
\newblock
{\BBOQ}\APACrefatitle {{ImageNet}: A large-scale hierarchical image database}
  {{ImageNet}: A large-scale hierarchical image database}.{\BBCQ}
\newblock
\BIn{} \APACrefbtitle {{2009 IEEE Conference on Computer Vision and Pattern
  Recognition}} {{2009 IEEE Conference on Computer Vision and Pattern
  Recognition}}\ (\BPGS\ 248--255).
\newblock
\APACaddressPublisher{}{{IEEE}}.
\newblock
\begin{APACrefDOI} \doi{10.1109/CVPR.2009.5206848} \end{APACrefDOI}
\PrintBackRefs{\CurrentBib}

\bibitem [\protect \citeauthoryear {%
Dosovitskiy%
\ \protect \BOthers {.}}{%
Dosovitskiy%
\ \protect \BOthers {.}}{%
{\protect \APACyear {2021}}%
}]{%
Dosovitskiy2020-ay}
\APACinsertmetastar {%
Dosovitskiy2020-ay}%
\begin{APACrefauthors}%
Dosovitskiy, A.%
, Beyer, L.%
, Kolesnikov, A.%
, Weissenborn, D.%
, Zhai, X.%
, Unterthiner, T.%
, Dehghani, M.%
, Minderer, M.%
, Heigold, G.%
, Gelly, S.%
, Uszkoreit, J.%
\BCBL {}\ \BBA {} Houlsby, N.%
\end{APACrefauthors}%
\unskip\
\newblock
\APACrefYearMonthDay{2021}{}{}.
\newblock
{\BBOQ}\APACrefatitle {An image is worth 16x16 words: Transformers for image
  recognition at scale} {An image is worth 16x16 words: Transformers for image
  recognition at scale}.{\BBCQ}
\newblock
\BIn{} \APACrefbtitle {{International Conference on Learning Representations}.}
  {{International Conference on Learning Representations}.}
\PrintBackRefs{\CurrentBib}

\bibitem [\protect \citeauthoryear {%
Flamary%
\ \protect \BOthers {.}}{%
Flamary%
\ \protect \BOthers {.}}{%
{\protect \APACyear {2021}}%
}]{%
Flamary2021-mg}
\APACinsertmetastar {%
Flamary2021-mg}%
\begin{APACrefauthors}%
Flamary, R.%
, Courty, N.%
, Gramfort, A.%
, Alaya, M\BPBI Z.%
, Boisbunon, A.%
, Chambon, S.%
, Chapel, L.%
, Corenflos, A.%
, Fatras, K.%
, Fournier, N.%
, Gautheron, L.%
, Gayraud, N\BPBI T\BPBI H.%
, Janati, H.%
, Rakotomamonjy, A.%
, Redko, I.%
, Rolet, A.%
, Schutz, A.%
, Seguy, V.%
, Sutherland, D\BPBI J.%
\BDBL {}Vayer, T.%
\end{APACrefauthors}%
\unskip\
\newblock
\APACrefYearMonthDay{2021}{}{}.
\newblock
{\BBOQ}\APACrefatitle {{POT}: {Python} optimal transport} {{POT}: {Python}
  optimal transport}.{\BBCQ}
\newblock
\APACjournalVolNumPages{Journal of Machine Learning Research}{22}{78}{1--8}.
\PrintBackRefs{\CurrentBib}

\bibitem [\protect \citeauthoryear {%
Fu%
\ \protect \BOthers {.}}{%
Fu%
\ \protect \BOthers {.}}{%
{\protect \APACyear {2023}}%
}]{%
Fu2023-ds}
\APACinsertmetastar {%
Fu2023-ds}%
\begin{APACrefauthors}%
Fu, S.%
, Tamir, N\BPBI Y.%
, Sundaram, S.%
, Chai, L.%
, Zhang, R.%
, Dekel, T.%
\BCBL {}\ \BBA {} Isola, P.%
\end{APACrefauthors}%
\unskip\
\newblock
\APACrefYearMonthDay{2023}{}{}.
\newblock
{\BBOQ}\APACrefatitle {{DreamSim}: Learning new dimensions of human visual
  similarity using synthetic data} {{DreamSim}: Learning new dimensions of
  human visual similarity using synthetic data}.{\BBCQ}
\newblock
\BIn{} \APACrefbtitle {{Advances in Neural Information Processing Systems}}
  {{Advances in Neural Information Processing Systems}}\ (\BVOL~36, \BPGS\
  50742--50768).
\newblock
\APACaddressPublisher{}{{Curran Associates}}.
\newblock
\begin{APACrefDOI} \doi{10.52202/075280-2208} \end{APACrefDOI}
\PrintBackRefs{\CurrentBib}

\bibitem [\protect \citeauthoryear {%
Gadre%
\ \protect \BOthers {.}}{%
Gadre%
\ \protect \BOthers {.}}{%
{\protect \APACyear {2023}}%
}]{%
Gadre2023-dr}
\APACinsertmetastar {%
Gadre2023-dr}%
\begin{APACrefauthors}%
Gadre, S\BPBI Y.%
, Ilharco, G.%
, Fang, A.%
, Hayase, J.%
, Smyrnis, G.%
, Nguyen, T.%
, Marten, R.%
, Wortsman, M.%
, Ghosh, D.%
, Zhang, J.%
, Orgad, E.%
, Entezari, R.%
, Daras, G.%
, Pratt, S.%
, Ramanujan, V.%
, Bitton, Y.%
, Marathe, K.%
, Mussmann, S.%
, Vencu, R.%
\BDBL {}Schmidt, L.%
\end{APACrefauthors}%
\unskip\
\newblock
\APACrefYearMonthDay{2023}{}{}.
\newblock
{\BBOQ}\APACrefatitle {{DataComp}: In search of the next generation of
  multimodal datasets} {{DataComp}: In search of the next generation of
  multimodal datasets}.{\BBCQ}
\newblock
\BIn{} \APACrefbtitle {{Advances in Neural Information Processing Systems}}
  {{Advances in Neural Information Processing Systems}}\ (\BVOL~36, \BPGS\
  27092--27112).
\newblock
\APACaddressPublisher{}{{Curran Associates}}.
\newblock
\begin{APACrefDOI} \doi{10.52202/075280-1179} \end{APACrefDOI}
\PrintBackRefs{\CurrentBib}

\bibitem [\protect \citeauthoryear {%
Gidaris%
\ \protect \BOthers {.}}{%
Gidaris%
\ \protect \BOthers {.}}{%
{\protect \APACyear {2018}}%
}]{%
Gidaris2018-qs}
\APACinsertmetastar {%
Gidaris2018-qs}%
\begin{APACrefauthors}%
Gidaris, S.%
, Singh, P.%
\BCBL {}\ \BBA {} Komodakis, N.%
\end{APACrefauthors}%
\unskip\
\newblock
\APACrefYearMonthDay{2018}{}{}.
\newblock
{\BBOQ}\APACrefatitle {Unsupervised representation learning by predicting image
  rotations} {Unsupervised representation learning by predicting image
  rotations}.{\BBCQ}
\newblock
\BIn{} \APACrefbtitle {{International Conference on Learning Representations}.}
  {{International Conference on Learning Representations}.}
\PrintBackRefs{\CurrentBib}

\bibitem [\protect \citeauthoryear {%
Gr{\"o}ger%
\ \protect \BOthers {.}}{%
Gr{\"o}ger%
\ \protect \BOthers {.}}{%
{\protect \APACyear {2026}}%
}]{%
Groger2026-pc}
\APACinsertmetastar {%
Groger2026-pc}%
\begin{APACrefauthors}%
Gr{\"o}ger, F.%
, Wen, S.%
\BCBL {}\ \BBA {} Brbi{\'c}, M.%
\end{APACrefauthors}%
\unskip\
\newblock
\APACrefYearMonthDay{2026}{}{}.
\newblock
\APACrefbtitle {Revisiting the {Platonic} representation hypothesis: An
  {Aristotelian} view} {Revisiting the {Platonic} representation hypothesis: An
  {Aristotelian} view}\ [{Preprint}].
\newblock
\APAChowpublished {ar{X}iv}.
\newblock
\APACrefnote{{Accepted at the 43rd International Conference on Machine
  Learning}}
\newblock
\begin{APACrefDOI} \doi{10.48550/arXiv.2602.14486} \end{APACrefDOI}
\PrintBackRefs{\CurrentBib}

\bibitem [\protect \citeauthoryear {%
G{\"u}{\c c}l{\"u}%
\ \BBA {} van Gerven%
}{%
G{\"u}{\c c}l{\"u}%
\ \BBA {} van Gerven%
}{%
{\protect \APACyear {2015}}%
}]{%
Guclu2015-jk}
\APACinsertmetastar {%
Guclu2015-jk}%
\begin{APACrefauthors}%
G{\"u}{\c c}l{\"u}, U.%
\BCBT {}\ \BBA {} van Gerven, M\BPBI A\BPBI J.%
\end{APACrefauthors}%
\unskip\
\newblock
\APACrefYearMonthDay{2015}{}{}.
\newblock
{\BBOQ}\APACrefatitle {Deep neural networks reveal a gradient in the complexity
  of neural representations across the ventral stream} {Deep neural networks
  reveal a gradient in the complexity of neural representations across the
  ventral stream}.{\BBCQ}
\newblock
\APACjournalVolNumPages{The Journal of Neuroscience}{35}{27}{10005--10014}.
\newblock
\begin{APACrefDOI} \doi{10.1523/JNEUROSCI.5023-14.2015} \end{APACrefDOI}
\PrintBackRefs{\CurrentBib}

\bibitem [\protect \citeauthoryear {%
He%
\ \protect \BOthers {.}}{%
He%
\ \protect \BOthers {.}}{%
{\protect \APACyear {2016}}%
}]{%
He2016-je}
\APACinsertmetastar {%
He2016-je}%
\begin{APACrefauthors}%
He, K.%
, Zhang, X.%
, Ren, S.%
\BCBL {}\ \BBA {} Sun, J.%
\end{APACrefauthors}%
\unskip\
\newblock
\APACrefYearMonthDay{2016}{}{}.
\newblock
{\BBOQ}\APACrefatitle {Deep residual learning for image recognition} {Deep
  residual learning for image recognition}.{\BBCQ}
\newblock
\BIn{} \APACrefbtitle {{2016 IEEE Conference on Computer Vision and Pattern
  Recognition}} {{2016 IEEE Conference on Computer Vision and Pattern
  Recognition}}\ (\BPGS\ 770--778).
\newblock
\APACaddressPublisher{}{{IEEE}}.
\newblock
\begin{APACrefDOI} \doi{10.1109/CVPR.2016.90} \end{APACrefDOI}
\PrintBackRefs{\CurrentBib}

\bibitem [\protect \citeauthoryear {%
Hebart%
\ \protect \BOthers {.}}{%
Hebart%
\ \protect \BOthers {.}}{%
{\protect \APACyear {2023}}%
}]{%
Hebart2023-uf}
\APACinsertmetastar {%
Hebart2023-uf}%
\begin{APACrefauthors}%
Hebart, M\BPBI N.%
, Contier, O.%
, Teichmann, L.%
, Rockter, A\BPBI H.%
, Zheng, C\BPBI Y.%
, Kidder, A.%
, Corriveau, A.%
, Vaziri-Pashkam, M.%
\BCBL {}\ \BBA {} Baker, C\BPBI I.%
\end{APACrefauthors}%
\unskip\
\newblock
\APACrefYearMonthDay{2023}{}{}.
\newblock
{\BBOQ}\APACrefatitle {{THINGS}-data, a multimodal collection of large-scale
  datasets for investigating object representations in human brain and
  behavior} {{THINGS}-data, a multimodal collection of large-scale datasets for
  investigating object representations in human brain and behavior}.{\BBCQ}
\newblock
\APACjournalVolNumPages{{eLife}}{12}{}{Article e82580}.
\newblock
\begin{APACrefDOI} \doi{10.7554/eLife.82580} \end{APACrefDOI}
\PrintBackRefs{\CurrentBib}

\bibitem [\protect \citeauthoryear {%
Hebart%
\ \protect \BOthers {.}}{%
Hebart%
\ \protect \BOthers {.}}{%
{\protect \APACyear {2019}}%
}]{%
Hebart2019-bn}
\APACinsertmetastar {%
Hebart2019-bn}%
\begin{APACrefauthors}%
Hebart, M\BPBI N.%
, Dickter, A\BPBI H.%
, Kidder, A.%
, Kwok, W\BPBI Y.%
, Corriveau, A.%
, Van~Wicklin, C.%
\BCBL {}\ \BBA {} Baker, C\BPBI I.%
\end{APACrefauthors}%
\unskip\
\newblock
\APACrefYearMonthDay{2019}{}{}.
\newblock
{\BBOQ}\APACrefatitle {{THINGS}: A database of 1,854 object concepts and more
  than 26,000 naturalistic object images} {{THINGS}: A database of 1,854 object
  concepts and more than 26,000 naturalistic object images}.{\BBCQ}
\newblock
\APACjournalVolNumPages{{PLOS} {ONE}}{14}{10}{Article e0223792}.
\newblock
\begin{APACrefDOI} \doi{10.1371/journal.pone.0223792} \end{APACrefDOI}
\PrintBackRefs{\CurrentBib}

\bibitem [\protect \citeauthoryear {%
Hebart%
\ \protect \BOthers {.}}{%
Hebart%
\ \protect \BOthers {.}}{%
{\protect \APACyear {2020}}%
}]{%
Hebart2020-pd}
\APACinsertmetastar {%
Hebart2020-pd}%
\begin{APACrefauthors}%
Hebart, M\BPBI N.%
, Zheng, C\BPBI Y.%
, Pereira, F.%
\BCBL {}\ \BBA {} Baker, C\BPBI I.%
\end{APACrefauthors}%
\unskip\
\newblock
\APACrefYearMonthDay{2020}{}{}.
\newblock
{\BBOQ}\APACrefatitle {Revealing the multidimensional mental representations of
  natural objects underlying human similarity judgements} {Revealing the
  multidimensional mental representations of natural objects underlying human
  similarity judgements}.{\BBCQ}
\newblock
\APACjournalVolNumPages{Nature Human Behaviour}{4}{11}{1173--1185}.
\newblock
\begin{APACrefDOI} \doi{10.1038/s41562-020-00951-3} \end{APACrefDOI}
\PrintBackRefs{\CurrentBib}

\bibitem [\protect \citeauthoryear {%
Huber%
}{%
Huber%
}{%
{\protect \APACyear {1964}}%
}]{%
Huber1964-rb}
\APACinsertmetastar {%
Huber1964-rb}%
\begin{APACrefauthors}%
Huber, P\BPBI J.%
\end{APACrefauthors}%
\unskip\
\newblock
\APACrefYearMonthDay{1964}{}{}.
\newblock
{\BBOQ}\APACrefatitle {Robust estimation of a location parameter} {Robust
  estimation of a location parameter}.{\BBCQ}
\newblock
\APACjournalVolNumPages{The Annals of Mathematical Statistics}{35}{1}{73--101}.
\newblock
\begin{APACrefDOI} \doi{10.1214/aoms/1177703732} \end{APACrefDOI}
\PrintBackRefs{\CurrentBib}

\bibitem [\protect \citeauthoryear {%
Huh%
\ \protect \BOthers {.}}{%
Huh%
\ \protect \BOthers {.}}{%
{\protect \APACyear {2024}}%
}]{%
Huh2024-wv}
\APACinsertmetastar {%
Huh2024-wv}%
\begin{APACrefauthors}%
Huh, M.%
, Cheung, B.%
, Wang, T.%
\BCBL {}\ \BBA {} Isola, P.%
\end{APACrefauthors}%
\unskip\
\newblock
\APACrefYearMonthDay{2024}{}{}.
\newblock
{\BBOQ}\APACrefatitle {Position: The {Platonic} representation hypothesis}
  {Position: The {Platonic} representation hypothesis}.{\BBCQ}
\newblock
\BIn{} \APACrefbtitle {{Proceedings of the 41st International Conference on
  Machine Learning}} {{Proceedings of the 41st International Conference on
  Machine Learning}}\ (\BVOL~235, \BPGS\ 20617--20642).
\newblock
\APACaddressPublisher{}{{PMLR}}.
\PrintBackRefs{\CurrentBib}

\bibitem [\protect \citeauthoryear {%
Ilharco%
\ \protect \BOthers {.}}{%
Ilharco%
\ \protect \BOthers {.}}{%
{\protect \APACyear {2021}}%
}]{%
Ilharco2021-oc}
\APACinsertmetastar {%
Ilharco2021-oc}%
\begin{APACrefauthors}%
Ilharco, G.%
, Wortsman, M.%
, Wightman, R.%
, Gordon, C.%
, Carlini, N.%
, Taori, R.%
, Dave, A.%
, Shankar, V.%
, Namkoong, H.%
, Miller, J.%
, Hajishirzi, H.%
, Farhadi, A.%
\BCBL {}\ \BBA {} Schmidt, L.%
\end{APACrefauthors}%
\unskip\
\newblock
\APACrefYearMonthDay{2021}{}{}.
\newblock
\APACrefbtitle {{OpenCLIP} ({Version} 0.1)} {{OpenCLIP} ({Version} 0.1)}\
  [{Computer software}].
\newblock
\APACaddressPublisher{}{Zenodo}.
\newblock
\begin{APACrefDOI} \doi{10.5281/zenodo.5143773} \end{APACrefDOI}
\PrintBackRefs{\CurrentBib}

\bibitem [\protect \citeauthoryear {%
Jha%
\ \protect \BOthers {.}}{%
Jha%
\ \protect \BOthers {.}}{%
{\protect \APACyear {2023}}%
}]{%
Jha2023-ou}
\APACinsertmetastar {%
Jha2023-ou}%
\begin{APACrefauthors}%
Jha, A.%
, Peterson, J\BPBI C.%
\BCBL {}\ \BBA {} Griffiths, T\BPBI L.%
\end{APACrefauthors}%
\unskip\
\newblock
\APACrefYearMonthDay{2023}{}{}.
\newblock
{\BBOQ}\APACrefatitle {Extracting low-dimensional psychological representations
  from convolutional neural networks} {Extracting low-dimensional psychological
  representations from convolutional neural networks}.{\BBCQ}
\newblock
\APACjournalVolNumPages{Cognitive Science}{47}{1}{Article e13226}.
\newblock
\begin{APACrefDOI} \doi{10.1111/cogs.13226} \end{APACrefDOI}
\PrintBackRefs{\CurrentBib}

\bibitem [\protect \citeauthoryear {%
Jozwik%
\ \protect \BOthers {.}}{%
Jozwik%
\ \protect \BOthers {.}}{%
{\protect \APACyear {2017}}%
}]{%
Jozwik2017-ha}
\APACinsertmetastar {%
Jozwik2017-ha}%
\begin{APACrefauthors}%
Jozwik, K\BPBI M.%
, Kriegeskorte, N.%
, Storrs, K\BPBI R.%
\BCBL {}\ \BBA {} Mur, M.%
\end{APACrefauthors}%
\unskip\
\newblock
\APACrefYearMonthDay{2017}{}{}.
\newblock
{\BBOQ}\APACrefatitle {Deep convolutional neural networks outperform
  feature-based but not categorical models in explaining object similarity
  judgments} {Deep convolutional neural networks outperform feature-based but
  not categorical models in explaining object similarity judgments}.{\BBCQ}
\newblock
\APACjournalVolNumPages{Frontiers in Psychology}{8}{}{Article 1726}.
\newblock
\begin{APACrefDOI} \doi{10.3389/fpsyg.2017.01726} \end{APACrefDOI}
\PrintBackRefs{\CurrentBib}

\bibitem [\protect \citeauthoryear {%
Khaligh-Razavi%
\ \BBA {} Kriegeskorte%
}{%
Khaligh-Razavi%
\ \BBA {} Kriegeskorte%
}{%
{\protect \APACyear {2014}}%
}]{%
Khaligh-Razavi2014-qu}
\APACinsertmetastar {%
Khaligh-Razavi2014-qu}%
\begin{APACrefauthors}%
Khaligh-Razavi, S\BHBI M.%
\BCBT {}\ \BBA {} Kriegeskorte, N.%
\end{APACrefauthors}%
\unskip\
\newblock
\APACrefYearMonthDay{2014}{}{}.
\newblock
{\BBOQ}\APACrefatitle {Deep supervised, but not unsupervised, models may
  explain {IT} cortical representation} {Deep supervised, but not unsupervised,
  models may explain {IT} cortical representation}.{\BBCQ}
\newblock
\APACjournalVolNumPages{{PLOS} Computational Biology}{10}{11}{Article
  e1003915}.
\newblock
\begin{APACrefDOI} \doi{10.1371/journal.pcbi.1003915} \end{APACrefDOI}
\PrintBackRefs{\CurrentBib}

\bibitem [\protect \citeauthoryear {%
Kheradpisheh%
\ \protect \BOthers {.}}{%
Kheradpisheh%
\ \protect \BOthers {.}}{%
{\protect \APACyear {2016}}%
}]{%
Kheradpisheh2016-jj}
\APACinsertmetastar {%
Kheradpisheh2016-jj}%
\begin{APACrefauthors}%
Kheradpisheh, S\BPBI R.%
, Ghodrati, M.%
, Ganjtabesh, M.%
\BCBL {}\ \BBA {} Masquelier, T.%
\end{APACrefauthors}%
\unskip\
\newblock
\APACrefYearMonthDay{2016}{}{}.
\newblock
{\BBOQ}\APACrefatitle {Deep networks can resemble human feed-forward vision in
  invariant object recognition} {Deep networks can resemble human feed-forward
  vision in invariant object recognition}.{\BBCQ}
\newblock
\APACjournalVolNumPages{Scientific Reports}{6}{}{Article 32672}.
\newblock
\begin{APACrefDOI} \doi{10.1038/srep32672} \end{APACrefDOI}
\PrintBackRefs{\CurrentBib}

\bibitem [\protect \citeauthoryear {%
King%
\ \protect \BOthers {.}}{%
King%
\ \protect \BOthers {.}}{%
{\protect \APACyear {2019}}%
}]{%
King2019-jw}
\APACinsertmetastar {%
King2019-jw}%
\begin{APACrefauthors}%
King, M\BPBI L.%
, Groen, I\BPBI I\BPBI A.%
, Steel, A.%
, Kravitz, D\BPBI J.%
\BCBL {}\ \BBA {} Baker, C\BPBI I.%
\end{APACrefauthors}%
\unskip\
\newblock
\APACrefYearMonthDay{2019}{}{}.
\newblock
{\BBOQ}\APACrefatitle {Similarity judgments and cortical visual responses
  reflect different properties of object and scene categories in naturalistic
  images} {Similarity judgments and cortical visual responses reflect different
  properties of object and scene categories in naturalistic images}.{\BBCQ}
\newblock
\APACjournalVolNumPages{NeuroImage}{197}{}{368--382}.
\newblock
\begin{APACrefDOI} \doi{10.1016/j.neuroimage.2019.04.079} \end{APACrefDOI}
\PrintBackRefs{\CurrentBib}

\bibitem [\protect \citeauthoryear {%
Kingma%
\ \BBA {} Ba%
}{%
Kingma%
\ \BBA {} Ba%
}{%
{\protect \APACyear {2015}}%
}]{%
Kingma2014-av}
\APACinsertmetastar {%
Kingma2014-av}%
\begin{APACrefauthors}%
Kingma, D\BPBI P.%
\BCBT {}\ \BBA {} Ba, J.%
\end{APACrefauthors}%
\unskip\
\newblock
\APACrefYearMonthDay{2015}{}{}.
\newblock
{\BBOQ}\APACrefatitle {{Adam}: A method for stochastic optimization} {{Adam}: A
  method for stochastic optimization}.{\BBCQ}
\newblock
\BIn{} \APACrefbtitle {{International Conference on Learning Representations}.}
  {{International Conference on Learning Representations}.}
\PrintBackRefs{\CurrentBib}

\bibitem [\protect \citeauthoryear {%
Kriegeskorte%
}{%
Kriegeskorte%
}{%
{\protect \APACyear {2015}}%
}]{%
Kriegeskorte2015-bi}
\APACinsertmetastar {%
Kriegeskorte2015-bi}%
\begin{APACrefauthors}%
Kriegeskorte, N.%
\end{APACrefauthors}%
\unskip\
\newblock
\APACrefYearMonthDay{2015}{}{}.
\newblock
{\BBOQ}\APACrefatitle {Deep neural networks: A new framework for modeling
  biological vision and brain information processing} {Deep neural networks: A
  new framework for modeling biological vision and brain information
  processing}.{\BBCQ}
\newblock
\APACjournalVolNumPages{Annual Review of Vision Science}{1}{}{417--446}.
\newblock
\begin{APACrefDOI} \doi{10.1146/annurev-vision-082114-035447} \end{APACrefDOI}
\PrintBackRefs{\CurrentBib}

\bibitem [\protect \citeauthoryear {%
Kriegeskorte%
\ \protect \BOthers {.}}{%
Kriegeskorte%
\ \protect \BOthers {.}}{%
{\protect \APACyear {2008}}%
}]{%
Kriegeskorte2008-ep}
\APACinsertmetastar {%
Kriegeskorte2008-ep}%
\begin{APACrefauthors}%
Kriegeskorte, N.%
, Mur, M.%
\BCBL {}\ \BBA {} Bandettini, P.%
\end{APACrefauthors}%
\unskip\
\newblock
\APACrefYearMonthDay{2008}{}{}.
\newblock
{\BBOQ}\APACrefatitle {Representational similarity analysis -- connecting the
  branches of systems neuroscience} {Representational similarity analysis --
  connecting the branches of systems neuroscience}.{\BBCQ}
\newblock
\APACjournalVolNumPages{Frontiers in Systems Neuroscience}{2}{}{Article 4}.
\newblock
\begin{APACrefDOI} \doi{10.3389/neuro.06.004.2008} \end{APACrefDOI}
\PrintBackRefs{\CurrentBib}

\bibitem [\protect \citeauthoryear {%
M{\'e}moli%
}{%
M{\'e}moli%
}{%
{\protect \APACyear {2011}}%
}]{%
Memoli2011-nt}
\APACinsertmetastar {%
Memoli2011-nt}%
\begin{APACrefauthors}%
M{\'e}moli, F.%
\end{APACrefauthors}%
\unskip\
\newblock
\APACrefYearMonthDay{2011}{}{}.
\newblock
{\BBOQ}\APACrefatitle {{Gromov--Wasserstein} distances and the metric approach
  to object matching} {{Gromov--Wasserstein} distances and the metric approach
  to object matching}.{\BBCQ}
\newblock
\APACjournalVolNumPages{Foundations of Computational
  Mathematics}{11}{4}{417--487}.
\newblock
\begin{APACrefDOI} \doi{10.1007/s10208-011-9093-5} \end{APACrefDOI}
\PrintBackRefs{\CurrentBib}

\bibitem [\protect \citeauthoryear {%
Miller%
\ \protect \BOthers {.}}{%
Miller%
\ \protect \BOthers {.}}{%
{\protect \APACyear {1990}}%
}]{%
Miller1990-nw}
\APACinsertmetastar {%
Miller1990-nw}%
\begin{APACrefauthors}%
Miller, G\BPBI A.%
, Beckwith, R.%
, Fellbaum, C.%
, Gross, D.%
\BCBL {}\ \BBA {} Miller, K\BPBI J.%
\end{APACrefauthors}%
\unskip\
\newblock
\APACrefYearMonthDay{1990}{}{}.
\newblock
{\BBOQ}\APACrefatitle {Introduction to {WordNet}: An on-line lexical database}
  {Introduction to {WordNet}: An on-line lexical database}.{\BBCQ}
\newblock
\APACjournalVolNumPages{International Journal of Lexicography}{3}{4}{235--244}.
\newblock
\begin{APACrefDOI} \doi{10.1093/ijl/3.4.235} \end{APACrefDOI}
\PrintBackRefs{\CurrentBib}

\bibitem [\protect \citeauthoryear {%
Muttenthaler%
, Dippel%
\BCBL {}\ \protect \BOthers {.}}{%
Muttenthaler%
, Dippel%
\BCBL {}\ \protect \BOthers {.}}{%
{\protect \APACyear {2023}}%
}]{%
Muttenthaler2022-mf}
\APACinsertmetastar {%
Muttenthaler2022-mf}%
\begin{APACrefauthors}%
Muttenthaler, L.%
, Dippel, J.%
, Linhardt, L.%
, Vandermeulen, R\BPBI A.%
\BCBL {}\ \BBA {} Kornblith, S.%
\end{APACrefauthors}%
\unskip\
\newblock
\APACrefYearMonthDay{2023}{}{}.
\newblock
{\BBOQ}\APACrefatitle {Human alignment of neural network representations}
  {Human alignment of neural network representations}.{\BBCQ}
\newblock
\BIn{} \APACrefbtitle {{International Conference on Learning Representations}.}
  {{International Conference on Learning Representations}.}
\PrintBackRefs{\CurrentBib}

\bibitem [\protect \citeauthoryear {%
Muttenthaler%
\ \protect \BOthers {.}}{%
Muttenthaler%
\ \protect \BOthers {.}}{%
{\protect \APACyear {2025}}%
}]{%
Muttenthaler2025-nb}
\APACinsertmetastar {%
Muttenthaler2025-nb}%
\begin{APACrefauthors}%
Muttenthaler, L.%
, Greff, K.%
, Born, F.%
, Spitzer, B.%
, Kornblith, S.%
, Mozer, M\BPBI C.%
, M{\"u}ller, K\BHBI R.%
, Unterthiner, T.%
\BCBL {}\ \BBA {} Lampinen, A\BPBI K.%
\end{APACrefauthors}%
\unskip\
\newblock
\APACrefYearMonthDay{2025}{}{}.
\newblock
{\BBOQ}\APACrefatitle {Aligning machine and human visual representations across
  abstraction levels} {Aligning machine and human visual representations across
  abstraction levels}.{\BBCQ}
\newblock
\APACjournalVolNumPages{Nature}{647}{8089}{349--355}.
\newblock
\begin{APACrefDOI} \doi{10.1038/s41586-025-09631-6} \end{APACrefDOI}
\PrintBackRefs{\CurrentBib}

\bibitem [\protect \citeauthoryear {%
Muttenthaler%
, Linhardt%
\BCBL {}\ \protect \BOthers {.}}{%
Muttenthaler%
, Linhardt%
\BCBL {}\ \protect \BOthers {.}}{%
{\protect \APACyear {2023}}%
}]{%
Muttenthaler2023-aw}
\APACinsertmetastar {%
Muttenthaler2023-aw}%
\begin{APACrefauthors}%
Muttenthaler, L.%
, Linhardt, L.%
, Dippel, J.%
, Vandermeulen, R\BPBI A.%
, Hermann, K.%
, Lampinen, A\BPBI K.%
\BCBL {}\ \BBA {} Kornblith, S.%
\end{APACrefauthors}%
\unskip\
\newblock
\APACrefYearMonthDay{2023}{}{}.
\newblock
{\BBOQ}\APACrefatitle {Improving neural network representations using human
  similarity judgments} {Improving neural network representations using human
  similarity judgments}.{\BBCQ}
\newblock
\BIn{} \APACrefbtitle {{Advances in Neural Information Processing Systems}}
  {{Advances in Neural Information Processing Systems}}\ (\BVOL~36, \BPGS\
  50978--51007).
\newblock
\APACaddressPublisher{}{{Curran Associates}}.
\newblock
\begin{APACrefDOI} \doi{10.52202/075280-2218} \end{APACrefDOI}
\PrintBackRefs{\CurrentBib}

\bibitem [\protect \citeauthoryear {%
Park%
\ \protect \BOthers {.}}{%
Park%
\ \protect \BOthers {.}}{%
{\protect \APACyear {2019}}%
}]{%
Park2019-nt}
\APACinsertmetastar {%
Park2019-nt}%
\begin{APACrefauthors}%
Park, W.%
, Kim, D.%
, Lu, Y.%
\BCBL {}\ \BBA {} Cho, M.%
\end{APACrefauthors}%
\unskip\
\newblock
\APACrefYearMonthDay{2019}{}{}.
\newblock
{\BBOQ}\APACrefatitle {Relational knowledge distillation} {Relational knowledge
  distillation}.{\BBCQ}
\newblock
\BIn{} \APACrefbtitle {{2019 IEEE/CVF Conference on Computer Vision and Pattern
  Recognition}} {{2019 IEEE/CVF Conference on Computer Vision and Pattern
  Recognition}}\ (\BPGS\ 3962--3971).
\newblock
\APACaddressPublisher{}{{IEEE}}.
\newblock
\begin{APACrefDOI} \doi{10.1109/CVPR.2019.00409} \end{APACrefDOI}
\PrintBackRefs{\CurrentBib}

\bibitem [\protect \citeauthoryear {%
Pedregosa%
\ \protect \BOthers {.}}{%
Pedregosa%
\ \protect \BOthers {.}}{%
{\protect \APACyear {2011}}%
}]{%
Pedregosa2011-sk}
\APACinsertmetastar {%
Pedregosa2011-sk}%
\begin{APACrefauthors}%
Pedregosa, F.%
, Varoquaux, G.%
, Gramfort, A.%
, Michel, V.%
, Thirion, B.%
, Grisel, O.%
, Blondel, M.%
, Prettenhofer, P.%
, Weiss, R.%
, Dubourg, V.%
, Vanderplas, J.%
, Passos, A.%
, Cournapeau, D.%
, Brucher, M.%
, Perrot, M.%
\BCBL {}\ \BBA {} Duchesnay, {\'E}.%
\end{APACrefauthors}%
\unskip\
\newblock
\APACrefYearMonthDay{2011}{}{}.
\newblock
{\BBOQ}\APACrefatitle {Scikit-learn: Machine learning in {Python}}
  {Scikit-learn: Machine learning in {Python}}.{\BBCQ}
\newblock
\APACjournalVolNumPages{Journal of Machine Learning
  Research}{12}{}{2825--2830}.
\PrintBackRefs{\CurrentBib}

\bibitem [\protect \citeauthoryear {%
Peterson%
\ \protect \BOthers {.}}{%
Peterson%
\ \protect \BOthers {.}}{%
{\protect \APACyear {2016}}%
}]{%
Peterson2016-hx}
\APACinsertmetastar {%
Peterson2016-hx}%
\begin{APACrefauthors}%
Peterson, J\BPBI C.%
, Abbott, J\BPBI T.%
\BCBL {}\ \BBA {} Griffiths, T\BPBI L.%
\end{APACrefauthors}%
\unskip\
\newblock
\APACrefYearMonthDay{2016}{}{}.
\newblock
{\BBOQ}\APACrefatitle {Adapting deep network features to capture psychological
  representations} {Adapting deep network features to capture psychological
  representations}.{\BBCQ}
\newblock
\BIn{} \APACrefbtitle {{Proceedings of the 38th Annual Meeting of the Cognitive
  Science Society}} {{Proceedings of the 38th Annual Meeting of the Cognitive
  Science Society}}\ (\BPGS\ 2363--2368).
\newblock
\APACaddressPublisher{}{{Cognitive Science Society}}.
\PrintBackRefs{\CurrentBib}

\bibitem [\protect \citeauthoryear {%
Peterson%
\ \protect \BOthers {.}}{%
Peterson%
\ \protect \BOthers {.}}{%
{\protect \APACyear {2018}}%
}]{%
Peterson2018-hb}
\APACinsertmetastar {%
Peterson2018-hb}%
\begin{APACrefauthors}%
Peterson, J\BPBI C.%
, Abbott, J\BPBI T.%
\BCBL {}\ \BBA {} Griffiths, T\BPBI L.%
\end{APACrefauthors}%
\unskip\
\newblock
\APACrefYearMonthDay{2018}{}{}.
\newblock
{\BBOQ}\APACrefatitle {Evaluating (and improving) the correspondence between
  deep neural networks and human representations} {Evaluating (and improving)
  the correspondence between deep neural networks and human
  representations}.{\BBCQ}
\newblock
\APACjournalVolNumPages{Cognitive Science}{42}{8}{2648--2669}.
\newblock
\begin{APACrefDOI} \doi{10.1111/cogs.12670} \end{APACrefDOI}
\PrintBackRefs{\CurrentBib}

\bibitem [\protect \citeauthoryear {%
Peyr{\'e}%
\ \BBA {} Cuturi%
}{%
Peyr{\'e}%
\ \BBA {} Cuturi%
}{%
{\protect \APACyear {2019}}%
}]{%
Peyre2019-ct}
\APACinsertmetastar {%
Peyre2019-ct}%
\begin{APACrefauthors}%
Peyr{\'e}, G.%
\BCBT {}\ \BBA {} Cuturi, M.%
\end{APACrefauthors}%
\unskip\
\newblock
\APACrefYearMonthDay{2019}{}{}.
\newblock
{\BBOQ}\APACrefatitle {Computational optimal transport: With applications to
  data science} {Computational optimal transport: With applications to data
  science}.{\BBCQ}
\newblock
\APACjournalVolNumPages{Foundations and Trends in Machine
  Learning}{11}{5--6}{355--607}.
\newblock
\begin{APACrefDOI} \doi{10.1561/2200000073} \end{APACrefDOI}
\PrintBackRefs{\CurrentBib}

\bibitem [\protect \citeauthoryear {%
Peyr{\'e}%
\ \protect \BOthers {.}}{%
Peyr{\'e}%
\ \protect \BOthers {.}}{%
{\protect \APACyear {2016}}%
}]{%
Peyre2016-dj}
\APACinsertmetastar {%
Peyre2016-dj}%
\begin{APACrefauthors}%
Peyr{\'e}, G.%
, Cuturi, M.%
\BCBL {}\ \BBA {} Solomon, J.%
\end{APACrefauthors}%
\unskip\
\newblock
\APACrefYearMonthDay{2016}{}{}.
\newblock
{\BBOQ}\APACrefatitle {{Gromov--Wasserstein} averaging of kernel and distance
  matrices} {{Gromov--Wasserstein} averaging of kernel and distance
  matrices}.{\BBCQ}
\newblock
\BIn{} \APACrefbtitle {{Proceedings of the 33rd International Conference on
  Machine Learning}} {{Proceedings of the 33rd International Conference on
  Machine Learning}}\ (\BVOL~48, \BPGS\ 2664--2672).
\newblock
\APACaddressPublisher{}{{PMLR}}.
\PrintBackRefs{\CurrentBib}

\bibitem [\protect \citeauthoryear {%
Radford%
\ \protect \BOthers {.}}{%
Radford%
\ \protect \BOthers {.}}{%
{\protect \APACyear {2021}}%
}]{%
Radford2021-fu}
\APACinsertmetastar {%
Radford2021-fu}%
\begin{APACrefauthors}%
Radford, A.%
, Kim, J\BPBI W.%
, Hallacy, C.%
, Ramesh, A.%
, Goh, G.%
, Agarwal, S.%
, Sastry, G.%
, Askell, A.%
, Mishkin, P.%
, Clark, J.%
, Krueger, G.%
\BCBL {}\ \BBA {} Sutskever, I.%
\end{APACrefauthors}%
\unskip\
\newblock
\APACrefYearMonthDay{2021}{}{}.
\newblock
{\BBOQ}\APACrefatitle {Learning transferable visual models from natural
  language supervision} {Learning transferable visual models from natural
  language supervision}.{\BBCQ}
\newblock
\BIn{} \APACrefbtitle {{Proceedings of the 38th International Conference on
  Machine Learning}} {{Proceedings of the 38th International Conference on
  Machine Learning}}\ (\BVOL~139, \BPGS\ 8748--8763).
\newblock
\APACaddressPublisher{}{{PMLR}}.
\PrintBackRefs{\CurrentBib}

\bibitem [\protect \citeauthoryear {%
Roads%
\ \BBA {} Love%
}{%
Roads%
\ \BBA {} Love%
}{%
{\protect \APACyear {2021}}%
}]{%
Roads2021-nz}
\APACinsertmetastar {%
Roads2021-nz}%
\begin{APACrefauthors}%
Roads, B\BPBI D.%
\BCBT {}\ \BBA {} Love, B\BPBI C.%
\end{APACrefauthors}%
\unskip\
\newblock
\APACrefYearMonthDay{2021}{}{}.
\newblock
{\BBOQ}\APACrefatitle {Enriching {ImageNet} with human similarity judgments and
  psychological embeddings} {Enriching {ImageNet} with human similarity
  judgments and psychological embeddings}.{\BBCQ}
\newblock
\BIn{} \APACrefbtitle {{2021 IEEE/CVF Conference on Computer Vision and Pattern
  Recognition}} {{2021 IEEE/CVF Conference on Computer Vision and Pattern
  Recognition}}\ (\BPGS\ 3546--3556).
\newblock
\APACaddressPublisher{}{{IEEE}}.
\newblock
\begin{APACrefDOI} \doi{10.1109/CVPR46437.2021.00355} \end{APACrefDOI}
\PrintBackRefs{\CurrentBib}

\bibitem [\protect \citeauthoryear {%
Roads%
\ \BBA {} Mozer%
}{%
Roads%
\ \BBA {} Mozer%
}{%
{\protect \APACyear {2019}}%
}]{%
Roads2019-jk}
\APACinsertmetastar {%
Roads2019-jk}%
\begin{APACrefauthors}%
Roads, B\BPBI D.%
\BCBT {}\ \BBA {} Mozer, M\BPBI C.%
\end{APACrefauthors}%
\unskip\
\newblock
\APACrefYearMonthDay{2019}{}{}.
\newblock
{\BBOQ}\APACrefatitle {Obtaining psychological embeddings through joint kernel
  and metric learning} {Obtaining psychological embeddings through joint kernel
  and metric learning}.{\BBCQ}
\newblock
\APACjournalVolNumPages{Behavior Research Methods}{51}{5}{2180--2193}.
\newblock
\begin{APACrefDOI} \doi{10.3758/s13428-019-01285-3} \end{APACrefDOI}
\PrintBackRefs{\CurrentBib}

\bibitem [\protect \citeauthoryear {%
Russakovsky%
\ \protect \BOthers {.}}{%
Russakovsky%
\ \protect \BOthers {.}}{%
{\protect \APACyear {2015}}%
}]{%
Russakovsky2015-il}
\APACinsertmetastar {%
Russakovsky2015-il}%
\begin{APACrefauthors}%
Russakovsky, O.%
, Deng, J.%
, Su, H.%
, Krause, J.%
, Satheesh, S.%
, Ma, S.%
, Huang, Z.%
, Karpathy, A.%
, Khosla, A.%
, Bernstein, M.%
, Berg, A\BPBI C.%
\BCBL {}\ \BBA {} Fei-Fei, L.%
\end{APACrefauthors}%
\unskip\
\newblock
\APACrefYearMonthDay{2015}{}{}.
\newblock
{\BBOQ}\APACrefatitle {{ImageNet} large scale visual recognition challenge}
  {{ImageNet} large scale visual recognition challenge}.{\BBCQ}
\newblock
\APACjournalVolNumPages{International Journal of Computer
  Vision}{115}{3}{211--252}.
\newblock
\begin{APACrefDOI} \doi{10.1007/s11263-015-0816-y} \end{APACrefDOI}
\PrintBackRefs{\CurrentBib}

\bibitem [\protect \citeauthoryear {%
Sch{\"o}nemann%
}{%
Sch{\"o}nemann%
}{%
{\protect \APACyear {1966}}%
}]{%
Schonemann1966-op}
\APACinsertmetastar {%
Schonemann1966-op}%
\begin{APACrefauthors}%
Sch{\"o}nemann, P\BPBI H.%
\end{APACrefauthors}%
\unskip\
\newblock
\APACrefYearMonthDay{1966}{}{}.
\newblock
{\BBOQ}\APACrefatitle {A generalized solution of the orthogonal {Procrustes}
  problem} {A generalized solution of the orthogonal {Procrustes}
  problem}.{\BBCQ}
\newblock
\APACjournalVolNumPages{Psychometrika}{31}{1}{1--10}.
\newblock
\begin{APACrefDOI} \doi{10.1007/BF02289451} \end{APACrefDOI}
\PrintBackRefs{\CurrentBib}

\bibitem [\protect \citeauthoryear {%
Schuhmann%
\ \protect \BOthers {.}}{%
Schuhmann%
\ \protect \BOthers {.}}{%
{\protect \APACyear {2022}}%
}]{%
Schuhmann2022-lx}
\APACinsertmetastar {%
Schuhmann2022-lx}%
\begin{APACrefauthors}%
Schuhmann, C.%
, Beaumont, R.%
, Vencu, R.%
, Gordon, C.%
, Wightman, R.%
, Cherti, M.%
, Coombes, T.%
, Katta, A.%
, Mullis, C.%
, Wortsman, M.%
, Schramowski, P.%
, Kundurthy, S.%
, Crowson, K.%
, Schmidt, L.%
, Kaczmarczyk, R.%
\BCBL {}\ \BBA {} Jitsev, J.%
\end{APACrefauthors}%
\unskip\
\newblock
\APACrefYearMonthDay{2022}{}{}.
\newblock
{\BBOQ}\APACrefatitle {{LAION}-5{B}: An open large-scale dataset for training
  next generation image-text models} {{LAION}-5{B}: An open large-scale dataset
  for training next generation image-text models}.{\BBCQ}
\newblock
\BIn{} \APACrefbtitle {{Advances in Neural Information Processing Systems}}
  {{Advances in Neural Information Processing Systems}}\ (\BVOL~35, \BPGS\
  25278--25294).
\newblock
\APACaddressPublisher{}{{Curran Associates}}.
\newblock
\begin{APACrefDOI} \doi{10.52202/068431-1833} \end{APACrefDOI}
\PrintBackRefs{\CurrentBib}

\bibitem [\protect \citeauthoryear {%
Schuhmann%
\ \protect \BOthers {.}}{%
Schuhmann%
\ \protect \BOthers {.}}{%
{\protect \APACyear {2021}}%
}]{%
Schuhmann2021-dn}
\APACinsertmetastar {%
Schuhmann2021-dn}%
\begin{APACrefauthors}%
Schuhmann, C.%
, Vencu, R.%
, Beaumont, R.%
, Kaczmarczyk, R.%
, Mullis, C.%
, Katta, A.%
, Coombes, T.%
, Jitsev, J.%
\BCBL {}\ \BBA {} Komatsuzaki, A.%
\end{APACrefauthors}%
\unskip\
\newblock
\APACrefYearMonthDay{2021}{}{}.
\newblock
\APACrefbtitle {{LAION}-400{M}: Open dataset of {CLIP}-filtered 400 million
  image-text pairs} {{LAION}-400{M}: Open dataset of {CLIP}-filtered 400
  million image-text pairs}\ [{Preprint}].
\newblock
\APAChowpublished {ar{X}iv}.
\newblock
\APACrefnote{{Presented at the NeurIPS 2021 Workshop on Data-Centric AI}}
\newblock
\begin{APACrefDOI} \doi{10.48550/arXiv.2111.02114} \end{APACrefDOI}
\PrintBackRefs{\CurrentBib}

\bibitem [\protect \citeauthoryear {%
Shepard%
}{%
Shepard%
}{%
{\protect \APACyear {1987}}%
}]{%
Shepard1987-ky}
\APACinsertmetastar {%
Shepard1987-ky}%
\begin{APACrefauthors}%
Shepard, R\BPBI N.%
\end{APACrefauthors}%
\unskip\
\newblock
\APACrefYearMonthDay{1987}{}{}.
\newblock
{\BBOQ}\APACrefatitle {Toward a universal law of generalization for
  psychological science} {Toward a universal law of generalization for
  psychological science}.{\BBCQ}
\newblock
\APACjournalVolNumPages{Science}{237}{4820}{1317--1323}.
\newblock
\begin{APACrefDOI} \doi{10.1126/science.3629243} \end{APACrefDOI}
\PrintBackRefs{\CurrentBib}

\bibitem [\protect \citeauthoryear {%
Storrs%
\ \protect \BOthers {.}}{%
Storrs%
\ \protect \BOthers {.}}{%
{\protect \APACyear {2021}}%
}]{%
Storrs2021-dd}
\APACinsertmetastar {%
Storrs2021-dd}%
\begin{APACrefauthors}%
Storrs, K\BPBI R.%
, Kietzmann, T\BPBI C.%
, Walther, A.%
, Mehrer, J.%
\BCBL {}\ \BBA {} Kriegeskorte, N.%
\end{APACrefauthors}%
\unskip\
\newblock
\APACrefYearMonthDay{2021}{}{}.
\newblock
{\BBOQ}\APACrefatitle {Diverse deep neural networks all predict human inferior
  temporal cortex well, after training and fitting} {Diverse deep neural
  networks all predict human inferior temporal cortex well, after training and
  fitting}.{\BBCQ}
\newblock
\APACjournalVolNumPages{Journal of Cognitive Neuroscience}{33}{10}{2044--2064}.
\newblock
\begin{APACrefDOI} \doi{10.1162/jocn_a_01755} \end{APACrefDOI}
\PrintBackRefs{\CurrentBib}

\bibitem [\protect \citeauthoryear {%
Takahashi%
\ \protect \BOthers {.}}{%
Takahashi%
\ \protect \BOthers {.}}{%
{\protect \APACyear {2026}}%
}]{%
Takahashi2026-lk}
\APACinsertmetastar {%
Takahashi2026-lk}%
\begin{APACrefauthors}%
Takahashi, S.%
, Sasaki, M.%
, Takeda, K.%
\BCBL {}\ \BBA {} Oizumi, M.%
\end{APACrefauthors}%
\unskip\
\newblock
\APACrefYearMonthDay{2026}{}{}.
\newblock
{\BBOQ}\APACrefatitle {Investigating fine- and coarse-grained structural
  correspondences between deep neural networks and human object image
  similarity judgments using unsupervised alignment} {Investigating fine- and
  coarse-grained structural correspondences between deep neural networks and
  human object image similarity judgments using unsupervised alignment}.{\BBCQ}
\newblock
\APACjournalVolNumPages{Neural Networks}{195}{}{Article 108222}.
\newblock
\begin{APACrefDOI} \doi{10.1016/j.neunet.2025.108222} \end{APACrefDOI}
\PrintBackRefs{\CurrentBib}

\bibitem [\protect \citeauthoryear {%
Takeda%
, Abe%
\BCBL {}\ \protect \BOthers {.}}{%
Takeda%
, Abe%
\BCBL {}\ \protect \BOthers {.}}{%
{\protect \APACyear {2025}}%
}]{%
Takeda2025-za}
\APACinsertmetastar {%
Takeda2025-za}%
\begin{APACrefauthors}%
Takeda, K.%
, Abe, K.%
, Kitazono, J.%
\BCBL {}\ \BBA {} Oizumi, M.%
\end{APACrefauthors}%
\unskip\
\newblock
\APACrefYearMonthDay{2025}{}{}.
\newblock
{\BBOQ}\APACrefatitle {Unsupervised alignment reveals structural commonalities
  and differences in neural representations of natural scenes across
  individuals and brain areas} {Unsupervised alignment reveals structural
  commonalities and differences in neural representations of natural scenes
  across individuals and brain areas}.{\BBCQ}
\newblock
\APACjournalVolNumPages{{iScience}}{28}{5}{Article 112427}.
\newblock
\begin{APACrefDOI} \doi{10.1016/j.isci.2025.112427} \end{APACrefDOI}
\PrintBackRefs{\CurrentBib}

\bibitem [\protect \citeauthoryear {%
Takeda%
, Sasaki%
\BCBL {}\ \protect \BOthers {.}}{%
Takeda%
, Sasaki%
\BCBL {}\ \protect \BOthers {.}}{%
{\protect \APACyear {2025}}%
}]{%
Takeda2025-wi}
\APACinsertmetastar {%
Takeda2025-wi}%
\begin{APACrefauthors}%
Takeda, K.%
, Sasaki, M.%
, Abe, K.%
\BCBL {}\ \BBA {} Oizumi, M.%
\end{APACrefauthors}%
\unskip\
\newblock
\APACrefYearMonthDay{2025}{}{}.
\newblock
{\BBOQ}\APACrefatitle {Unsupervised alignment in neuroscience: Introducing a
  toolbox for {Gromov--Wasserstein} optimal transport} {Unsupervised alignment
  in neuroscience: Introducing a toolbox for {Gromov--Wasserstein} optimal
  transport}.{\BBCQ}
\newblock
\APACjournalVolNumPages{Journal of Neuroscience Methods}{419}{}{Article
  110443}.
\newblock
\begin{APACrefDOI} \doi{10.1016/j.jneumeth.2025.110443} \end{APACrefDOI}
\PrintBackRefs{\CurrentBib}

\bibitem [\protect \citeauthoryear {%
Virtanen%
\ \protect \BOthers {.}}{%
Virtanen%
\ \protect \BOthers {.}}{%
{\protect \APACyear {2020}}%
}]{%
Virtanen2020-sp}
\APACinsertmetastar {%
Virtanen2020-sp}%
\begin{APACrefauthors}%
Virtanen, P.%
, Gommers, R.%
, Oliphant, T\BPBI E.%
, Haberland, M.%
, Reddy, T.%
, Cournapeau, D.%
, Burovski, E.%
, Peterson, P.%
, Weckesser, W.%
, Bright, J.%
, van~der Walt, S\BPBI J.%
, Brett, M.%
, Wilson, J.%
, Millman, K\BPBI J.%
, Mayorov, N.%
, Nelson, A\BPBI R\BPBI J.%
, Jones, E.%
, Kern, R.%
, Larson, E.%
\BDBL {}van Mulbregt, P.%
\end{APACrefauthors}%
\unskip\
\newblock
\APACrefYearMonthDay{2020}{}{}.
\newblock
{\BBOQ}\APACrefatitle {{SciPy} 1.0: Fundamental algorithms for scientific
  computing in {Python}} {{SciPy} 1.0: Fundamental algorithms for scientific
  computing in {Python}}.{\BBCQ}
\newblock
\APACjournalVolNumPages{Nature Methods}{17}{3}{261--272}.
\newblock
\begin{APACrefDOI} \doi{10.1038/s41592-019-0686-2} \end{APACrefDOI}
\PrintBackRefs{\CurrentBib}

\bibitem [\protect \citeauthoryear {%
Yamins%
\ \BBA {} DiCarlo%
}{%
Yamins%
\ \BBA {} DiCarlo%
}{%
{\protect \APACyear {2016}}%
}]{%
Yamins2016-uf}
\APACinsertmetastar {%
Yamins2016-uf}%
\begin{APACrefauthors}%
Yamins, D\BPBI L\BPBI K.%
\BCBT {}\ \BBA {} DiCarlo, J\BPBI J.%
\end{APACrefauthors}%
\unskip\
\newblock
\APACrefYearMonthDay{2016}{}{}.
\newblock
{\BBOQ}\APACrefatitle {Using goal-driven deep learning models to understand
  sensory cortex} {Using goal-driven deep learning models to understand sensory
  cortex}.{\BBCQ}
\newblock
\APACjournalVolNumPages{Nature Neuroscience}{19}{3}{356--365}.
\newblock
\begin{APACrefDOI} \doi{10.1038/nn.4244} \end{APACrefDOI}
\PrintBackRefs{\CurrentBib}

\bibitem [\protect \citeauthoryear {%
Yamins%
\ \protect \BOthers {.}}{%
Yamins%
\ \protect \BOthers {.}}{%
{\protect \APACyear {2014}}%
}]{%
Yamins2014-xv}
\APACinsertmetastar {%
Yamins2014-xv}%
\begin{APACrefauthors}%
Yamins, D\BPBI L\BPBI K.%
, Hong, H.%
, Cadieu, C\BPBI F.%
, Solomon, E\BPBI A.%
, Seibert, D.%
\BCBL {}\ \BBA {} DiCarlo, J\BPBI J.%
\end{APACrefauthors}%
\unskip\
\newblock
\APACrefYearMonthDay{2014}{}{}.
\newblock
{\BBOQ}\APACrefatitle {Performance-optimized hierarchical models predict neural
  responses in higher visual cortex} {Performance-optimized hierarchical models
  predict neural responses in higher visual cortex}.{\BBCQ}
\newblock
\APACjournalVolNumPages{Proceedings of the National Academy of
  Sciences}{111}{23}{8619--8624}.
\newblock
\begin{APACrefDOI} \doi{10.1073/pnas.1403112111} \end{APACrefDOI}
\PrintBackRefs{\CurrentBib}

\bibitem [\protect \citeauthoryear {%
Zheng%
\ \protect \BOthers {.}}{%
Zheng%
\ \protect \BOthers {.}}{%
{\protect \APACyear {2019}}%
}]{%
Zheng2019-io}
\APACinsertmetastar {%
Zheng2019-io}%
\begin{APACrefauthors}%
Zheng, C\BPBI Y.%
, Pereira, F.%
, Baker, C\BPBI I.%
\BCBL {}\ \BBA {} Hebart, M\BPBI N.%
\end{APACrefauthors}%
\unskip\
\newblock
\APACrefYearMonthDay{2019}{}{}.
\newblock
{\BBOQ}\APACrefatitle {Revealing interpretable object representations from
  human behavior} {Revealing interpretable object representations from human
  behavior}.{\BBCQ}
\newblock
\BIn{} \APACrefbtitle {{International Conference on Learning Representations}.}
  {{International Conference on Learning Representations}.}
\PrintBackRefs{\CurrentBib}

\end{thebibliography}


\begin{thebibliography}{}

\bibitem [\protect \citeauthoryear {%
Caron%
\ \protect \BOthers {.}}{%
Caron%
\ \protect \BOthers {.}}{%
{\protect \APACyear {2021}}%
}]{%
Caron2021-dn}
\APACinsertmetastar {%
Caron2021-dn}%
\begin{APACrefauthors}%
Caron, M.%
, Touvron, H.%
, Misra, I.%
, J{\'e}gou, H.%
, Mairal, J.%
, Bojanowski, P.%
\BCBL {}\ \BBA {} Joulin, A.%
\end{APACrefauthors}%
\unskip\
\newblock
\APACrefYearMonthDay{2021}{}{}.
\newblock
{\BBOQ}\APACrefatitle {Emerging properties in self-supervised vision
  transformers} {Emerging properties in self-supervised vision
  transformers}.{\BBCQ}
\newblock
\BIn{} \APACrefbtitle {{2021 IEEE/CVF International Conference on Computer
  Vision}} {{2021 IEEE/CVF International Conference on Computer Vision}}\
  (\BPGS\ 9630--9640).
\newblock
\APACaddressPublisher{}{{IEEE}}.
\newblock
\begin{APACrefDOI} \doi{10.1109/ICCV48922.2021.00951} \end{APACrefDOI}
\PrintBackRefs{\CurrentBib}

\bibitem [\protect \citeauthoryear {%
Chen%
\ \protect \BOthers {.}}{%
Chen%
\ \protect \BOthers {.}}{%
{\protect \APACyear {2020}}%
}]{%
Chen2020-sv}
\APACinsertmetastar {%
Chen2020-sv}%
\begin{APACrefauthors}%
Chen, T.%
, Kornblith, S.%
, Swersky, K.%
, Norouzi, M.%
\BCBL {}\ \BBA {} Hinton, G\BPBI E.%
\end{APACrefauthors}%
\unskip\
\newblock
\APACrefYearMonthDay{2020}{}{}.
\newblock
{\BBOQ}\APACrefatitle {Big self-supervised models are strong semi-supervised
  learners} {Big self-supervised models are strong semi-supervised
  learners}.{\BBCQ}
\newblock
\BIn{} \APACrefbtitle {{Advances in Neural Information Processing Systems}}
  {{Advances in Neural Information Processing Systems}}\ (\BVOL~33, \BPGS\
  22243--22255).
\newblock
\APACaddressPublisher{}{{Curran Associates}}.
\PrintBackRefs{\CurrentBib}

\bibitem [\protect \citeauthoryear {%
Cherti%
\ \protect \BOthers {.}}{%
Cherti%
\ \protect \BOthers {.}}{%
{\protect \APACyear {2023}}%
}]{%
Cherti2022-fl}
\APACinsertmetastar {%
Cherti2022-fl}%
\begin{APACrefauthors}%
Cherti, M.%
, Beaumont, R.%
, Wightman, R.%
, Wortsman, M.%
, Ilharco, G.%
, Gordon, C.%
, Schuhmann, C.%
, Schmidt, L.%
\BCBL {}\ \BBA {} Jitsev, J.%
\end{APACrefauthors}%
\unskip\
\newblock
\APACrefYearMonthDay{2023}{}{}.
\newblock
{\BBOQ}\APACrefatitle {Reproducible scaling laws for contrastive language-image
  learning} {Reproducible scaling laws for contrastive language-image
  learning}.{\BBCQ}
\newblock
\BIn{} \APACrefbtitle {{2023 IEEE/CVF Conference on Computer Vision and Pattern
  Recognition}} {{2023 IEEE/CVF Conference on Computer Vision and Pattern
  Recognition}}\ (\BPGS\ 2818--2829).
\newblock
\APACaddressPublisher{}{{IEEE}}.
\newblock
\begin{APACrefDOI} \doi{10.1109/CVPR52729.2023.00276} \end{APACrefDOI}
\PrintBackRefs{\CurrentBib}

\bibitem [\protect \citeauthoryear {%
Deng%
\ \protect \BOthers {.}}{%
Deng%
\ \protect \BOthers {.}}{%
{\protect \APACyear {2009}}%
}]{%
Deng2009-ea}
\APACinsertmetastar {%
Deng2009-ea}%
\begin{APACrefauthors}%
Deng, J.%
, Dong, W.%
, Socher, R.%
, Li, L\BHBI J.%
, Li, K.%
\BCBL {}\ \BBA {} Fei-Fei, L.%
\end{APACrefauthors}%
\unskip\
\newblock
\APACrefYearMonthDay{2009}{}{}.
\newblock
{\BBOQ}\APACrefatitle {{ImageNet}: A large-scale hierarchical image database}
  {{ImageNet}: A large-scale hierarchical image database}.{\BBCQ}
\newblock
\BIn{} \APACrefbtitle {{2009 IEEE Conference on Computer Vision and Pattern
  Recognition}} {{2009 IEEE Conference on Computer Vision and Pattern
  Recognition}}\ (\BPGS\ 248--255).
\newblock
\APACaddressPublisher{}{{IEEE}}.
\newblock
\begin{APACrefDOI} \doi{10.1109/CVPR.2009.5206848} \end{APACrefDOI}
\PrintBackRefs{\CurrentBib}

\bibitem [\protect \citeauthoryear {%
He%
\ \protect \BOthers {.}}{%
He%
\ \protect \BOthers {.}}{%
{\protect \APACyear {2022}}%
}]{%
He2022-ma}
\APACinsertmetastar {%
He2022-ma}%
\begin{APACrefauthors}%
He, K.%
, Chen, X.%
, Xie, S.%
, Li, Y.%
, Doll{\'a}r, P.%
\BCBL {}\ \BBA {} Girshick, R.%
\end{APACrefauthors}%
\unskip\
\newblock
\APACrefYearMonthDay{2022}{}{}.
\newblock
{\BBOQ}\APACrefatitle {Masked autoencoders are scalable vision learners}
  {Masked autoencoders are scalable vision learners}.{\BBCQ}
\newblock
\BIn{} \APACrefbtitle {{2022 IEEE/CVF Conference on Computer Vision and Pattern
  Recognition}} {{2022 IEEE/CVF Conference on Computer Vision and Pattern
  Recognition}}\ (\BPGS\ 15979--15988).
\newblock
\APACaddressPublisher{}{{IEEE}}.
\newblock
\begin{APACrefDOI} \doi{10.1109/CVPR52688.2022.01553} \end{APACrefDOI}
\PrintBackRefs{\CurrentBib}

\bibitem [\protect \citeauthoryear {%
He%
\ \protect \BOthers {.}}{%
He%
\ \protect \BOthers {.}}{%
{\protect \APACyear {2016}}%
}]{%
He2016-je}
\APACinsertmetastar {%
He2016-je}%
\begin{APACrefauthors}%
He, K.%
, Zhang, X.%
, Ren, S.%
\BCBL {}\ \BBA {} Sun, J.%
\end{APACrefauthors}%
\unskip\
\newblock
\APACrefYearMonthDay{2016}{}{}.
\newblock
{\BBOQ}\APACrefatitle {Deep residual learning for image recognition} {Deep
  residual learning for image recognition}.{\BBCQ}
\newblock
\BIn{} \APACrefbtitle {{2016 IEEE Conference on Computer Vision and Pattern
  Recognition}} {{2016 IEEE Conference on Computer Vision and Pattern
  Recognition}}\ (\BPGS\ 770--778).
\newblock
\APACaddressPublisher{}{{IEEE}}.
\newblock
\begin{APACrefDOI} \doi{10.1109/CVPR.2016.90} \end{APACrefDOI}
\PrintBackRefs{\CurrentBib}

\bibitem [\protect \citeauthoryear {%
Hebart%
\ \protect \BOthers {.}}{%
Hebart%
\ \protect \BOthers {.}}{%
{\protect \APACyear {2023}}%
}]{%
Hebart2023-uf}
\APACinsertmetastar {%
Hebart2023-uf}%
\begin{APACrefauthors}%
Hebart, M\BPBI N.%
, Contier, O.%
, Teichmann, L.%
, Rockter, A\BPBI H.%
, Zheng, C\BPBI Y.%
, Kidder, A.%
, Corriveau, A.%
, Vaziri-Pashkam, M.%
\BCBL {}\ \BBA {} Baker, C\BPBI I.%
\end{APACrefauthors}%
\unskip\
\newblock
\APACrefYearMonthDay{2023}{}{}.
\newblock
{\BBOQ}\APACrefatitle {{THINGS}-data, a multimodal collection of large-scale
  datasets for investigating object representations in human brain and
  behavior} {{THINGS}-data, a multimodal collection of large-scale datasets for
  investigating object representations in human brain and behavior}.{\BBCQ}
\newblock
\APACjournalVolNumPages{{eLife}}{12}{}{Article e82580}.
\newblock
\begin{APACrefDOI} \doi{10.7554/eLife.82580} \end{APACrefDOI}
\PrintBackRefs{\CurrentBib}

\bibitem [\protect \citeauthoryear {%
Hebart%
\ \protect \BOthers {.}}{%
Hebart%
\ \protect \BOthers {.}}{%
{\protect \APACyear {2019}}%
}]{%
Hebart2019-bn}
\APACinsertmetastar {%
Hebart2019-bn}%
\begin{APACrefauthors}%
Hebart, M\BPBI N.%
, Dickter, A\BPBI H.%
, Kidder, A.%
, Kwok, W\BPBI Y.%
, Corriveau, A.%
, Van~Wicklin, C.%
\BCBL {}\ \BBA {} Baker, C\BPBI I.%
\end{APACrefauthors}%
\unskip\
\newblock
\APACrefYearMonthDay{2019}{}{}.
\newblock
{\BBOQ}\APACrefatitle {{THINGS}: A database of 1,854 object concepts and more
  than 26,000 naturalistic object images} {{THINGS}: A database of 1,854 object
  concepts and more than 26,000 naturalistic object images}.{\BBCQ}
\newblock
\APACjournalVolNumPages{{PLOS} {ONE}}{14}{10}{Article e0223792}.
\newblock
\begin{APACrefDOI} \doi{10.1371/journal.pone.0223792} \end{APACrefDOI}
\PrintBackRefs{\CurrentBib}

\bibitem [\protect \citeauthoryear {%
Hebart%
\ \protect \BOthers {.}}{%
Hebart%
\ \protect \BOthers {.}}{%
{\protect \APACyear {2020}}%
}]{%
Hebart2020-pd}
\APACinsertmetastar {%
Hebart2020-pd}%
\begin{APACrefauthors}%
Hebart, M\BPBI N.%
, Zheng, C\BPBI Y.%
, Pereira, F.%
\BCBL {}\ \BBA {} Baker, C\BPBI I.%
\end{APACrefauthors}%
\unskip\
\newblock
\APACrefYearMonthDay{2020}{}{}.
\newblock
{\BBOQ}\APACrefatitle {Revealing the multidimensional mental representations of
  natural objects underlying human similarity judgements} {Revealing the
  multidimensional mental representations of natural objects underlying human
  similarity judgements}.{\BBCQ}
\newblock
\APACjournalVolNumPages{Nature Human Behaviour}{4}{11}{1173--1185}.
\newblock
\begin{APACrefDOI} \doi{10.1038/s41562-020-00951-3} \end{APACrefDOI}
\PrintBackRefs{\CurrentBib}

\bibitem [\protect \citeauthoryear {%
Ilharco%
\ \protect \BOthers {.}}{%
Ilharco%
\ \protect \BOthers {.}}{%
{\protect \APACyear {2021}}%
}]{%
Ilharco2021-oc}
\APACinsertmetastar {%
Ilharco2021-oc}%
\begin{APACrefauthors}%
Ilharco, G.%
, Wortsman, M.%
, Wightman, R.%
, Gordon, C.%
, Carlini, N.%
, Taori, R.%
, Dave, A.%
, Shankar, V.%
, Namkoong, H.%
, Miller, J.%
, Hajishirzi, H.%
, Farhadi, A.%
\BCBL {}\ \BBA {} Schmidt, L.%
\end{APACrefauthors}%
\unskip\
\newblock
\APACrefYearMonthDay{2021}{}{}.
\newblock
\APACrefbtitle {{OpenCLIP} ({Version} 0.1)} {{OpenCLIP} ({Version} 0.1)}\
  [{Computer software}].
\newblock
\APACaddressPublisher{}{Zenodo}.
\newblock
\begin{APACrefDOI} \doi{10.5281/zenodo.5143773} \end{APACrefDOI}
\PrintBackRefs{\CurrentBib}

\bibitem [\protect \citeauthoryear {%
Krizhevsky%
}{%
Krizhevsky%
}{%
{\protect \APACyear {2009}}%
}]{%
Krizhevsky2009-ez}
\APACinsertmetastar {%
Krizhevsky2009-ez}%
\begin{APACrefauthors}%
Krizhevsky, A.%
\end{APACrefauthors}%
\unskip\
\newblock
\APACrefYearMonthDay{2009}{}{}.
\newblock
\APACrefbtitle {Learning multiple layers of features from tiny images}
  {Learning multiple layers of features from tiny images}\ \APACbVolEdTR
  {}{{Technical report}}.
\newblock
\APACaddressInstitution{}{University of Toronto}.
\PrintBackRefs{\CurrentBib}

\bibitem [\protect \citeauthoryear {%
Lloyd%
}{%
Lloyd%
}{%
{\protect \APACyear {1982}}%
}]{%
Lloyd1982-ls}
\APACinsertmetastar {%
Lloyd1982-ls}%
\begin{APACrefauthors}%
Lloyd, S\BPBI P.%
\end{APACrefauthors}%
\unskip\
\newblock
\APACrefYearMonthDay{1982}{}{}.
\newblock
{\BBOQ}\APACrefatitle {Least squares quantization in {PCM}} {Least squares
  quantization in {PCM}}.{\BBCQ}
\newblock
\APACjournalVolNumPages{{IEEE} Transactions on Information
  Theory}{28}{2}{129--137}.
\newblock
\begin{APACrefDOI} \doi{10.1109/TIT.1982.1056489} \end{APACrefDOI}
\PrintBackRefs{\CurrentBib}

\bibitem [\protect \citeauthoryear {%
Muttenthaler%
, Dippel%
\BCBL {}\ \protect \BOthers {.}}{%
Muttenthaler%
, Dippel%
\BCBL {}\ \protect \BOthers {.}}{%
{\protect \APACyear {2023}}%
}]{%
Muttenthaler2022-mf}
\APACinsertmetastar {%
Muttenthaler2022-mf}%
\begin{APACrefauthors}%
Muttenthaler, L.%
, Dippel, J.%
, Linhardt, L.%
, Vandermeulen, R\BPBI A.%
\BCBL {}\ \BBA {} Kornblith, S.%
\end{APACrefauthors}%
\unskip\
\newblock
\APACrefYearMonthDay{2023}{}{}.
\newblock
{\BBOQ}\APACrefatitle {Human alignment of neural network representations}
  {Human alignment of neural network representations}.{\BBCQ}
\newblock
\BIn{} \APACrefbtitle {{International Conference on Learning Representations}.}
  {{International Conference on Learning Representations}.}
\PrintBackRefs{\CurrentBib}

\bibitem [\protect \citeauthoryear {%
Muttenthaler%
\ \protect \BOthers {.}}{%
Muttenthaler%
\ \protect \BOthers {.}}{%
{\protect \APACyear {2025}}%
}]{%
Muttenthaler2025-nb}
\APACinsertmetastar {%
Muttenthaler2025-nb}%
\begin{APACrefauthors}%
Muttenthaler, L.%
, Greff, K.%
, Born, F.%
, Spitzer, B.%
, Kornblith, S.%
, Mozer, M\BPBI C.%
, M{\"u}ller, K\BHBI R.%
, Unterthiner, T.%
\BCBL {}\ \BBA {} Lampinen, A\BPBI K.%
\end{APACrefauthors}%
\unskip\
\newblock
\APACrefYearMonthDay{2025}{}{}.
\newblock
{\BBOQ}\APACrefatitle {Aligning machine and human visual representations across
  abstraction levels} {Aligning machine and human visual representations across
  abstraction levels}.{\BBCQ}
\newblock
\APACjournalVolNumPages{Nature}{647}{8089}{349--355}.
\newblock
\begin{APACrefDOI} \doi{10.1038/s41586-025-09631-6} \end{APACrefDOI}
\PrintBackRefs{\CurrentBib}

\bibitem [\protect \citeauthoryear {%
Muttenthaler%
, Linhardt%
\BCBL {}\ \protect \BOthers {.}}{%
Muttenthaler%
, Linhardt%
\BCBL {}\ \protect \BOthers {.}}{%
{\protect \APACyear {2023}}%
}]{%
Muttenthaler2023-aw}
\APACinsertmetastar {%
Muttenthaler2023-aw}%
\begin{APACrefauthors}%
Muttenthaler, L.%
, Linhardt, L.%
, Dippel, J.%
, Vandermeulen, R\BPBI A.%
, Hermann, K.%
, Lampinen, A\BPBI K.%
\BCBL {}\ \BBA {} Kornblith, S.%
\end{APACrefauthors}%
\unskip\
\newblock
\APACrefYearMonthDay{2023}{}{}.
\newblock
{\BBOQ}\APACrefatitle {Improving neural network representations using human
  similarity judgments} {Improving neural network representations using human
  similarity judgments}.{\BBCQ}
\newblock
\BIn{} \APACrefbtitle {{Advances in Neural Information Processing Systems}}
  {{Advances in Neural Information Processing Systems}}\ (\BVOL~36, \BPGS\
  50978--51007).
\newblock
\APACaddressPublisher{}{{Curran Associates}}.
\newblock
\begin{APACrefDOI} \doi{10.52202/075280-2218} \end{APACrefDOI}
\PrintBackRefs{\CurrentBib}

\bibitem [\protect \citeauthoryear {%
Pedregosa%
\ \protect \BOthers {.}}{%
Pedregosa%
\ \protect \BOthers {.}}{%
{\protect \APACyear {2011}}%
}]{%
Pedregosa2011-sk}
\APACinsertmetastar {%
Pedregosa2011-sk}%
\begin{APACrefauthors}%
Pedregosa, F.%
, Varoquaux, G.%
, Gramfort, A.%
, Michel, V.%
, Thirion, B.%
, Grisel, O.%
, Blondel, M.%
, Prettenhofer, P.%
, Weiss, R.%
, Dubourg, V.%
, Vanderplas, J.%
, Passos, A.%
, Cournapeau, D.%
, Brucher, M.%
, Perrot, M.%
\BCBL {}\ \BBA {} Duchesnay, {\'E}.%
\end{APACrefauthors}%
\unskip\
\newblock
\APACrefYearMonthDay{2011}{}{}.
\newblock
{\BBOQ}\APACrefatitle {Scikit-learn: Machine learning in {Python}}
  {Scikit-learn: Machine learning in {Python}}.{\BBCQ}
\newblock
\APACjournalVolNumPages{Journal of Machine Learning
  Research}{12}{}{2825--2830}.
\PrintBackRefs{\CurrentBib}

\bibitem [\protect \citeauthoryear {%
Radford%
\ \protect \BOthers {.}}{%
Radford%
\ \protect \BOthers {.}}{%
{\protect \APACyear {2021}}%
}]{%
Radford2021-fu}
\APACinsertmetastar {%
Radford2021-fu}%
\begin{APACrefauthors}%
Radford, A.%
, Kim, J\BPBI W.%
, Hallacy, C.%
, Ramesh, A.%
, Goh, G.%
, Agarwal, S.%
, Sastry, G.%
, Askell, A.%
, Mishkin, P.%
, Clark, J.%
, Krueger, G.%
\BCBL {}\ \BBA {} Sutskever, I.%
\end{APACrefauthors}%
\unskip\
\newblock
\APACrefYearMonthDay{2021}{}{}.
\newblock
{\BBOQ}\APACrefatitle {Learning transferable visual models from natural
  language supervision} {Learning transferable visual models from natural
  language supervision}.{\BBCQ}
\newblock
\BIn{} \APACrefbtitle {{Proceedings of the 38th International Conference on
  Machine Learning}} {{Proceedings of the 38th International Conference on
  Machine Learning}}\ (\BVOL~139, \BPGS\ 8748--8763).
\newblock
\APACaddressPublisher{}{{PMLR}}.
\PrintBackRefs{\CurrentBib}

\bibitem [\protect \citeauthoryear {%
Roads%
\ \BBA {} Love%
}{%
Roads%
\ \BBA {} Love%
}{%
{\protect \APACyear {2021}}%
}]{%
Roads2021-nz}
\APACinsertmetastar {%
Roads2021-nz}%
\begin{APACrefauthors}%
Roads, B\BPBI D.%
\BCBT {}\ \BBA {} Love, B\BPBI C.%
\end{APACrefauthors}%
\unskip\
\newblock
\APACrefYearMonthDay{2021}{}{}.
\newblock
{\BBOQ}\APACrefatitle {Enriching {ImageNet} with human similarity judgments and
  psychological embeddings} {Enriching {ImageNet} with human similarity
  judgments and psychological embeddings}.{\BBCQ}
\newblock
\BIn{} \APACrefbtitle {{2021 IEEE/CVF Conference on Computer Vision and Pattern
  Recognition}} {{2021 IEEE/CVF Conference on Computer Vision and Pattern
  Recognition}}\ (\BPGS\ 3546--3556).
\newblock
\APACaddressPublisher{}{{IEEE}}.
\newblock
\begin{APACrefDOI} \doi{10.1109/CVPR46437.2021.00355} \end{APACrefDOI}
\PrintBackRefs{\CurrentBib}

\bibitem [\protect \citeauthoryear {%
Snell%
\ \protect \BOthers {.}}{%
Snell%
\ \protect \BOthers {.}}{%
{\protect \APACyear {2017}}%
}]{%
Snell2017-pn}
\APACinsertmetastar {%
Snell2017-pn}%
\begin{APACrefauthors}%
Snell, J.%
, Swersky, K.%
\BCBL {}\ \BBA {} Zemel, R\BPBI S.%
\end{APACrefauthors}%
\unskip\
\newblock
\APACrefYearMonthDay{2017}{}{}.
\newblock
{\BBOQ}\APACrefatitle {Prototypical networks for few-shot learning}
  {Prototypical networks for few-shot learning}.{\BBCQ}
\newblock
\BIn{} \APACrefbtitle {{Advances in Neural Information Processing Systems}}
  {{Advances in Neural Information Processing Systems}}\ (\BVOL~30, \BPGS\
  4077--4087).
\newblock
\APACaddressPublisher{}{{Curran Associates}}.
\PrintBackRefs{\CurrentBib}

\bibitem [\protect \citeauthoryear {%
{TorchVision maintainers and contributors}%
}{%
{TorchVision maintainers and contributors}%
}{%
{\protect \APACyear {2016}}%
}]{%
Torchvision2016-tv}
\APACinsertmetastar {%
Torchvision2016-tv}%
\begin{APACrefauthors}%
{TorchVision maintainers and contributors}.%
\end{APACrefauthors}%
\unskip\
\newblock
\APACrefYearMonthDay{2016}{}{}.
\newblock
\APACrefbtitle {{TorchVision}: {PyTorch}'s computer vision library ({Version}
  0.17.2)} {{TorchVision}: {PyTorch}'s computer vision library ({Version}
  0.17.2)}\ [{Computer software}].
\newblock
\APACaddressPublisher{}{{GitHub}}.
\newblock
\begin{APACrefURL} [{{August 26, 2026}}]\url{https://github.com/pytorch/vision}
  \end{APACrefURL}
\PrintBackRefs{\CurrentBib}

\bibitem [\protect \citeauthoryear {%
Wightman%
}{%
Wightman%
}{%
{\protect \APACyear {2019}}%
}]{%
Wightman2019-tm}
\APACinsertmetastar {%
Wightman2019-tm}%
\begin{APACrefauthors}%
Wightman, R.%
\end{APACrefauthors}%
\unskip\
\newblock
\APACrefYearMonthDay{2019}{}{}.
\newblock
\APACrefbtitle {{PyTorch} image models ({Version} 0.9.16)} {{PyTorch} image
  models ({Version} 0.9.16)}\ [{Computer software}].
\newblock
\APACaddressPublisher{}{Zenodo}.
\newblock
\begin{APACrefDOI} \doi{10.5281/zenodo.4414861} \end{APACrefDOI}
\PrintBackRefs{\CurrentBib}

\bibitem [\protect \citeauthoryear {%
Wolf%
\ \protect \BOthers {.}}{%
Wolf%
\ \protect \BOthers {.}}{%
{\protect \APACyear {2020}}%
}]{%
Wolf2020-hf}
\APACinsertmetastar {%
Wolf2020-hf}%
\begin{APACrefauthors}%
Wolf, T.%
, Debut, L.%
, Sanh, V.%
, Chaumond, J.%
, Delangue, C.%
, Moi, A.%
, Cistac, P.%
, Rault, T.%
, Louf, R.%
, Funtowicz, M.%
, Davison, J.%
, Shleifer, S.%
, von Platen, P.%
, Ma, C.%
, Jernite, Y.%
, Plu, J.%
, Xu, C.%
, Le~Scao, T.%
, Gugger, S.%
\BDBL {}Rush, A\BPBI M.%
\end{APACrefauthors}%
\unskip\
\newblock
\APACrefYearMonthDay{2020}{}{}.
\newblock
{\BBOQ}\APACrefatitle {Transformers: State-of-the-art natural language
  processing} {Transformers: State-of-the-art natural language
  processing}.{\BBCQ}
\newblock
\BIn{} \APACrefbtitle {{Proceedings of the 2020 Conference on Empirical Methods
  in Natural Language Processing: System Demonstrations}} {{Proceedings of the
  2020 Conference on Empirical Methods in Natural Language Processing: System
  Demonstrations}}\ (\BPGS\ 38--45).
\newblock
\APACaddressPublisher{}{{Association for Computational Linguistics}}.
\newblock
\begin{APACrefDOI} \doi{10.18653/v1/2020.emnlp-demos.6} \end{APACrefDOI}
\PrintBackRefs{\CurrentBib}

\end{thebibliography}
\clearpage

\setcounter{section}{0}
\setcounter{figure}{0}
\setcounter{table}{0}
\renewcommand{\thesection}{S\arabic{section}}
\renewcommand{\thefigure}{S\arabic{figure}}
\renewcommand{\thetable}{S\arabic{table}}
\renewcommand{\theHsection}{supp.\arabic{section}}
\renewcommand{\theHfigure}{supp.\arabic{figure}}
\renewcommand{\theHtable}{supp.\arabic{table}}

\clearpage

\setcounter{page}{1}
\resetlinenumber

\thispagestyle{empty}
\begin{center}
{\large\bfseries Supplementary Information for}\\[1.5em]
{\large Relational Knowledge Distillation Brings\\
DNN Representations Close Enough to Humans\\
to Be Aligned Without Supervision}\\[1.5em]
Yuria Shimizu, Soh Takahashi, Takato Horii, and Masafumi Oizumi
\end{center}

\clearpage

\section*{Table of Contents}
\makeatletter
\newcommand{\suptocline}[3]{%
  \par\begingroup
    \leftskip #1\relax
    \renewcommand{\@pnumwidth}{1.3em}%
    \rightskip \@tocrmarg
    \parfillskip -\rightskip
    \noindent #2\nobreak
    \leaders\hbox{$\m@th\mkern\@dotsep mu\hbox{.}\mkern\@dotsep mu$}\hfill
    \nobreak\hb@xt@\@pnumwidth{\hfil\normalfont\normalcolor\pageref{#3}}%
    \par
  \endgroup}
\makeatother
\suptocline{0pt}{\textbf{\ref*{sec:supp_data_driven}.~Coarse-category-level analyses using data-driven clusters}}{sec:supp_data_driven}
\suptocline{1.5em}{Fig.~\ref*{fig:data_driven_clustering}.~GWOT matching accuracy and inter-cluster arrangement for the data-driven clusters, before and after RKD}{fig:data_driven_clustering}

\suptocline{0pt}{\textbf{\ref*{sec:supp_additional}.~RKD applied to DNNs with different architectures and pre-training schemes}}{sec:supp_additional}
\suptocline{1.5em}{Table~\ref*{tab:rsa_gwot_additional}.~RSA and GWOT results for seven additional DNN models before and after RKD}{tab:rsa_gwot_additional}

\suptocline{0pt}{\textbf{\ref*{sec:supp_split}.~Control experiment: a within-dataset split of THINGS}}{sec:supp_split}
\suptocline{1.5em}{Table~\ref*{tab:rsa_gwot_things_split_ep50}.~RSA and GWOT results for five CLIP ViT-B/16 variants under the within-dataset split of THINGS}{tab:rsa_gwot_things_split_ep50}

\suptocline{0pt}{\textbf{\ref*{sec:supp_neighborhoods}.~Reorganization of the DNN's own local neighborhoods after RKD}}{sec:supp_neighborhoods}
\suptocline{1.5em}{Table~\ref*{tab:knn_before_after}.~$k$-nearest-neighbor overlap rate between the pre-RKD and post-RKD CLIP ViT-B/16 representations}{tab:knn_before_after}

\suptocline{0pt}{\textbf{\ref*{sec:supp_fewshot}.~Few-shot classification performance before and after RKD}}{sec:supp_fewshot}
\suptocline{1.5em}{Table~\ref*{tab:fewshot_results}.~Few-shot classification accuracy for the five CLIP ViT-B/16 variants before and after RKD}{tab:fewshot_results}

\suptocline{0pt}{\textbf{Supplementary References}}{sec:supp_references}
\clearpage

\section{Coarse-category-level analyses using data-driven clusters}\label{sec:supp_data_driven}

    To confirm that our coarse-category-level results did not depend on the human-annotated THINGS categories \citepsupp{Hebart2019-bn}, we conducted the same evaluations on data-driven clusters derived from the human embedding. These evaluations concern the coarse-category-level GWOT matching accuracy (Section~\ref{sec:gwot_results}) and the arrangement of the coarse categories relative to one another (Section~\ref{sec:category}).

\begin{figure}[p]
                \centering
                \includegraphics[width=0.879\textwidth]{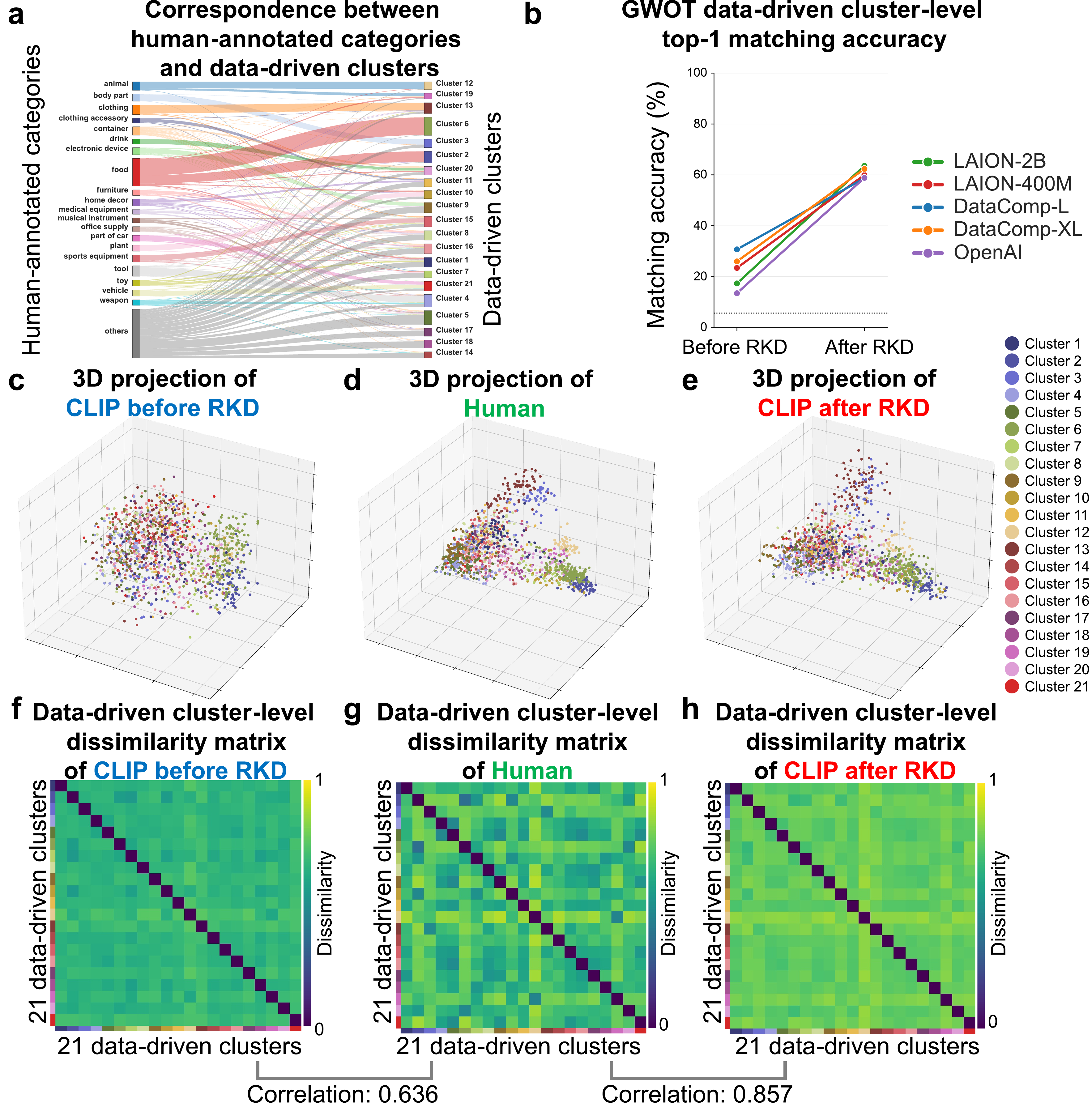}
                \caption{\textbf{Data-driven cluster-level GWOT matching accuracy and inter-cluster arrangement, before and after RKD.}
                \textbf{(a)} Sankey diagram relating the 21 human-annotated THINGS categories, including \emph{others}, to the 21 data-driven clusters. The categories on the left are ordered alphabetically, with \emph{others} last. Each cluster on the right is then placed at the average position of the categories to which its objects belong, which reduces the crossing of the flows. Flow widths are proportional to the number of objects.
                \textbf{(b)} Data-driven cluster-level top-1 GWOT matching accuracy, before and after RKD, for the five CLIP ViT-B/16 variants labeled in the panel. The dotted line marks chance ($5.8\%$).
                \textbf{(c--e)} Three-dimensional projections of the 1{,}249 test objects for CLIP ViT-B/16 (LAION-2B) as an example, with each object colored by its data-driven cluster: CLIP before RKD (c), human psychological embedding (d), and CLIP after RKD (e).
                \textbf{(f--h)} $21 \times 21$ inter-cluster dissimilarity matrices for the same model: CLIP before RKD (f), human psychological embedding (g), and CLIP after RKD (h). The color bars along the left and bottom edges indicate cluster identity, shown in the legend at the right of (c--e). Brackets below the matrices indicate Spearman correlations between the CLIP and human inter-cluster dissimilarity matrices before and after RKD.
                }
                \label{fig:data_driven_clustering}
            \end{figure}

    To obtain the data-driven clusters, we applied $k$-means clustering \citepsupp{Lloyd1982-ls} to the 66-dimensional human psychological embedding of the 1{,}249 THINGS test objects \citepsupp{Hebart2023-uf}, using scikit-learn \citepsupp{Pedregosa2011-sk}. We set the number of clusters to 21, matching the number of human-annotated groups (the 20 THINGS coarse categories plus \emph{others}), so that the data-driven clusters have the same granularity as the human-annotated categories. As the Sankey diagram in Fig.~\ref{fig:data_driven_clustering}a shows, the resulting 21 clusters overlap partially, but not one-to-one, with the human-annotated categories. Thus, they constitute a different grouping of the same objects. Since these clusters include no heterogeneous \emph{others} category, we used all 1{,}249 test objects, rather than the 794 objects in 20 categories used for the human-annotated arrangement analyses (Section~\ref{sec:pca}, Section~\ref{sec:inter_category}).

    First, to repeat the analysis of Section~\ref{sec:gwot_results} for the data-driven clusters, we rescored the GWOT transport plans using these clusters.

    We found that data-driven cluster-level top-1 matching accuracy increased after RKD for all five variants (Fig.~\ref{fig:data_driven_clustering}b). It rose from a pre-RKD range of $13.5\%$--$30.7\%$ to a post-RKD range of $58.7\%$--$63.6\%$ (chance $5.8\%$). This post-RKD range is comparable to that obtained with the human-annotated categories ($57.6\%$--$61.2\%$; Table~\ref{tab:rsa_gwot} of the main article). We therefore conclude that the coarse-category-level gain reported in Section~\ref{sec:gwot_results} does not depend on the human-annotated grouping.

    Second, to repeat the analyses of Section~\ref{sec:pca} and Section~\ref{sec:inter_category} for the data-driven clusters, we conducted the same two analyses on the 21 clusters. These are the three-dimensional projection of the human, pre-RKD, and post-RKD embeddings, and the Spearman correlation between the human and CLIP $21 \times 21$ inter-cluster dissimilarity matrices. We followed the same procedures as in those sections.

    We found that both analyses yielded the same pattern of results as that obtained with the human-annotated categories. In the projections, which we show for CLIP ViT-B/16 (LAION-2B) as an example, the post-RKD clusters were arranged in much the same way as in the human embedding, whereas the pre-RKD clusters were not (Fig.~\ref{fig:data_driven_clustering}c--e, in which each object is colored by its data-driven cluster). The inter-cluster Spearman correlation increased after RKD for all five variants, from a pre-RKD range of $\rho = 0.636$--$0.771$ to a post-RKD range of $\rho = 0.846$--$0.874$. For the same example model, it rose from $\rho = 0.636$ to $\rho = 0.857$ (Fig.~\ref{fig:data_driven_clustering}f--h). These values are close to those obtained with the human-annotated categories ($\rho = 0.572$--$0.751$ before RKD and $\rho = 0.838$--$0.879$ after RKD; Section~\ref{sec:category}). We therefore conclude that the results on coarse-category arrangement reported in Section~\ref{sec:category} do not depend on the human-annotated grouping.

\section{RKD applied to DNNs with different architectures and pre-training schemes}\label{sec:supp_additional}

    To test whether the improvement in human--DNN alignment after RKD was specific to the CLIP ViT-B/16 architecture and to CLIP-style image--text pre-training, we applied the same RKD procedure and the same RSA and GWOT evaluations to seven additional DNN models. We selected these models so that together they varied both the architecture (convolutional ResNet versus Vision Transformer) and the pre-training scheme (image--text contrastive, supervised classification, self-supervised, and no pre-training). We varied the pre-training scheme more widely, because prior work reports that the training objective and the training data affect human--DNN alignment strongly, whereas the architecture and the model scale have comparatively little effect \citepsupp{Muttenthaler2022-mf}.

    We used seven models that span two architectures and four pre-training schemes, and we obtained every pre-trained weight set from its public release. The first was CLIP ResNet-50 (OpenAI pre-training; \citealpsupp{Radford2021-fu}), for which we used OpenCLIP \citepsupp{Ilharco2021-oc, Cherti2022-fl}. The next two were ResNet-50 models \citepsupp{He2016-je} pre-trained on ImageNet-1k \citepsupp{Deng2009-ea}, one by supervised classification, for which we used torchvision \citepsupp{Torchvision2016-tv}, and one by self-supervised SimCLRv2, for which we used the SimCLRv2 release \citepsupp{Chen2020-sv}. The next three were ViT-B/16 models: two pre-trained on ImageNet-1k by self-supervised DINO \citepsupp{Caron2021-dn} and by self-supervised MAE \citepsupp{He2022-ma}, for which we used \texttt{timm} \citepsupp{Wightman2019-tm}, and one pre-trained by supervised classification (ImageNet-21k \citepsupp{Deng2009-ea} followed by ImageNet-1k), for which we used Hugging Face Transformers \citepsupp{Wolf2020-hf}. The last was a randomly initialized ViT-B/16 with no pre-training. We took the representation of each model from its final layer: the vision encoder projection output for CLIP ResNet-50, the globally average-pooled output for the two ResNet-50 models, the class token for the DINO, supervised, and randomly initialized ViT-B/16 models, and the mean of the patch tokens for the MAE ViT-B/16. These representations range from 768 to 2{,}048 dimensions, compared with 512 for CLIP ViT-B/16.

    We implemented RKD, and the RSA and GWOT evaluations, following the same procedures as for the five CLIP ViT-B/16 variants. For RKD, we distilled the same human teacher embedding into each model, using the same ImageNet training images, relational objective, and training schedule (50 epochs) as in Section~\ref{sec:rkd_training_setup}, with the learning rate selected per model by the same validation procedure, which yielded $1 \times 10^{-6}$ for all seven models. For the RSA and GWOT evaluations, we compared the human and DNN representations before and after RKD on the 1{,}249-object THINGS test set, following the procedures of Section~\ref{sec:methods_rsa} and Section~\ref{sec:methods_gwot}, including the GWOT hyperparameter search.

    We found that RKD was effective for every model except the randomly initialized ViT-B/16, although the size of the gain varied widely across models (Table~\ref{tab:rsa_gwot_additional}). For the six pre-trained models, the RSA correlation and the GWOT matching accuracy both rose after RKD, and for most of them the post-RKD values were comparable to those of the five CLIP ViT-B/16 variants. Notably, for CLIP ResNet-50, the two measures did not move together: the RSA correlation barely changed after RKD ($\rho = 0.500$ to $\rho = 0.501$), whereas the model's GWOT matching accuracy rose substantially at both levels of granularity. The supervised comparison alone can therefore miss a reorganization that the unsupervised comparison detects. The gain was smallest for the self-supervised MAE ViT-B/16, whose pre-trained representation was the most weakly aligned with humans of the six. Both measures rose for this model, but its gain in GWOT matching accuracy was far smaller than those of the other five.

    For the randomly initialized ViT-B/16, we found that alignment also improved after RKD, but that it remained close to chance. Its individual-object top-1 matching accuracy rose from $0.00\%$ to only $0.160\%$ against a chance level of $0.0801\%$, and its coarse-category top-1 matching accuracy reached $20.7\%$ against a chance level of $18.4\%$.

    We conclude that alignment improved after RKD for every model across architectures and pre-training schemes, but that the pre-trained representation needs to be aligned with humans to some extent for RKD to produce a substantial gain. 

\begin{table}[t]
            \centering
            \caption{\textbf{RSA and GWOT results for seven additional DNN models before and after RKD on the 1{,}249-object THINGS test set.} Values in the RSA column are Spearman correlations ($\rho$) between human and DNN dissimilarity matrices. Values in the GWOT columns are matching accuracies. For each human object, the top-1 and top-5 columns consider, respectively, the single DNN object and the five DNN objects that receive the most transport mass from it. Fine-grained (individual-object-level) matching counts only the identical object as correct, whereas coarse-grained (coarse-category-level) matching counts any object from the same coarse category as correct. Chance-level values are listed for reference.}
            \label{tab:rsa_gwot_additional}
            \scriptsize
            \setlength{\tabcolsep}{2.2pt}
            \setgwotcolwd{Coarse-grained}{GWOT matching accuracy (\%)}
            \begin{tabular}{@{}llllS[table-format=1.4]cccc}
                \toprule
                \multirow{3}{*}{Architecture} & \multirow{3}{*}{\makecell[l]{Pre-training\\data}} & \multirow{3}{*}{\makecell[l]{Learning\\paradigm}} & & {\multirow{3}{*}{\makecell{RSA correlation\\(Spearman $\rho$)}}} & \multicolumn{4}{c}{GWOT matching accuracy (\%)} \\
                \cmidrule(lr){6-9}
                & & & & & \multicolumn{2}{c}{Fine-grained} & \multicolumn{2}{c}{Coarse-grained} \\
                \cmidrule(lr){6-7} \cmidrule(lr){8-9}
                & & & & & \gwotcell{Top-1} & \gwotcell{Top-5} & \gwotcell{Top-1} & \gwotcell{Top-5} \\
                \midrule
                \multirow{2}{*}{ResNet-50} & \multirow{2}{*}{OpenAI} & \multirow{2}{*}{\makecell[l]{Image--text\\contrastive (CLIP)}}
                    & Before & 0.500 & 0.400 & 1.52 & 36.7 & 61.2 \\
                    & & & After  & 0.501 & 10.3 & 31.7 & 53.8 & 87.3 \\
                \midrule
                \multirow{2}{*}{ResNet-50} & \multirow{2}{*}{ImageNet-1k} & \multirow{2}{*}{Supervised}
                    & Before & 0.351 & 1.92 & 8.17 & 38.8 & 71.8 \\
                    & & & After  & 0.458 & 4.40 & 15.3 & 45.1 & 78.1 \\
                \midrule
                \multirow{2}{*}{ResNet-50} & \multirow{2}{*}{ImageNet-1k} & \multirow{2}{*}{\makecell[l]{Self-supervised\\(SimCLRv2)}}
                    & Before & 0.316 & 0.320 & 1.44 & 28.9 & 55.8 \\
                    & & & After  & 0.471 & 5.12 & 16.7 & 44.8 & 79.2 \\
                \midrule
                \multirow{2}{*}{ViT-B/16} & \multirow{2}{*}{ImageNet-1k} & \multirow{2}{*}{\makecell[l]{Self-supervised\\(DINO)}}
                    & Before & 0.439 & 0.400 & 2.48 & 35.7 & 66.0 \\
                    & & & After  & 0.488 & 5.84 & 20.3 & 49.2 & 83.9 \\
                \midrule
                \multirow{2}{*}{ViT-B/16} & \multirow{2}{*}{ImageNet-1k} & \multirow{2}{*}{\makecell[l]{Self-supervised\\(MAE)}}
                    & Before & 0.138 & 0.00 & 0.400 & 15.2 & 41.6 \\
                    & & & After  & 0.291 & 0.560 & 2.08 & 34.7 & 66.4 \\
                \midrule
                \multirow{2}{*}{ViT-B/16} & \multirow{2}{*}{\makecell[l]{ImageNet-\\21k$\to$1k}} & \multirow{2}{*}{Supervised}
                    & Before & 0.308 & 0.560 & 3.92 & 35.1 & 65.6 \\
                    & & & After  & 0.575 & 16.6 & 39.1 & 59.4 & 90.6 \\
                \midrule
                \multirow{2}{*}{ViT-B/16} & \multirow{2}{*}{None} & \multirow{2}{*}{\makecell[l]{None\\(random init.)}}
                    & Before & 0.0627 & 0.00 & 0.560 & 16.7 & 46.4 \\
                    & & & After  & 0.102  & 0.160  & 0.721 & 20.7 & 55.8 \\
                \midrule
                Chance rate
                    & & & & {N/A} & 0.0801 & 0.400 & 18.4 & 51.9 \\
                \bottomrule
            \end{tabular}%
        \end{table}

\section{Control experiment: a within-dataset split of THINGS}\label{sec:supp_split}

    To examine how demanding the generalization required by our primary evaluation (Section~\ref{sec:results_alignment}) is, we conducted a control experiment using a within-dataset split of the THINGS dataset \citepsupp{Hebart2019-bn}. Our primary evaluation was demanding in two respects. First, the human embedding used for RKD and the human embedding used for evaluation were derived from different judgment tasks on different image datasets and by different embedding models (the 8-rank-2 task on ImageNet images, \citealpsupp{Roads2021-nz}, versus the triplet odd-one-out task on THINGS images, \citealpsupp{Hebart2020-pd, Hebart2023-uf}; see Section~\ref{sec:dataset_imagenet} and Section~\ref{sec:dataset_things}). Second, the test concepts were curated so as not to overlap with the RKD training concepts (Section~\ref{sec:dataset_things}). Therefore, as a control, we conducted the same procedure without either of these two demands, using a conventional split of the THINGS objects into training and test sets.

    We conducted the control experiment as follows. We sorted the 1{,}854 THINGS objects alphabetically and assigned the first 1{,}554 to the training set and the remaining 300 to the test set. We fine-tuned each of the five CLIP ViT-B/16 variants on the 1{,}554 objects, with the same relational objective and optimization hyperparameters as in our primary RKD experiments. We then compared the human and DNN representations of the 300 held-out objects with the same RSA and GWOT procedures as in Section~\ref{sec:methods_rsa} and Section~\ref{sec:methods_gwot}. We used the same 21-category assignment (20 categories plus \emph{others}), which was defined once on all 1{,}854 objects and applied here to the 300 test objects.

    We found that the RSA correlation and the GWOT matching accuracy rose after RKD for all five CLIP ViT-B/16 variants, and that the gains were larger than in our primary evaluation (Table~\ref{tab:rsa_gwot_things_split_ep50}). Before RKD, the RSA correlation ranged from $\rho = 0.448$ to $\rho = 0.464$, GWOT fine-grained top-1 matching accuracy from $0.00\%$ to $1.33\%$, and GWOT coarse-grained top-1 matching accuracy from $16.0\%$ to $24.7\%$. After RKD, these three measures reached $\rho = 0.803$--$0.825$, $58.0\%$--$65.0\%$, and $77.7\%$--$84.7\%$, respectively.

    The two settings use different test sets, so we do not compare their values directly. Even so, every measure was substantially higher under the within-dataset split than under our primary evaluation. This was especially marked for GWOT fine-grained top-1 matching accuracy, the measure of primary interest in this study: $58.0\%$--$65.0\%$ versus $13.1\%$--$19.5\%$ (Table~\ref{tab:rsa_gwot} of the main article). We therefore conclude that our primary evaluation is a far stricter generalization test.

    \begin{table}[t]
                \centering
                \caption{\textbf{RSA and GWOT results for five CLIP ViT-B/16 variants before and after RKD under the within-dataset split of THINGS.} Values in the RSA column are Spearman correlations ($\rho$) between human and DNN dissimilarity matrices. Values in the GWOT columns are matching accuracies. For each human object, the top-1 and top-5 columns consider, respectively, the single DNN object and the five DNN objects that receive the most transport mass from it. Fine-grained (individual-object-level) matching counts only the identical object as correct, whereas coarse-grained (coarse-category-level) matching counts any object from the same coarse category as correct. Chance-level values are listed for reference.}
                \label{tab:rsa_gwot_things_split_ep50}
                \small
                \setlength{\tabcolsep}{4pt}
                \setgwotcolwd{Coarse-grained}{GWOT matching accuracy (\%)}
                \begin{tabular}{@{}llRcccc}
                    \toprule
                    \multirow{3}{*}{Pre-training data} & & {\multirow{3}{*}{\makecell{RSA correlation\\(Spearman $\rho$)}}} & \multicolumn{4}{c}{GWOT matching accuracy (\%)} \\
                    \cmidrule(lr){4-7}
                    & & & \multicolumn{2}{c}{Fine-grained} & \multicolumn{2}{c}{Coarse-grained} \\
                    \cmidrule(lr){4-5} \cmidrule(lr){6-7}
                    & & & \gwotcell{Top-1} & \gwotcell{Top-5} & \gwotcell{Top-1} & \gwotcell{Top-5} \\
                    \midrule
                    \multirow{2}{*}{LAION-2B}
                        & Before & 0.451 & 1.33 & 4.00 & 23.7 & 47.7 \\
                        & After  & 0.822 & 65.0 & 92.7 & 84.7 & 99.0 \\
                    \midrule
                    \multirow{2}{*}{LAION-400M}
                        & Before & 0.460 & 0.00 & 2.67 & 16.0 & 43.0 \\
                        & After  & 0.803 & 62.3 & 87.3 & 80.7 & 97.7 \\
                    \midrule
                    \multirow{2}{*}{DataComp-L}
                        & Before & 0.448 & 0.667 & 2.00 & 21.7 & 46.0 \\
                        & After  & 0.805 & 63.0 & 86.7 & 78.0 & 97.3 \\
                    \midrule
                    \multirow{2}{*}{DataComp-XL}
                        & Before & 0.449 & 0.667 & 5.00 & 24.7 & 50.7 \\
                        & After  & 0.825 & 58.0 & 90.0 & 77.7 & 97.7 \\
                    \midrule
                    \multirow{2}{*}{OpenAI}
                        & Before & 0.464 & 0.333 & 2.67 & 16.0 & 41.7 \\
                        & After  & 0.823 & 60.3 & 89.7 & 82.3 & 98.3 \\
                    \midrule
                    Chance rate
                        & & {N/A} & 0.333 & 1.67 & 15.0 & 45.8 \\
                    \bottomrule
                \end{tabular}%
            \end{table}

\section{Reorganization of the DNN's own local neighborhoods after RKD}\label{sec:supp_neighborhoods}

    Motivated by the finding that the human--DNN nearest-neighbor overlap rate remained largely unchanged after RKD (Section~\ref{sec:knn}), we investigated how much RKD changed the DNN's own local neighborhoods. We applied the same procedure to compare the CLIP representations before and after RKD, independently of the human reference (see Section~\ref{sec:methods_knn} for details). The two sets of neighbors compared here are the pre-RKD and post-RKD CLIP neighbors, rather than the human and DNN neighbors. For each of the 1{,}249 THINGS test objects, we computed the proportion of its $k$ nearest neighbors in the pre-RKD CLIP embedding that remained among its $k$ nearest neighbors in the post-RKD embedding. We then averaged this proportion across all objects for each of the five CLIP ViT-B/16 variants, and we report the mean together with a 95\% confidence interval.

    We found that RKD replaced more than half of each object's nearest neighbors on average. The mean proportion retained after RKD was $33.8\%$--$44.6\%$ at $k = 5$ and $35.0\%$--$45.5\%$ at $k = 10$ across the five variants (Table~\ref{tab:knn_before_after}). Therefore, RKD substantially reorganized the DNN's own local neighborhoods, even though the human--DNN nearest-neighbor overlap rate shown in Section~\ref{sec:knn} remained largely unchanged.

    \begin{table}[t]
                \centering
                \caption{\textbf{$k$-nearest-neighbor overlap rate between the pre-RKD and post-RKD CLIP ViT-B/16 representations on the 1{,}249-object THINGS test set.} For each object, the value is the percentage of its $k$ nearest neighbors that are shared between the pre-RKD and post-RKD embeddings. Values are the mean across all 1{,}249 objects, shown with the 95\% confidence interval (CI).}
                \label{tab:knn_before_after}
                \small
                \setlength{\tabcolsep}{6pt}
                \begin{tabular}{@{}l>{\centering\arraybackslash}p{4.15cm}>{\centering\arraybackslash}p{4.15cm}}
                \toprule
                \multirow{2}{*}{Pre-training data} & \multicolumn{2}{c}{Nearest-neighbor overlap rate (\%, mean $\pm$ 95\% CI)} \\
                \cmidrule(lr){2-3}
                & {$k=5$} & {$k=10$} \\
                \midrule
                LAION-2B    & $41.9 \pm 1.3$ & $42.8 \pm 1.0$ \\
                LAION-400M  & $41.7 \pm 1.3$ & $42.9 \pm 1.0$ \\
                DataComp-L  & $44.6 \pm 1.3$ & $45.5 \pm 1.0$ \\
                DataComp-XL & $40.8 \pm 1.2$ & $41.0 \pm 1.0$ \\
                OpenAI      & $33.8 \pm 1.2$ & $35.0 \pm 1.0$ \\
                \bottomrule
                \end{tabular}
            \end{table}

\section{Few-shot classification performance before and after RKD}\label{sec:supp_fewshot}

    To examine how RKD affected DNNs' downstream task performance, we measured accuracy on few-shot classification, a representative example of a widely used downstream task, before and after RKD. As stated in Section~\ref{sec:methods_rkd}, we optimized the relational objective alone, with no task-specific loss and no regularization term; this analysis quantifies the consequence of that design.

    We conducted the few-shot classification analysis in the following steps. First, we built class prototypes for each dataset by averaging the features of five randomly selected examples per class. Second, we assigned each remaining image to the prototype with the smallest dissimilarity (defined in Eq.~\ref{eq:combined_distance} of the main article). Third, we repeated this random selection five times per dataset, using the same five selections before and after RKD so that the comparison was paired, and we report the mean and standard deviation (SD) of accuracy across them. Throughout, we left the representations unchanged and fitted no parameters for this task, classifying images by prototype matching \citepsupp{Snell2017-pn} rather than by fitting a classifier head.

    The four datasets used in the evaluation differ in their relation to our training and evaluation domains. THINGS (1{,}854 classes; \citealpsupp{Hebart2019-bn}) is the domain used for the human--DNN alignment evaluation. The ILSVRC 2012 validation set of ImageNet (1{,}000 classes; \citealpsupp{Deng2009-ea}) is the image set used for RKD training (Section~\ref{sec:dataset_imagenet}). CIFAR-100 (100 classes) and its coarse-label variant CIFAR-20 (20 classes) \citepsupp{Krizhevsky2009-ez} are out of domain for both RKD training and the human--DNN alignment evaluation.

    \begin{table}[t]
            \centering
            \caption{\textbf{Few-shot classification accuracy for the five CLIP ViT-B/16 variants before and after RKD.} Values are accuracy (\%), mean $\pm$ SD across five random draws of the five support examples per class, reported for four datasets: THINGS, ImageNet, CIFAR-100, and its coarse-label variant CIFAR-20. The same five draws were used before and after RKD. Chance-level values are listed for reference.}
            \label{tab:fewshot_results}
            \small
            \setlength{\tabcolsep}{6pt}
            \begin{tabular}{@{}llcccc@{}}
                \toprule
                \multirow{2}{*}{Pre-training data} & & \multicolumn{4}{c}{Few-shot accuracy (\%, mean $\pm$ SD)} \\
                \cmidrule(lr){3-6}
                & & THINGS & ImageNet & CIFAR-100 & CIFAR-20 \\
                \midrule
                \multirow{2}{*}{LAION-2B}
                    & Before & $87.8 \pm 0.3$ & $61.8 \pm 0.2$ & $67.4 \pm 0.7$ & $65.1 \pm 1.3$ \\
                    & After  & $84.9 \pm 0.3$ & $53.5 \pm 0.3$ & $55.1 \pm 0.7$ & $58.1 \pm 1.4$ \\
                \midrule
                \multirow{2}{*}{LAION-400M}
                    & Before & $86.7 \pm 0.3$ & $58.7 \pm 0.5$ & $61.1 \pm 0.3$ & $60.0 \pm 1.2$ \\
                    & After  & $82.2 \pm 0.1$ & $49.9 \pm 0.1$ & $50.1 \pm 0.5$ & $54.3 \pm 0.6$ \\
                \midrule
                \multirow{2}{*}{DataComp-L}
                    & Before & $85.6 \pm 0.4$ & $55.6 \pm 0.3$ & $66.4 \pm 0.5$ & $65.6 \pm 0.9$ \\
                    & After  & $81.3 \pm 0.2$ & $46.6 \pm 0.3$ & $57.1 \pm 0.4$ & $60.2 \pm 0.6$ \\
                \midrule
                \multirow{2}{*}{DataComp-XL}
                    & Before & $90.1 \pm 0.3$ & $64.1 \pm 0.3$ & $73.4 \pm 0.5$ & $70.3 \pm 0.6$ \\
                    & After  & $87.4 \pm 0.2$ & $55.5 \pm 0.3$ & $64.8 \pm 0.6$ & $65.7 \pm 1.6$ \\
                \midrule
                \multirow{2}{*}{OpenAI}
                    & Before & $81.0 \pm 0.5$ & $56.1 \pm 0.3$ & $56.8 \pm 0.6$ & $58.7 \pm 1.6$ \\
                    & After  & $81.0 \pm 0.3$ & $53.7 \pm 0.3$ & $56.5 \pm 0.8$ & $61.6 \pm 0.6$ \\
                \midrule
                Chance rate
                    & & $0.0539$ & $0.100$ & $1.00$ & $5.00$ \\
                \bottomrule
            \end{tabular}%
        \end{table}

    We found that RKD reduced few-shot accuracy across models and datasets, but that the reductions were modest (Table~\ref{tab:fewshot_results}). For the four variants other than the one pre-trained on the OpenAI dataset, the reductions were largest on CIFAR-100 and on ImageNet (8.6--12.3 and 8.3--9.0 percentage points, respectively), and smallest on THINGS (2.7--4.5 percentage points). These four variants still classified THINGS, the dataset with the largest number of classes, with high accuracy ($85.6\%$--$90.1\%$ before RKD and $81.3\%$--$87.4\%$ after). We regard the reduction on ImageNet as notable, because we performed RKD on these very images. The CLIP ViT-B/16 pre-trained on the OpenAI dataset was the exception to this downward trend: its accuracy was essentially unchanged on THINGS and CIFAR-100, rose on CIFAR-20, and fell only modestly on ImageNet.

    We conclude that equipping a DNN with human relational structure does not always benefit task performance, at least for the few-shot classification accuracy we examined. Pre-trained CLIP models have been reported to achieve high few-shot classification accuracy, one of the standard performance benchmarks in the computer vision field \citepsupp{Cherti2022-fl}. The RKD objective rewarded only agreement with the human relational structure and contained no term that anchored the student to that pre-trained structure, so RKD could overwrite it without penalty. Indeed, prior work reported a similar few-shot degradation when representations were aligned to human judgments without an explicit preservation term \citepsupp{Muttenthaler2023-aw}. If increasing or preserving downstream task performance were an additional goal beyond bringing DNNs' internal representations closer to those of humans, an explicit preservation term would be warranted \citepsupp{Muttenthaler2023-aw, Muttenthaler2025-nb}.

\clearpage
\phantomsection
\label{sec:supp_references}
\bibliographystylesupp{apa7}
\bibliographysupp{references}

\end{document}